\documentclass[letterpaper]{article} 
\usepackage{aaai2026}  
\usepackage{times}  
\usepackage{helvet}  
\usepackage{courier}  
\usepackage[hyphens]{url}  
\usepackage{graphicx} 
\usepackage{natbib}  
\usepackage{caption} 
\usepackage{algorithm}
\usepackage{algorithmic}

\usepackage{newfloat}
\usepackage{listings}
\DeclareCaptionStyle{ruled}{labelfont=normalfont,labelsep=colon,strut=off} 
\floatstyle{ruled}
\newfloat{listing}{tb}{lst}{}
\floatname{listing}{Listing}
\usepackage{amsmath}
\usepackage{amsfonts}
\usepackage{xcolor}

\usepackage{rotating}
\usepackage{booktabs}

\def\qed{\space$\square$ \par \vspace{.15in}}
\def\hat{\widehat}

\newcommand{\bz}{{\bf z}}

\newcommand{\bx}{{\bf x}}

\newcommand{\E}{\mbox{{\rm E}}}

\newcommand{\bc}{\begin{center}}
\newcommand{\ec}{\end{center}}
\newcommand{\be}{\begin{equation}}
\newcommand{\ee}{\end{equation}}
\newcommand{\ba}{\begin{array}}
\newcommand{\ea}{\end{array}}
\newcommand{\bean}{\begin{eqnarray*}}
\newcommand{\eean}{\end{eqnarray*}}
\newcommand{\bea}{\begin{eqnarray}}
\newcommand{\eea}{\end{eqnarray}}
\newcommand{\ben}{\begin{enumerate}}
\newcommand{\een}{\end{enumerate}}
\newcommand{\bed}{\begin{itemize}}
\newcommand{\eed}{\end{itemize}}

\newtheorem{assumption}{\bf Assumption}

\newtheorem{lemma}{\bf Lemma}

\newtheorem{proposition}{\bf Proposition}

\makeatletter
\def\copyright@on{F}
\def\showauthors@on{T}
\makeatother

\title{Memorize Early, Then Query: Inlier-Memorization-Guided Active Outlier Detection}
\author{
    Minseo Kang\textsuperscript{\rm 1},
    Seunghwan Park\textsuperscript{\rm 2},
    Dongha Kim\textsuperscript{\rm 1,3}\thanks{Corresponding author.}
}
\affiliations{
    \textsuperscript{\rm 1}Department of Statistics, Sungshin Women's University\\
    \textsuperscript{\rm 2}Information Statistics, Kangwon National University \\
    \textsuperscript{\rm 3}Center for Data Science, Sungshin Women's University \\
    
}

\usepackage{bibentry}

\begin{document}

\maketitle

\begin{abstract}
Outlier detection (OD) aims to identify abnormal instances, known as outliers or anomalies, by learning typical patterns of normal data, or inliers.
Performing OD under an unsupervised regime--without any information about anomalous instances in the training data--is challenging.
A recently observed phenomenon, known as the \textit{inlier-memorization (IM) effect}, where deep generative models (DGMs) tend to memorize inlier patterns during early training, provides a promising signal for distinguishing outliers. 
However, existing unsupervised approaches that rely solely on the IM effect still struggle when inliers and outliers are not well-separated or when outliers form dense clusters. 
To address these limitations, we incorporate \textit{active learning} to selectively acquire informative labels, and propose \textit{IMBoost}, a novel framework that explicitly reinforces the IM effect to improve outlier detection.
Our method consists of two stages: 1) a \textit{warm-up} phase that induces and promotes the IM effect, and 2) a \textit{polarization} phase in which actively queried samples are used to maximize the discrepancy between inlier and outlier scores.
In particular, we propose a novel query strategy and tailored loss function in the polarization phase to effectively identify informative samples and fully leverage the limited labeling budget.
We provide a theoretical analysis showing that the IMBoost consistently decreases inlier risk while increasing outlier risk throughout training, thereby amplifying their separation.
Extensive experiments on diverse benchmark datasets demonstrate that IMBoost not only significantly outperforms state-of-the-art active OD methods but also requires \textcolor{black}{substantially} less computational cost.
\end{abstract}


\section{Introduction}

\paragraph{Outlier Detection}
An outlier, or anomaly, refers to a data point that deviates significantly from the majority of typical observations \citep{pang2021deep}.
Such instances can contaminate data integrity and negatively impact downstream analyses, such as regression and classification tasks.
Therefore, identifying and removing outliers is essential for ensuring reliable data-driven modeling.
Outlier detection (OD) is the task of detecting these abnormal instances.
In addition to serving as a pre-processing step for downstream supervised learning applications, OD itself plays a critical role across various domains, including fraud detection, network intrusion detection, medical diagnosis, and sensor data monitoring \citep{nassif2021machine}.

OD tasks can be broadly categorized into three settings based on the availability of anomaly-related information in the training data.
Supervised OD (SOD) assumes access to labeled data, where each instance is explicitly annotated as either an inlier or an outlier.
Semi-supervised OD (SSOD), also referred to as out-of-distribution (OOD) detection, assumes that all training data are inliers and builds models using only these normal samples.
And unsupervised OD (UOD) operates without any label information, relying on training data that may include both inliers and outliers without explicit annotation.
In general, many real-world anomaly detection tasks belong to UOD because outliers in large datasets are usually unknown beforehand. 


To address UOD tasks, a couple of recent studies have leveraged the so-called \textit{inlier-memorization (IM) effect}, a phenomenon where DGMs tend to learn inlier patterns \textit{earlier} than outlier patterns during training. 
This arises because inliers are typically more \textit{prevalent} and \textit{densely} distributed, making it more effective to reduce their loss values first when minimizing the overall loss.
This early-phase memorization provides a useful signal for distinguishing inliers from outliers without supervision. 
ODIM \citep{DBLP:conf/icml/KimHLKK24} is the first to exploit this effect by identifying an optimal training point and using per-sample loss as an outlier score.
ALTBI \citep{DBLP:conf/aaai/ChoHBK25} further enhances the IM effect through adaptive mini-batch-size scheduling and trimmed loss optimization with a dynamic threshold. 

\begin{figure*}[t]
\centering
\includegraphics[width=0.75\textwidth]{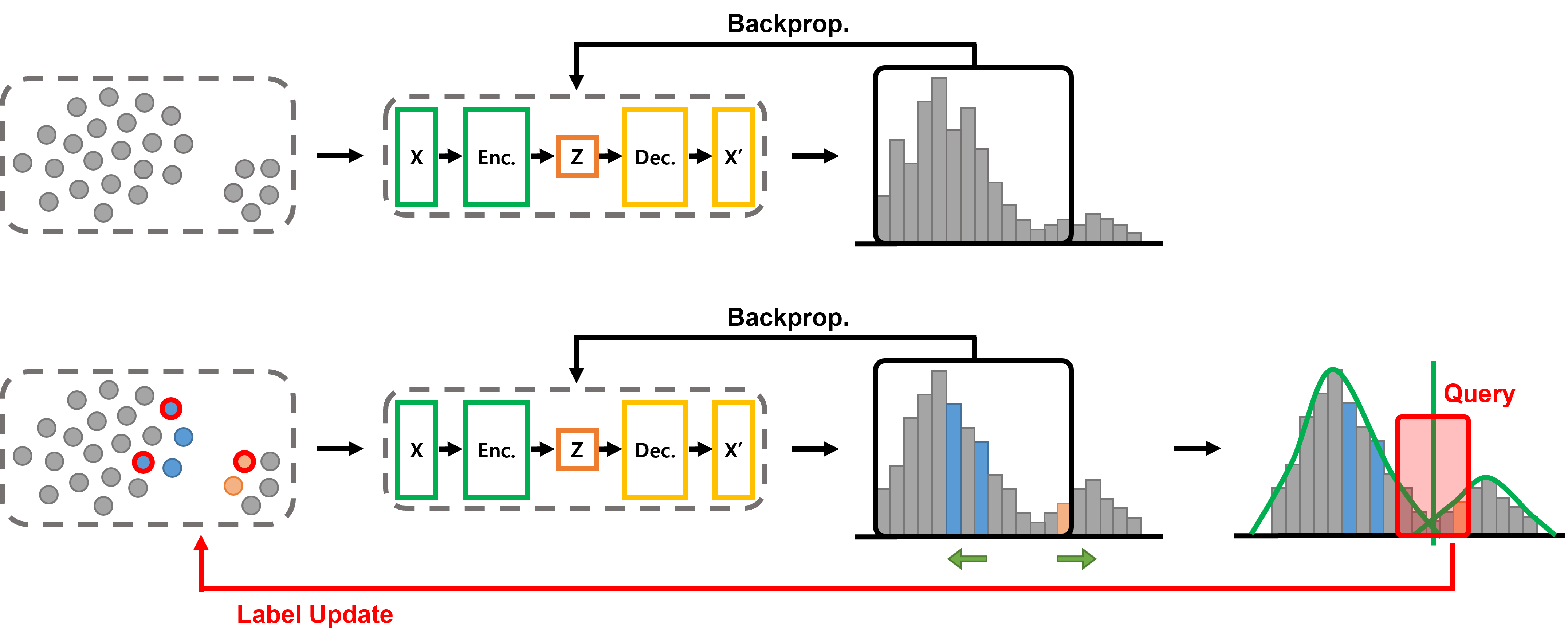} 
\caption{An illustration of IMBoost: \textbf{(Top)} \textit{warm-up} phase and \textbf{(Down)} \textit{polarization} phase.}
\label{fig:imboost_illustration}
\end{figure*}

Despite their effectiveness, these methods operate in a fully unsupervised manner, which leads to several limitations.
First, when inliers and outliers are closely distributed, it becomes inherently difficult to distinguish them without any label supervision.
Second, if outliers are more densely clustered than inliers, DGMs may incorrectly memorize outliers before inliers during early training, reversing the intended IM effect and resulting in performance degradation.
This issue is especially pronounced when outliers originate from a single point due to a measurement error, causing the model to focus on them prematurely.
These limitations underscore the need for a new paradigm that can address such failure modes of unsupervised learning.

\paragraph{Overview of Our Method}
As mentioned above, access to anomaly-related information is essential for accurately identifying outliers; however, obtaining such labels is often expensive and time-consuming.
To address this challenge, we adopt the idea of \textit{active learning} and propose \textit{IMBoost}, a novel active outlier detection framework that reinforces the IM effect to enhance detection performance.
IMBoost operates in two phases: 1) an \textit{warm-up} phase that induces and promotes the IM effect, and 2) a \textit{polarization} phase in which actively queried samples are used to maximize the discrepancy between inlier and outlier scores. 
In particular, during the polarization phase, we introduce a new querying strategy and a loss function to effectively select informative labeled samples and fully exploit their benefits.
An illustration of the IMBoost is shown in Figure \ref{fig:imboost_illustration}.

The IMBoost offers a couple of appealing features.
It consistently outperforms existing methods by a significant margin in identifying outliers.
Through extensive experiments on nearly 60 benchmark datasets, we demonstrate that the IMBoost achieves the best performance among recent approaches with large margins.
We also provide a rigorous theoretical analysis, showing that our method progressively increases the separation between inlier and outlier scores, particularly during the polarization phase.
Furthermore, by leveraging the IM effect, IMBoost achieves this performance with \textcolor{black}{considerably} lower computational cost, as it does not require full model training.
These findings highlight the IMBoost as a promising approach for practical OD tasks.

The remainder of this paper is organized as follows.
First, we review related work on outlier detection, with a particular focus on approaches that incorporate active learning.
We then present a detailed description of the proposed IMBoost framework and discuss its theoretical properties.
Next, we report extensive experimental results, including performance comparisons and ablation studies.
Finally, we conclude with a summary of our findings and outline potential directions for future work.

\section{Related Works}



We review algorithms for both SSOD and active OD, as the former are often used as base frameworks in active OD tasks.
A representative SSOD approach is support vector data description (SVDD, \citet{tax2004support}), which models normal data by enclosing it within a hypersphere in feature space.
Deep SVDD (DSVDD, \citet{ruff2018deep}) extends SVDD by mapping data into a latent space and learning a hypersphere that encloses normal samples. 
Anomalies are identified as samples that fall outside this sphere.
Deep SAD (DSAD, \citet{deepsad}) extends Deep SVDD by incorporating a few labeled anomalies, pushing them away from the center while keeping normal samples close.

In real-world scenarios, obtaining a clean dataset composed only of normal samples is challenging due to the high cost of human annotation. 
Active learning strategies have been introduced to selectively query anomaly labels for informative samples and improve outlier detection performance \citep{trittenbach2021overview}.   
Many existing strategies adopt a greedy query rule, selecting the top-1 or top-K samples with the highest anomaly scores \citep{lamba2019learning,das2016incorporating}.
However, such approaches may be suboptimal in active OD tasks, as some lower-ranked instances may provide more valuable information \citep{zha2020meta}. 

Instead, adaptive query strategies that prioritizes highly uncertain samples have been proposed to provide more informative guidance. 
\citet{kim2023active} introduced an active OD framework based to DSVDD with an uncertainty-based querying mechanism.  
The method adaptively updates the latent boundary using feedback from previously queried samples, reducing sensitivity to hyperparameters.
They also employed noise contrastive estimation (NCE, \citet{gutmann2010noise}) to iteratively refine the decision boundary using both labeled normal and abnormal samples.

\section{Proposed Method}

\subsection{Preliminaries}

\paragraph{Mathematical Notations}

We begin by introducing the notations and definitions used throughout the paper.  
Let $X_1, \ldots, X_n \in \mathcal{X} \sim \mathbb{P}$ be $n$ independent input samples drawn from the true data distribution $\mathbb{P}$, and let $\mathcal{D}^{\text{tr}}=\{\boldsymbol{x}_1,\ldots,\boldsymbol{x}_n\}$ denote their observed realization, i.e., training data.  
We assume that inlier-outlier labels are unavailable unless they are explicitly acquired through oracle queries. 

The data distribution $\mathbb{P}$ is modeled as a mixture of inlier and outlier distributions, i.e., $\mathbb{P} = (1-p_o)\cdot\mathbb{P}_{i}+p_o\cdot \mathbb{P}_{o}$, where $\mathbb{P}_{i}$ and $\mathbb{P}_{o}$ denote the inlier and outlier distributions, respectively, and $p_o \in (0,1)$ is the outlier ratio.
We define the supports of $\mathbb{P}_i$ and \( \mathbb{P}_o \) as \( \mathcal{X}_i \) and \( \mathcal{X}_o \), respectively, so that the total input space is \( \mathcal{X} = \mathcal{X}_i \cup \mathcal{X}_o \).  
We assume disjoint supports for inliers and outliers, i.e., $\mathcal{X}_i \cap \mathcal{X}_o = \emptyset$, which is a natural assumption in OD problems.




For a given sample $\boldsymbol{x}$, we define the per-sample loss with respect to a DGM as $l(\theta;\boldsymbol{x})$, such as the negative log-likelihood, where $\theta\in\Theta$ represents the model parameters of the DGM.
We assume that \( l(\theta; \boldsymbol{x}) \) is differentiable and bounded within the interval \([0, 1]\) for all \( \boldsymbol{x} \in \mathcal{X} \) and \( \theta \in \Theta \).

We denote by $\mathbb{E}_i$ and $\mathbb{E}_o$ the expectation operators with respect to $\mathbb{P}_i$ and $\mathbb{P}_o$, respectively.
The corresponding risk functions are defined as $R_i(\theta)=\mathbb{E}_i\left[ l(\theta;X) \right]$ and $R_o(\theta)=\mathbb{E}_o\left[ l(\theta;X) \right]$, which represent the expected loss over inlier and outlier distributions, respectively. 
Finally, we denote the minimizer of the inlier risk as $\theta_*$, i.e., $\theta_*=\text{argmin}_{\theta} R_i(\theta)$. 
We assume $R_i(\theta_*)=0$. 

\paragraph{Inlier-Memorization Effect}
The \textit{inlier-memorization (IM) effect} is a phenomenon observed in likelihood-based DGMs, where the model tends to learn inlier patterns \textit{earlier} than outlier patterns during training \citep{DBLP:conf/icml/KimHLKK24}. 
As training progresses, outliers are eventually learned as well, but the initial gap in memorization offers a useful signal for distinguishing them without supervision.


\citet{DBLP:conf/icml/KimHLKK24} is a pioneering work that first observed the IM effect and applied it to develop an UOD method called ODIM. 
ODIM trains a DGM for a limited number of updates and uses the per-sample loss as the outlier score, selecting the optimal update step where the IM effect is most prominent. 
To find this point, ODIM monitors the bi-modality of the loss distribution over time by fitting a two-clustered Gaussian mixture model (GMM) and measuring the separation between clusters. 

As a follow-up study, ALTBI \citep{DBLP:conf/aaai/ChoHBK25} adopts a more dynamic approach by continuously enhancing the IM effect throughout training. 
It gradually increases the batch size and employs trimmed loss optimization to filter out high-loss (potentially outlier) samples. 
This strategy helps enlarge the memorization gap between inliers and outliers during training and improves detection performance. 



\subsection{Loss Function Description of IMBoost}

Our procedure consists of two phases: 1) \textit{warm-up} and 2) \textit{polarization}. 
In the \textit{warm-up} phase, we pre-train a DGM to induce the IM effect and further refine it to promote this effect, in an unsupervised manner.
In the \textit{polarization} phase, we employ an active learning scheme to amplify the discrepancy between the inlier and outlier loss distributions.

\paragraph{Phase 1 (Warm-up)}
We begin by training a DGM using a conventional loss function to induce the IM effect. 
At each iteration, we draw a mini-batch $\mathcal{D}_0$ of size $n_0$ and compute the  loss function as follows: 
\begin{eqnarray}
\label{eq:loss_phase1}
\hat{R}_0(\theta)=\frac{1}{|\mathcal{D}_0|}\sum_{\boldsymbol{x}\in\mathcal{D}_0}l(\theta;\boldsymbol{x}).
\end{eqnarray}
The update process using \eqref{eq:loss_phase1} is iterated for $T_0$ times. 

Next, we train the DGM to further promote the IM effect.
We adopt using the strategy proposed in ALTBI \citep{DBLP:conf/aaai/ChoHBK25}, which gradually increases the mini-batch size and employs a trimmed mean loss with an adaptive threshold. 
At each iteration $t$, we sample a mini-batch $\mathcal{D}_t\subset\mathcal{D}^{\text{tr}}$ whose size increases exponentially with $t$, i.e., $|\mathcal{D}_t|=n_0\gamma^{t-1}$ for some constant $\gamma>1$.  
Instead of computing the loss over all samples in  $\mathcal{D}_t$, we use a \textit{trimmed loss function} defined as:
\begin{eqnarray}
\label{eq:loss_phase2}
\hat{R}_t(\theta)=
\frac{\sum_{\boldsymbol{x}\in\mathcal{D}_t}l(\theta;\boldsymbol{x})\cdot I(l(\theta;\boldsymbol{x})\le\tau_t)}{\sum_{\boldsymbol{x}'\in\mathcal{D}_t}I(l(\theta;\boldsymbol{x}')\le\tau_t)},
\end{eqnarray}
where $\tau_t>0$ is an adaptive threshold. 
We train the DGM with this trimmed loss \eqref{eq:loss_phase2} for $T_1$ updates.

\paragraph{Phase 2 (Polarization)}
After the warm-up phase ends, we adopt an active learning scheme.
Specifically, at each iteration, we query a small subset of samples to the oracle, resulting in two disjoint labeled subsets: $\mathcal{I}_t$ for inliers and $\mathcal{O}_t$ for outliers. 
In practice, we do not query new samples at every iteration; instead, we query them occasionally and reuse the labeled ones, gradually expanding both the inlier and outlier sets. 
Along with a randomly drawn unlabeled mini-batch $\mathcal{D}_t\subset \mathcal{D}^{\text{tr}}$ of size $|\mathcal{D}_t| = n_0 \gamma^{t-1}$ (with the same $n_0$ and $\gamma$ used in the phase 1), we use the following loss function:

\begin{eqnarray}
\label{eq:loss_phase3}
\hat{R}_t^{a}(\theta)=\hat{R}_t(\theta)+\lambda_{1,t}\cdot\hat{R}_{t}^i(\theta) - \lambda_{2,t}\cdot\hat{R}_{t}^o(\theta),
\end{eqnarray}
where 
\begin{align*}
 \hat{R}_{t}^i(\theta)=\frac{1}{|\mathcal{I}_t|}\sum_{\boldsymbol{x}\in\mathcal{I}_t}{l(\theta;\boldsymbol{x})}, \quad
\hat{R}_{t}^o(\theta)=\frac{1}{|\mathcal{O}_t|}\sum_{\boldsymbol{x}\in\mathcal{O}_t}l(\theta;\boldsymbol{x}),
\end{align*}
and  $\lambda_{1,t},\lambda_{2,t}>0$ are hyperparameters controlling the relative emphasis on inliers and outliers. 
We perform $T_2$ update steps using the loss function \eqref{eq:loss_phase3}.
After training, we compute the outlier score for a given input $\boldsymbol{x}$ as its per-sample loss, i.e., $l(\theta;\boldsymbol{x})$.
A sample is regarded as an outlier if it has a high score, and as an inlier otherwise.

The loss function in \eqref{eq:loss_phase3} can be viewed as an extended version of \eqref{eq:loss_phase2}, augmented with additional two terms computed from labeled inliers and outliers, respectively.
Minimizing the inlier loss $\hat{R}_t^i(\theta)$ is expected to promote a decrease in the inlier risk, while maximizing the outlier loss $\hat{R}_t^o(\theta)$ is intended to increase the outlier risk.
In the theoretical analysis, we rigorously validate that when the hyperparameters $\lambda_{1,t},\lambda_{2,t}$ are set proportional to $\gamma^{-t}$, the resulting optimization process indeed enlarges the gap between inliers and outliers, thereby enhancing the discriminative power of the per-sample loss as an outlier score.




\subsection{Practical Training Techniques for IMBoost}
In this section, we discuss several practical considerations that improve the performance and stability of our method during implementation.
The effectiveness of each technique is evaluated through detailed experiments, as presented in the ablation studies.

\paragraph{Implementation of $l(\theta;\boldsymbol{x})$}
Motivated by the empirical success of IM-effect-based approaches such as ODIM \citep{DBLP:conf/icml/KimHLKK24} and ALTBI \citep{DBLP:conf/aaai/ChoHBK25}, we adopt the negative log-likelihood as our loss function $l(\theta; \boldsymbol{x})$.
Various likelihood-based DGMs have been proposed, including VAE-based methods \citep{kingma2013auto}, normalizing flow (NF)-based models 
\citep{DBLP:conf/nips/KingmaD18}, 
and diffusion-based models \citep{DBLP:conf/iclr/0011SKKEP21}. 
Among them, we employ the importance-weighted autoencoder (IWAE) framework \citep{DBLP:journals/corr/BurdaGS15}, a VAE-based that provides a tight and tractable \textit{lower bound} on the log-likelihood and has demonstrated strong empirical performance in prior work.

Unlike the theoretical assumption that $l(\theta;\boldsymbol{x})$ lies within the range $[0,1]$, the upper bound of the IWAE loss often remains unsaturated in practice.
As a result, directly using $-l(\theta;\boldsymbol{x})$ as the outlier loss may lead to instability during training.
To address this issue, we adopt the CUBO framework \citep{dieng2017variational}, which provides a stable \textit{upper bound} on the log-likelihood.
As a result, we use the IWAE loss for the first two terms in \eqref{eq:loss_phase3}, and apply the CUBO loss for the last term, which corresponds to the outlier loss.
The detailed formulation is provided in Appendix A.

\paragraph{Threshold Modification}
Theoretically, the adaptive threshold $\tau_t$ is set to be the inlier risk, i.e., $\tau_t=R_i(\theta_{t-1})$. 
However, when no labeled data are available, this quantity cannot be computed directly. In such cases, using a quantile of the per-sample loss values has been shown to yield good empirical performance \citep{DBLP:conf/aaai/ChoHBK25}.
Therefore, we adopt this quantile-based strategy during the warm-up phase.
In contrast, during the polarization phase, labeled data are partially available. We incorporate this information into the computation of the adaptive threshold as follows:
\begin{align}
\label{eq:adap_thres}
\tau_t=(1-\xi)\cdot q_{\rho}(\theta_{t-1})+\xi\cdot\hat{R}_t^i(\theta_{t-1}),
\end{align}
where $q_{\rho}$ denotes the $\rho$-quantile of the per-sample loss values $l(\theta_{t-1}; \boldsymbol{x})$, and $\xi \in (0,1)$ is a weighting parameter.
We note that both hyperparameters, $\rho$ and $\xi$, are predetermined.

\paragraph{Quering Strategies}
In the polarization phase, selecting which samples to query based on per-loss values is crucial.
We consider three strategies:
\begin{itemize}
    \item \textbf{Random (RD):} randomly selecting samples for labeling.
    \item \textbf{Confidence Poles (CP):} selecting samples with the smallest and largest loss values, which are likely to be inliers and outliers, respectively. 
    Half of the queried samples are chosen from the lowest-loss instances, and the other half from the highest-loss ones. 
    \item \textbf{Decision Boundary Using Mixture Model (MM):} targeting ambiguous samples. 
    To this end, we fit a two-component GMM to the current loss values and compute each sample’s posterior probability of belonging to the inlier cluster (the one with the smaller mean). 
    We then select samples whose posterior probability is closest to a predefined threshold $\alpha\in(0,1)$.
\end{itemize}
Among the three, we adopt the MM approach with $\alpha=0.4$ as our final query strategy.

\begin{algorithm}[tb]
\caption{IMBoost\\ In practice, we set\\ 
$(n_0, \gamma, \rho, \xi, \alpha, \lambda_{1,t},\lambda_{2,t}) = (128, 1.03, 0.92, 0.4, 0.4, 2,1)$.}
\label{alg:algorithm}
\textbf{Input}: Training data: $\mathcal{D}^{\text{tr}}$,  parameters of a DGM : $\theta$, initial mini-batch size: $n_0$, mini-batch increment: $\gamma$, quantile value: $\rho$, optimizer: $\mathcal{L}$, four hyperparameters: $(\xi,\alpha,\lambda_{1,t},\lambda_{2,t})$, four time steps: $(T_0, T_1,T_2,T_a)$.
\begin{algorithmic}[1] 
\STATE Initialize $\theta_{0}$.\\
\STATE \textcolor{gray}{Phase 1: Warm-up}
\FOR{$(t=1$ \textbf{to} $T_0)$}
\STATE Draw a mini-batch with the fixed size of $n_0$, $\mathcal{D}_0=\{{\boldsymbol{x}}_{i}^{\text{mb}}\}_{i=1}^{n_0}$, from $\mathcal{D}^{\text{tr}}$. \\
Calculate the loss function $\hat{R}_0(\theta_0)$ in \eqref{eq:loss_phase1}. \\
$\theta_{0} \leftarrow \mathcal{L}(\hat{R}_0(\theta_0), \theta_0)$
\ENDFOR
\FOR{$(t=1$ \textbf{to} $T_1 )$}
\STATE Draw a mini-batch with a size of $n_t = n_0 \gamma^{t-1}$, $\mathcal{D}_t=\{\boldsymbol{x}_{i}^{\text{mb}}\}_{i=1}^{n_t}$ from $\mathcal{D}^{\text{tr}}$. \\
Calculate the loss function $\hat{R}_t(\theta_{t-1})$ in \eqref{eq:loss_phase2}.\\
$\theta_{t} \leftarrow \mathcal{L}(\hat{R}_t(\theta_{t-1}), \theta_{t-1})$
\ENDFOR
\STATE \textcolor{gray}{Phase 2: Polarization}
\STATE $\mathcal{I}_t,\mathcal{O}_t\leftarrow \emptyset,\emptyset$
\FOR{$(t=(T_1+1)$ \textbf{to} $(T_1+T_2) )$}
\STATE Draw a mini-batch with a size of $n_t = n_0 \gamma^{t-1}$, $\mathcal{D}_t=\{\boldsymbol{x}_{i}^{\text{mb}}\}_{i=1}^{n_t}$ from $\mathcal{D}^{\text{tr}}$. \\
\IF {($(t-T_1)$ \textbf{mod} $(T_2/T_a)$ \textbf{is} $0$)}
\STATE \textcolor{gray}{// Query labels and update inlier and outlier sets.} \\
Update $\mathcal{I}_t,\mathcal{O}_t$ using the MM strategy based on ensembled losses in \eqref{ensem_loss}. 
\ENDIF

Calculate the loss function $\hat{R}_t^a(\theta_{t-1})$ in \eqref{eq:loss_phase3}.\\
$\theta_{t} \leftarrow \mathcal{L}(\hat{R}_t^a(\theta_{t-1}), \theta_{t-1})$
\ENDFOR
\end{algorithmic}
\textbf{Output (IMBoost score)}: Per-sample loss $l(\theta_{T_1+T_2};\boldsymbol{x})$ computed on the training (or test) data
\end{algorithm}

For the CP and MM strategies, per-sample loss values over the training data are required, which can be unstable during training.
To mitigate this instability, we use loss scores averaged over several recent iterations before each query round.
Specifically, at the beginning of query round $t_q$, we compute the ensembled loss for each input $\boldsymbol{x}$ as:
\begin{align}
\label{ensem_loss}
l^{\text{ens}}(\boldsymbol{x})=\frac{1}{t_{e}}\sum_{t=t_q-t_e}^{t_q-1} l(\theta_t;\boldsymbol{x}),
\end{align}
and use these scores to apply the CP or MM strategies. 
In practice, we set $t_e=T_2/T_a$, where $T_a$ denotes the total number of query rounds, which is set to 5 in the experiments.


The complete pseudo-code of our proposed method, IMBoost, incorporating all the aforementioned techniques, is presented in Algorithm \ref{alg:algorithm}.

\section{Theoretical Analysis}
\label{sec:theory}

In this section, we provide a theoretical analysis of how IMBoost progressively enhances the separation between inlier and outlier risks during training. 
For simplicity in the theoretical analysis, we assume that $\mathcal{I}_t$ and $\mathcal{O}_t$ are independently drawn from $\mathbb{P}_i$ and $\mathbb{P}_o$, respectively, and that their sizes are fixed (possibly dependent on $t$). 
We also set the adaptive threshold to the inlier risk, i.e., $\tau_t = R_i(\theta_{t-1})$.

We begin by presenting three mathematical assumptions, following the setup introduced in \citet{DBLP:conf/aaai/ChoHBK25}.

\begin{assumption}[IM Effect]
\label{def_im}
There exist $0<a_1<a_2<1$ and $a_3\in(0,1-a_2)$ such that for any parameter $\theta$ satisfying $R_{i}(\theta)\in[a_1,a_2]$, $R_{o}(\theta)-R_{i}(\theta)\ge a_3$.
\end{assumption}



\begin{assumption}[Bounded and Smooth Gradient]
\label{assump_grad}
Denote the gradients of $l(\theta;\boldsymbol{x})$ and $R_i(\theta)$ as $\nabla_{\theta} l(\theta;\boldsymbol{x})$ and $\nabla_{\theta} R_i(\theta)$, respectively. Then the followings conditions are satisfied:
\begin{enumerate}
    \item[1)] For any $\boldsymbol{x}\in\mathcal{X}$ and $\theta\in\Theta$, there exists a constant $G > 0$, such that
    \begin{align*}
    \|\nabla_{\theta} l(\theta;\boldsymbol{x})\| \leq G.
    \end{align*}
    \item[2)]  $R_i(\theta)$ and $R_o(\theta)$ are smooth with a $L$-Lipschitz continuous gradient, i.e., there exists a constant $L > 0$ such that
    \begin{align*}
    &\|\nabla_{\theta} R_i(\theta) - \nabla_{\theta} R_i(\theta')\| \leq L \|\theta - \theta'\|, \quad \text{and} \\
    & \|\nabla_{\theta} R_o(\theta) - \nabla_{\theta} R_o(\theta')\| \leq L \|\theta - \theta'\|, \quad \forall \theta, \theta' \in \Theta.
    \end{align*}
    \item[3)]  There exists $\mu > 0$ such that for any $\theta\in\Theta$,
    \begin{align*}
    2\mu (R_i(\theta) -R_i(\theta_*))=2\mu R_i(\theta) \leq \|\nabla_{\theta} R_i(\theta)\|^2. 
    \end{align*}

    \end{enumerate}
\end{assumption}

\begin{assumption}[Uniform Margin]
\label{assump_unif}
There is a constant $0<c<1$ such that, for any $\theta$, the following inequality holds: 
$$\mathbb{E}_i\left[\sqrt{l(\theta;X)}\right]\le (1-c)\sqrt{R_i(\theta)}.$$     
\end{assumption}

Assumption \ref{def_im} describes a state of the DGM in which a gap has emerged between the risk values of inliers and outliers.
Assumption \ref{assump_grad} refers to the smoothness properties of the loss function and the associated risk functions. 
And Assumption \ref{assump_unif} ensures that Jensen's inequality holds with a uniform margin for any parameter $\theta$.

Additionally, we introduce an assumption stating that, under the presence of the IM effect, a gradient discrepancy exists between inliers and outliers.
\begin{assumption}[Gradient Discrepancy]
\label{assump_disc}
There are two constant $c_1,c_2>0$ such that, for any $\theta$ satisfying the IM effect (i.e., Assumption \ref{def_im}), the following inequalities hold: 
$$\|\nabla_{\theta}R_o(\theta)\|^2 \ge c_1 \quad\text{and}\quad \nabla_{\theta}R_i(\theta)^T \nabla_{\theta}R_o(\theta)\le -c_2.$$
\end{assumption}

Assumption \ref{assump_disc} implies that, during the IM effect, the gradients of inlier and outlier risks not only point in different directions, but also cannot be minimized simultaneously.
Based on the above assumptions, we establish the following two propositions. The proof of the second proposition is provided in the Appendix B.
We note that Proposition \ref{prop:1} is identical to Proposition 2 in \citet{DBLP:conf/aaai/ChoHBK25}, and therefore we omit its proof in this paper.

\begin{proposition}[Warm-up Property (\citet{DBLP:conf/aaai/ChoHBK25})]
\label{prop:1} 
Suppose that Assumptions \ref{def_im} to \ref{assump_unif} hold.  
During the \textit{warm-up} phase (i.e., when $t \le T_1$), assume that the current parameter $\theta_{t-1}$ satisfies $a_1 \le R_i(\theta_{t-1}) \le a_2 \gamma^{-(t-1)}$.  
Then for any given $\delta > 0$, there exists a learning rate $\eta_t > 0$ such that the inlier risk is further reduced, i.e., $R_i(\theta_t) \le \gamma^{-1}R_i(\theta_{t-1})$, with probability at least $1 - \delta$.
\end{proposition}

\begin{proposition}[Polarization Property]
\label{prop:2} 
Suppose that Assumptions \ref{def_im} to \ref{assump_disc} are satisfied.  
During the \textit{polarization} phase (i.e., when $t > T_1$), suppose that the current parameter $\theta_{t-1}$ satisfies $a_1 \le R_i(\theta_{t-1}) \le a_2 \gamma^{-(t-1)}$.  
Then, for any given $\delta > 0$, there exists a learning rate $\eta_t > 0$ and  $\lambda_{1,t}$ and $\lambda_{2,t}$, both proportional to $\gamma^{-(t-T_1-1)}$, such that the following holds with probability at least $1 - \delta$:
\begin{align*}
R_i(\theta_{t}) \le \gamma^{-1}R_i(\theta_{t-1}) \quad \text{and} \quad R_o(\theta_t) > R_o(\theta_{t-1}).
\end{align*}
\end{proposition}

Proposition \ref{prop:2} indicates that, during the polarization phase, our proposed loss function \eqref{eq:loss_phase3} guarantees a continued reduction in inlier risk and simultaneously induces an increase in outlier risk, thereby enlarging the discrepancy between inlier and outlier risks.

\begin{figure}[t]
\centering
\includegraphics[width=0.45\textwidth]{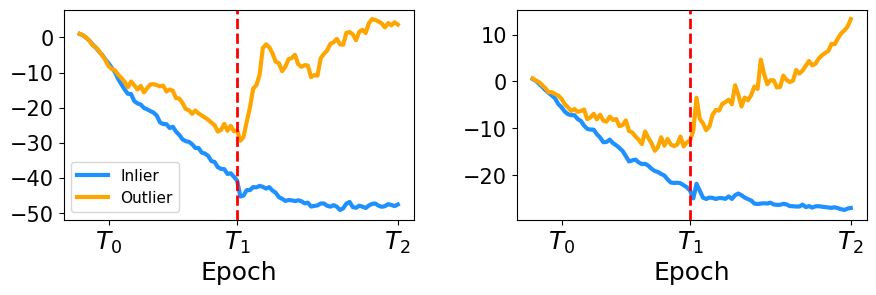}
\caption{
\textbf{(1st and 2nd)} Trace plots of inlier and outlier risks during the warm-up and polarization phases on the \texttt{PageBlocks} and \texttt{Thyroid} datasets, respectively.
}
\label{fig:theoretical_implications}
\end{figure}

\paragraph{Theoretical Implications}
These theoretical results provide key insights into our proposed method, with further discussion available in Appendix B.
First, a natural question arises regarding \textit{the necessity of employing the ALTBI loss function during the warm-up phase}, as initiating active learning immediately after the emergence of the IM effect may appear more intuitive and efficient by reducing the number of iterations.
And this concern is addressed by our theoretical results, particularly Lemma 4 in Appendix B.
Lemma 4 shows that repeated use of the trimmed loss function gradually reduces the proportion of outliers included in it.
In the early stages following the emergence of the IM effect, the trimmed loss function \eqref{eq:loss_phase2} still includes a relatively high proportion of outliers.
As a result, minimizing \eqref{eq:loss_phase2} at this point may unintentionally lead to a reduction in outlier risk, as illustrated in Figure \ref{fig:theoretical_implications}.
Therefore, applying active learning at this stage does not guarantee a meaningful increase in outlier risk.

Using a large value of $\lambda_{2,t}$ to amplify the effect of outlier loss term seems to be a plausible alternative. 
However, this approach may also adversely impact the inlier loss, potentially increasing the inlier risk, which is an undesirable outcome.
Therefore, it is crucial to first reduce the influence of outliers by minimizing \eqref{eq:loss_phase2} for several iterations--until it no longer decreases the outlier risk--and then introduce active learning to simultaneously enhance the reduction of inlier risk and the increase of outlier risk.

The second question is about \textit{the optimal ratio of labeled inlier and outlier samples}.  
The proof of Proposition \ref{prop:2} reveals that the increase in outlier risk is maximized when the sizes of the inlier and outlier sets are proportional to their corresponding hyperparameters, i.e., $|\mathcal{I}_t| \propto \lambda_{1,t}$ and $|\mathcal{O}_t| \propto \lambda_{2,t}$.
Assuming that the two hyperparameters are set to similar values, and the number of queried samples is fixed (i.e., $|\mathcal{I}_t| + |\mathcal{O}_t|=n_a$ with a constant $n_a$), the outlier risk increases the most when the sets are approximately balanced (i.e., $|\mathcal{I}_t| \approx |\mathcal{O}_t|\approx n_a/2$). 
This implies that we should query as \textit{many outliers} as inliers when acquiring labeled data.
This finding provides a theoretical justification for why using posterior probability thresholds close to 0.5 in the MM strategy generally yield the favorable results, as shown in the ablation study in the experimental section. 

\begin{figure}[t]
\centering
\includegraphics[width=0.45\textwidth]{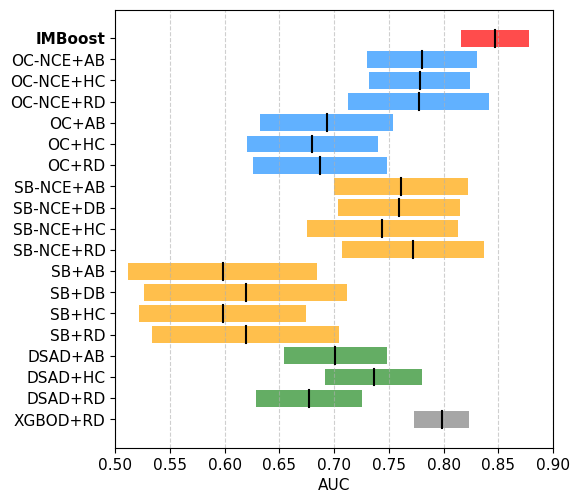}
\caption{Averaged test AUC results at the final (5th) round across 57 datasets from ADBench, with standard deviations over three independent runs. 
All implementations were done by us. 
Color scheme: red (IMBoost), blue (OC-based), orange (SB-based), and green (DSAD-based).
}
\label{fig:performace_results}
\end{figure}

\begin{figure}[t]
\centering
\includegraphics[width=0.38\textwidth]{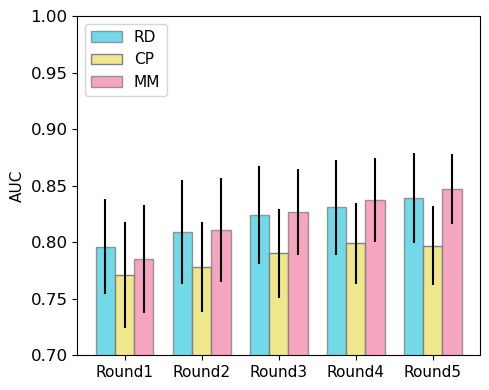}
\caption{
Averaged test AUC results (with standard deviations) of IMBoost using different querying strategies: 1) Random (RD), 2) Confidence Poles (CP), and 3) Mixture Model-based decision boundary (MM). 
Results are reported at the end of each active learning round. 
}
\label{fig:rd_cp_mm}
\end{figure}

\section{Experiments}
We validate the effectiveness of IMBoost through comprehensive experiments on 57 benchmark datasets across image, text, and tabular domains.
The results show that IMBoost achieves state-of-the-art outlier detection performance with notably higher accuracy and lower computational cost.
For each experiment, we report the average results over three independent runs with different random initializations.
All experiments are conducted using the \texttt{PyTorch} framework on two NVIDIA RTX 3090 GPUs.
The implementation code for our method is available at
\textcolor[rgb]{0,0.08,0.45}{https://github.com/mskang0306/IMBoost}.

\paragraph{Dataset Description} 
We evaluate our method on all 57 benchmark datasets from \texttt{ADBench} \citep{han2022adbench}, following the same preprocessing and dataset settings used in ALTBI \citep{DBLP:conf/aaai/ChoHBK25}.
We first consider 46 widely used tabular datasets covering diverse domains such as healthcare, finance, and astronautics.
Additionally, we include five text datasets using BERT \citep{devlin-etal-2019-bert} or RoBERTa \citep{DBLP:journals/corr/abs-1907-11692} embeddings provided by \texttt{ADBench}.
Finally, we evaluate on six image datasets using ViT-based embeddings \citep{DBLP:conf/iclr/DosovitskiyB0WZ21}, also from \texttt{ADBench}.
Full dataset details are provided in Appendix C.

All datasets are min-max normalized before training.
Each dataset is randomly split into training and test sets with a 7:3 ratio.
Following \citet{kim2023active}, we perform 5 rounds of active learning, that is, $T_a=5$, setting the query budget 1\% of each dataset.
For small datasets (fewer than 500 samples), such as \texttt{ionosphere}, \texttt{arrhythmia}, and \texttt{glass}, we instead query 6 samples per round.

\begin{figure*}[t]
\centering
\includegraphics[width=1.0\textwidth]{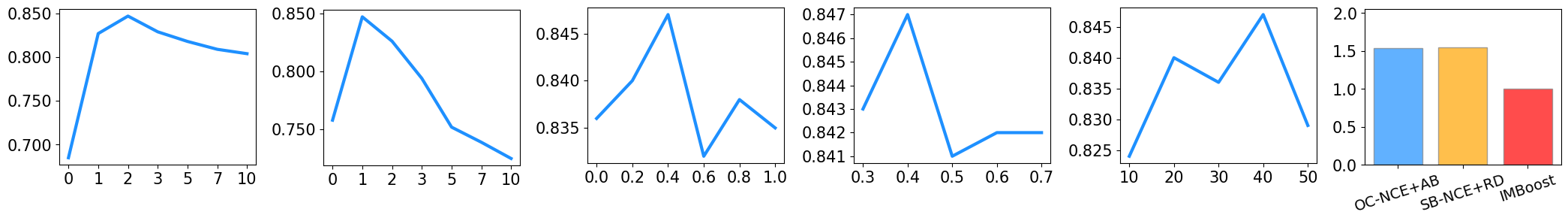}
\caption{(\textbf{1st–5th}) AUC results with varying values of hyperparameters: 1) $\lambda_{1,t}$, 2) $\lambda_{2,t}$, 3) $\xi$, 4) $\alpha$, and 5) $T_1$. 
(\textbf{6th}) Comparison of running time between the IMBoost and other approaches. 
Each runtime is rescaled relative to that of IMBoost.
}
\label{fig:ablation_results}
\end{figure*}

\paragraph{Competing Methods}
To evaluate the effectiveness of the proposed IMBoost framework, we compare it against recent state-of-the-art deep learning-based active OD methods. 
In particular, we mainly follow the experimental settings and baseline configurations analyzed in \citet{kim2023active}. 
As a baseline approach, we first consider DSVDD \citep{ruff2018deep}, originally developed for dealing with SSOD tasks. 
Following \citet{kim2023active}, we adopt two variants of DSVDD: one-class DSVDD (OC) and soft-boundary DSVDD (SB). 
To enhance the discrepancy between inlier and outlier scores, we additionally incorporate noise contrastive estimation (NCE) as a regularization technique. 
As another baseline, we include DSAD \citep{deepsad}, an advanced extension of DSVDD leveraging labeled samples.
We also consider a non–deep learning–based approach, XGBOD \citep{DBLP:conf/ijcnn/ZhaoH18}, known as one of the strongest OD solvers.

Based on these models, we explore various query strategies for active learning: 1) random sampling (RD), 2) high-confidence sampling (HC) that prioritizes highly abnormal instances, and 3) adaptive boundary (AB), which dynamically searches for a querying region where the ratio of inliers and outliers near the boundary is balanced. 
For XGBOD, we only adapt the RD strategy. 
For SB, we further consider an additional query strategy, called 4) decision boundary (DB), which selects samples near the estimated decision boundary. 
We follow all implementation configurations described in \citet{kim2023active}. 

\paragraph{Implementation Details}
As the DGM framework, we use IWAE, an ELBO-based model that has been shown to be both effective and efficient for IM-based methods by \citet{DBLP:conf/aaai/ChoHBK25}. 
We adopt the same encoder and decoder architectures as in \citet{DBLP:conf/aaai/ChoHBK25}, with detailed specifications provided in Appendix A.

We use the Adam optimizer \citep{kingma2014adam} with a learning rate of $1 \times 10^{-3}$. Our method involves several hyperparameters. 
For the warm-up phase, we use the same settings as in \citet{DBLP:conf/aaai/ChoHBK25}, namely $(n_0, \gamma, \rho) = (128, 1.03, 0.92)$, and set $(T_0, T_1) = (10, 40)$. 
In the polarization phase, we retain the same values for $(n_0, \gamma, \rho)$ and set $(\lambda_{1,t}, \lambda_{2,t}, \xi, \alpha) = (2, 1, 0.4, 0.4)$. 
Each active learning round is trained for 10 epochs, resulting in a total of 50 epochs for the polarization phase, i.e., $T_2=50$.
Note that, given the relatively small number of rounds (five), we keep the hyperparameters $(\lambda_{1,t}, \lambda_{2,t})$ constant across all rounds.
Among the three querying strategies, we adopt the mixture model-based strategy (MM) as the default querying method.  

\subsection{Performance Results}
We compare the anomaly detection performance of our method with other baselines. 
For each dataset, we compute the mean and standard deviation of the area under the ROC curve (AUC) and average precision (AP) scores over three independent runs.
Figure \ref{fig:performace_results} presents the average AUC scores and their standard deviations on the test data, aggregated across all datasets. 
Detailed results for each dataset, including AP metrics and additional evaluations on the training data, are provided in Appendix C.

We observe that our method achieves the best performance with an average margin of nearly 5\% over the second-best baseline (XGBOD+RD).
Moreover, it exhibits small standard deviations, indicating that IMBoost is not only superior in accuracy but also consistent in detecting outliers. 
Given that our method requires only a short warm-up phase, whereas other baselines involve pre-training over 100 epochs, IMBoost offers a highly practical and efficient solution for a wide range of OD tasks.

We also compare the performance of the three querying strategies--RD, CP, and MM--within the IMBoost framework. 
As shown in Figure~\ref{fig:rd_cp_mm}, the MM strategy usually outperforms both RD and CP throughout all rounds, while CP yields the worst performance among the three. 
This result aligns with the intuition that samples near the decision boundary--targeted by MM--are the most informative for improving the model.
In contrast, CP selects samples with the highest and lowest losses, which are often already well-classified and thus provide limited information gain--even less than randomly selected ones.

Inspired by RD's strong performance in the first round, we apply RD for the first round and MM for the remaining rounds, but this results in slightly degraded performance (0.840 vs. 0.847 with MM only). 
From the perspective of viewing RD and MM as \textit{exploration} and \textit{exploitation}, respectively, designing an advanced querying schedule that optimally combines the two would be an intriguing direction for future work.

\subsection{Ablation Studies}

We conduct additional experiments to examine the sensitivity of IMBoost to various hyperparameter choices, with the results illustrated in Figure~\ref{fig:ablation_results}.
To summarize:
1) Setting either $\lambda_{1,t}$ or $\lambda_{2,t}$ to zero leads to a significant performance drop.
2) Incorporating inlier risk from $\mathcal{I}_t$ when determining the adaptive threshold $\tau_t$ improves performance.
3) Performance remains stable as long as an appropriate posterior probability $\alpha$ is chosen.
4) Setting $T_1 = 40$ for the warm-up phase yields optimal results.
5) Our method has the highest computational efficiency among the competitive baselines (excluding XGBOD, which is highly inefficient for large datasets).
The detailed results are provided in Appendix C. 


\section{Concluding Remarks}
In this paper, we proposed IMBoost, a novel active outlier detection framework that leverages and enhances the IM effect in DGMs. 
Our method operates in two phases: a warm-up phase that induces and promotes the IM effect, and a polarization phase that leverages actively queried samples to simultaneously reduce inlier risk and increase outlier risk.
We theoretically showed that our approach indeed increases the separation between inlier and outlier risks, and empirically demonstrated that IMBoost significantly outperforms existing methods while remaining efficient.
A promising future work is to explore optimal querying strategies, for instance by incorporating advanced active learning techniques developed in other domains
\citep{DBLP:conf/iclr/SenerS18,DBLP:conf/nips/KirschAG19,DBLP:conf/cvpr/YooK19}.

\section*{Acknowledgements}
DK was supported by the National Research Foundation of Korea(NRF) grant funded by the Korea government(MSIT) (RS2023-00218231 and RS-2025-24683613).
SP was supported by Basic Science Research Program through the National Research Foundation of Korea(NRF) funded by the Ministry of Education(RS-2025-25415913).

\bibliography{references}

@article{pang2021deep,
  title={Deep learning for anomaly detection: A review},
  author={Pang, Guansong and Shen, Chunhua and Cao, Longbing and Hengel, Anton Van Den},
  journal={ACM computing surveys (CSUR)},
  volume={54},
  number={2},
  pages={1--38},
  year={2021},
  publisher={ACM New York, NY, USA}
}

@article{nassif2021machine,
  title={Machine learning for anomaly detection: A systematic review},
  author={Nassif, Ali Bou and Talib, Manar Abu and Nasir, Qassim and Dakalbab, Fatima Mohamad},
  journal={Ieee Access},
  volume={9},
  pages={78658--78700},
  year={2021},
  publisher={Ieee}
}

@inproceedings{DBLP:conf/icml/KimHLKK24,
  author       = {Dongha Kim and
                  Jaesung Hwang and
                  Jongjin Lee and
                  Kunwoong Kim and
                  Yongdai Kim},
  title        = {{ODIM:} Outlier Detection via Likelihood of Under-Fitted Generative
                  Models},
  booktitle    = {Forty-first International Conference on Machine Learning, {ICML} 2024,
                  Vienna, Austria, July 21-27, 2024},
  publisher    = {OpenReview.net},
  year         = {2024},
  url          = {https://openreview.net/forum?id=R8nbccD7kv},
  bibsource    = {dblp computer science bibliography, https://dblp.org}
}

@misc{kim2024odimoutlierdetectionlikelihood,
      title={ODIM: Outlier Detection via Likelihood of Under-Fitted Generative Models}, 
      author={Dongha Kim and Jaesung Hwang and Jongjin Lee and Kunwoong Kim and Yongdai Kim},
      year={2024},
      eprint={2301.04257},
      archivePrefix={arXiv},
      primaryClass={stat.ML},
      url={https://arxiv.org/abs/2301.04257}, 
}

@inproceedings{DBLP:conf/aaai/ChoHBK25,
  author       = {Seoyoung Cho and
                  Jaesung Hwang and
                  Kwan{-}Young Bak and
                  Dongha Kim},
  editor       = {Toby Walsh and
                  Julie Shah and
                  Zico Kolter},
  title        = {{ALTBI:} Constructing Improved Outlier Detection Models via Optimization
                  of Inlier-Memorization Effect},
  booktitle    = {AAAI-25, Sponsored by the Association for the Advancement of Artificial
                  Intelligence, February 25 - March 4, 2025, Philadelphia, PA, {USA}},
  pages        = {11544--11552},
  publisher    = {{AAAI} Press},
  year         = {2025},
  url          = {https://doi.org/10.1609/aaai.v39i11.33256},
  doi          = {10.1609/AAAI.V39I11.33256},
  bibsource    = {dblp computer science bibliography, https://dblp.org}
}

@article{tax2004support,
  title={Support vector data description},
  author={Tax, David MJ and Duin, Robert PW},
  journal={Machine learning},
  volume={54},
  number={1},
  pages={45--66},
  year={2004},
  publisher={Springer}
}

@inproceedings{ruff2018deep,
  title={Deep one-class classification},
  author={Ruff, Lukas and Vandermeulen, Robert and Goernitz, Nico and Deecke, Lucas and Siddiqui, Shoaib Ahmed and Binder, Alexander and M{\"u}ller, Emmanuel and Kloft, Marius},
  booktitle={International conference on machine learning},
  pages={4393--4402},
  year={2018},
  organization={PMLR}
}

@inproceedings{deepsad,
title={Deep Semi-Supervised Anomaly Detection},
author={Lukas Ruff and Robert A. Vandermeulen and Nico Görnitz and Alexander Binder and Emmanuel Müller and Klaus-Robert Müller and Marius Kloft},
booktitle={International Conference on Learning Representations},
year={2020},
}

@article{trittenbach2021overview,
  title={An overview and a benchmark of active learning for outlier detection with one-class classifiers},
  author={Trittenbach, Holger and Englhardt, Adrian and B{\"o}hm, Klemens},
  journal={Expert Systems with Applications},
  volume={168},
  pages={114372},
  year={2021},
  publisher={Elsevier}
}

@inproceedings{lamba2019learning,
  title={Learning on-the-job to re-rank anomalies from top-1 feedback},
  author={Lamba, Hemank and Akoglu, Leman},
  booktitle={Proceedings of the 2019 SIAM International Conference on Data Mining},
  pages={612--620},
  year={2019},
  organization={SIAM}
}

@inproceedings{das2016incorporating,
  title={Incorporating expert feedback into active anomaly discovery},
  author={Das, Shubhomoy and Wong, Weng-Keen and Dietterich, Thomas and Fern, Alan and Emmott, Andrew},
  booktitle={2016 IEEE 16th International Conference on Data Mining (ICDM)},
  pages={853--858},
  year={2016},
  organization={IEEE}
}

@inproceedings{zha2020meta,
  title={Meta-AAD: Active anomaly detection with deep reinforcement learning},
  author={Zha, Daochen and Lai, Kwei-Herng and Wan, Mingyang and Hu, Xia},
  booktitle={2020 IEEE International Conference on Data Mining (ICDM)},
  pages={771--780},
  year={2020},
  organization={IEEE}
}

@article{kim2023active,
  title={Active anomaly detection based on deep one-class classification},
  author={Kim, Minkyung and Kim, Junsik and Yu, Jongmin and Choi, Jun Kyun},
  journal={Pattern Recognition Letters},
  volume={167},
  pages={18--24},
  year={2023},
  publisher={Elsevier}
}

@inproceedings{gutmann2010noise,
  title={Noise-contrastive estimation: A new estimation principle for unnormalized statistical models},
  author={Gutmann, Michael and Hyv{\"a}rinen, Aapo},
  booktitle={Proceedings of the thirteenth international conference on artificial intelligence and statistics},
  pages={297--304},
  year={2010},
  organization={JMLR Workshop and Conference Proceedings}
}

@article{kingma2013auto,
  title={Auto-encoding variational bayes},
  author={Kingma, Diederik P and Welling, Max},
  journal={arXiv preprint arXiv:1312.6114},
  year={2013}
}

@inproceedings{DBLP:conf/nips/KingmaD18,
  author       = {Diederik P. Kingma and
                  Prafulla Dhariwal},
  editor       = {Samy Bengio and
                  Hanna M. Wallach and
                  Hugo Larochelle and
                  Kristen Grauman and
                  Nicol{\`{o}} Cesa{-}Bianchi and
                  Roman Garnett},
  title        = {Glow: Generative Flow with Invertible 1x1 Convolutions},
  booktitle    = {Advances in Neural Information Processing Systems 31: Annual Conference
                  on Neural Information Processing Systems 2018, NeurIPS 2018},
  pages        = {10236--10245},
  year         = {2018},
  bibsource    = {dblp computer science bibliography, https://dblp.org}
}

@inproceedings{DBLP:conf/iclr/0011SKKEP21,
  author       = {Yang Song and
                  Jascha Sohl{-}Dickstein and
                  Diederik P. Kingma and
                  Abhishek Kumar and
                  Stefano Ermon and
                  Ben Poole},
  title        = {Score-Based Generative Modeling through Stochastic Differential Equations},
  booktitle    = {9th International Conference on Learning Representations, {ICLR} 2021
                  },
  publisher    = {OpenReview.net},
  year         = {2021},
  url          = {https://openreview.net/forum?id=PxTIG12RRHS},
  bibsource    = {dblp computer science bibliography, https://dblp.org}
}

@inproceedings{DBLP:journals/corr/BurdaGS15,
  author    = {Yuri Burda and
               Roger B. Grosse and
               Ruslan Salakhutdinov},

  title     = {Importance Weighted Autoencoders},
  booktitle = {4th International Conference on Learning Representations, {ICLR} 2016,
               Conference Track Proceedings},
  year      = {2016},
  bibsource = {dblp computer science bibliography, https://dblp.org}
}

@article{dieng2017variational,
  title={Variational Inference via $\chi$-Upper Bound Minimization},
  author={Dieng, Adji Bousso and Tran, Dustin and Ranganath, Rajesh and Paisley, John and Blei, David},
  journal={Advances in Neural Information Processing Systems},
  volume={30},
  year={2017}
}

@inproceedings{han2022adbench,  
      title={ADBench: Anomaly Detection Benchmark},   
      author={Songqiao Han and Xiyang Hu and Hailiang Huang and Mingqi Jiang and Yue Zhao},  
      booktitle={Neural Information Processing Systems (NeurIPS)},
      year={2022},  
}

@inproceedings{devlin-etal-2019-bert,
    title = "{BERT}: Pre-training of Deep Bidirectional Transformers for Language Understanding",
    author = "Devlin, Jacob  and
      Chang, Ming-Wei  and
      Lee, Kenton  and
      Toutanova, Kristina",
    booktitle = "Proceedings of the 2019 Conference of the North {A}merican Chapter of the Association for Computational Linguistics: Human Language Technologies, Volume 1 (Long and Short Papers)",
    month = jun,
    year = "2019",
    publisher = "Association for Computational Linguistics",


    pages = "4171--4186",
}

@article{DBLP:journals/corr/abs-1907-11692,
  author       = {Yinhan Liu and
                  Myle Ott and
                  Naman Goyal and
                  Jingfei Du and
                  Mandar Joshi and
                  Danqi Chen and
                  Omer Levy and
                  Mike Lewis and
                  Luke Zettlemoyer and
                  Veselin Stoyanov},
  title        = {RoBERTa: {A} Robustly Optimized {BERT} Pretraining Approach},
  journal      = {CoRR},
  volume       = {abs/1907.11692},
  year         = {2019},

  eprinttype    = {arXiv},
  eprint       = {1907.11692},
  bibsource    = {dblp computer science bibliography, https://dblp.org}
}

@inproceedings{DBLP:conf/iclr/DosovitskiyB0WZ21,
  author       = {Alexey Dosovitskiy and
                  Lucas Beyer and
                  Alexander Kolesnikov and
                  Dirk Weissenborn and
                  Xiaohua Zhai and
                  Thomas Unterthiner and
                  Mostafa Dehghani and
                  Matthias Minderer and
                  Georg Heigold and
                  Sylvain Gelly and
                  Jakob Uszkoreit and
                  Neil Houlsby},
  title        = {An Image is Worth 16x16 Words: Transformers for Image Recognition
                  at Scale},
  booktitle    = {9th International Conference on Learning Representations, {ICLR} 2021},
  publisher    = {OpenReview.net},
  year         = {2021},
  bibsource    = {dblp computer science bibliography, https://dblp.org}
}

@inproceedings{DBLP:conf/ijcnn/ZhaoH18,
  author       = {Yue Zhao and
                  Maciej K. Hryniewicki},
  title        = {{XGBOD:} Improving Supervised Outlier Detection with Unsupervised
                  Representation Learning},
  booktitle    = {2018 International Joint Conference on Neural Networks, {IJCNN} 2018,
                  Rio de Janeiro, Brazil, July 8-13, 2018},
  pages        = {1--8},
  publisher    = {{IEEE}},
  year         = {2018},
  url          = {https://doi.org/10.1109/IJCNN.2018.8489605},
  doi          = {10.1109/IJCNN.2018.8489605},
  bibsource    = {dblp computer science bibliography, https://dblp.org}
}

@article{kingma2014adam,
  title={Adam: A method for stochastic optimization},
  author={Kingma, Diederik P and Ba, Jimmy},
  journal={arXiv preprint arXiv:1412.6980},
  year={2014}
}

@inproceedings{DBLP:conf/iclr/SenerS18,
  author       = {Ozan Sener and
                  Silvio Savarese},
  title        = {Active Learning for Convolutional Neural Networks: {A} Core-Set Approach},
  booktitle    = {6th International Conference on Learning Representations, {ICLR} 2018,
                  Vancouver, BC, Canada, April 30 - May 3, 2018, Conference Track Proceedings},
  publisher    = {OpenReview.net},
  year         = {2018},
  url          = {https://openreview.net/forum?id=H1aIuk-RW},
  bibsource    = {dblp computer science bibliography, https://dblp.org}
}

@inproceedings{DBLP:conf/nips/KirschAG19,
  author       = {Andreas Kirsch and
                  Joost van Amersfoort and
                  Yarin Gal},
  title        = {BatchBALD: Efficient and Diverse Batch Acquisition for Deep Bayesian
                  Active Learning},
  booktitle    = {Advances in Neural Information Processing Systems 32: Annual Conference
                  on Neural Information Processing Systems 2019, NeurIPS 2019, December
                  8-14, 2019, Vancouver, BC, Canada},
  pages        = {7024--7035},
  year         = {2019},
  url          = {https://proceedings.neurips.cc/paper/2019/hash/95323660ed2124450caaac2c46b5ed90-Abstract.html},
  bibsource    = {dblp computer science bibliography, https://dblp.org}
}

@inproceedings{DBLP:conf/cvpr/YooK19,
  author       = {Donggeun Yoo and
                  In So Kweon},
  title        = {Learning Loss for Active Learning},
  booktitle    = {{IEEE} Conference on Computer Vision and Pattern Recognition, {CVPR}
                  2019, Long Beach, CA, USA, June 16-20, 2019},
  pages        = {93--102},
  publisher    = {Computer Vision Foundation / {IEEE}},
  year         = {2019},
  url          = {http://openaccess.thecvf.com/content\_CVPR\_2019/html/Yoo\_Learning\_Loss\_for\_Active\_Learning\_CVPR\_2019\_paper.html},
  doi          = {10.1109/CVPR.2019.00018},
  bibsource    = {dblp computer science bibliography, https://dblp.org}
}

@article{chung2006concentration,
  title={Concentration inequalities and martingale inequalities: a survey},
  author={Chung, Fan and Lu, Linyuan},
  journal={Internet mathematics},
  volume={3},
  number={1},
  pages={79--127},
  year={2006},
  publisher={Taylor \& Francis}
}

@article{ghadimi2016mini,
  title={Mini-batch stochastic approximation methods for nonconvex stochastic composite optimization},
  author={Ghadimi, Saeed and Lan, Guanghui and Zhang, Hongchao},
  journal={Mathematical Programming},
  volume={155},
  number={1},
  pages={267--305},
  year={2016},
  publisher={Springer}
}

\appendix

\onecolumn

\vskip 0.2in
\begin{center}
    {\huge \textbf{Supplementary material for}}
    \\
    {\huge \textbf{``Memorize Early, Then Query: Inlier-Memorization-Guided Active Outlier Detection''}}
\end{center}
\vskip 0.2in

\section{A. Detailed Description of the Loss Function}

We begin by introducing a latent-variable-based DGM that we adopt in our study.  
Let the decoder and encoder be denoted by $p(\boldsymbol{x} | \boldsymbol{z};\theta)$ and $q(\boldsymbol{z} | \boldsymbol{x};\phi)$, respectively, where the latent variable $ \boldsymbol{z} \in \mathbb{R}^d $ typically lies in a lower-dimensional space than the input space $\mathcal{X}$.
The parameters \( \theta \) and \( \phi \) correspond to the decoder and encoder, respectively. 
Following \citet{kim2024odimoutlierdetectionlikelihood}, we implement both the encoder and decoder using two-layer deep neural networks, with each hidden layer consisting of 50 to 100 nodes.
The generative process assumes that the latent variable is drawn from a standard normal prior, $ \boldsymbol{Z} \sim \mathcal{N}(\mathbf{0}_d, \mathbf{I}_d) $, and the observation is generated via:
\begin{align*}
\boldsymbol{X} \sim p(\boldsymbol{x} \mid \boldsymbol{Z};\theta).    
\end{align*}


\subsection{A-\MakeUppercase{\romannumeral 1}. Description of Importance Weighted Autoencoders (IWAE)}

\textit{Importance Weighted Autoencoders (IWAE)} \citep{DBLP:journals/corr/BurdaGS15} is a variational inference method that yields arbitrarily tight lower bounds on the log-likelihood. Given an input \(\boldsymbol{x}\), IWAE derives the following lower bound by utilizing multiple samples drawn from the variational distribution \(q(\boldsymbol{z} | \boldsymbol{x}; \phi)\):

\begin{align*}
\mathbb{E}_{\boldsymbol{z}_1,...,\boldsymbol{z}_K\sim q(\boldsymbol{z}|\boldsymbol{x};\phi)}\left[ \log \left( \frac{1}{K} \sum_{k=1}^K \frac{p(\boldsymbol{x},\boldsymbol{z}_k;\theta)}{q(\boldsymbol{z}_k|\boldsymbol{x};\phi)} \right) \right],
\end{align*}
where \(K\) denotes the number of samples, which are set $K=2$ in practice. 
Thus, for a given input $\boldsymbol{x}$, the loss function used in the first two terms of our objective is defined as the negative of the above expression:
\begin{align}
\label{eq:iwae}
l_i(\theta,\phi;\boldsymbol{x})=-\mathbb{E}_{\boldsymbol{z}_1,...,\boldsymbol{z}_K\sim q(\boldsymbol{z}|\boldsymbol{x};\phi)}\left[ \log \left( \frac{1}{K} \sum_{k=1}^K \frac{p(\boldsymbol{x},\boldsymbol{z}_k;\theta)}{q(\boldsymbol{z}_k|\boldsymbol{x};\phi)} \right) \right].
\end{align}



\subsection{A-\MakeUppercase{\romannumeral 2}. Description of $\chi$ Upper Bound (CUBO)}


We adopt the idea introduced by \citet{dieng2017variational}, who first proposed an upper bound on the log-likelihood.  
For $u > 1$, this upper bound—known as the $\chi$ Upper Bound (CUBO)—is given by:

\begin{align*}
\frac{1}{v}\log\mathbb{E}_{\bz\sim q(\bz|\bx;\phi)}\left[ \left(\frac{p(\bx|\bz;\theta)p(\bz)}{q(\bz|\bx;\phi)}\right)^v \right].    
\end{align*}
In our paper, we set $v=2$. 
Then, for a given input $\boldsymbol{x}$, the loss function used in the last term of our objective is defined as the negative $\chi$ upper bound:
\begin{align}
\label{eq:cubo}
l_o(\theta,\phi;\boldsymbol{x})=-\frac{1}{2}\log\mathbb{E}_{\bz\sim q(\bz|\bx;\phi)}\left[ \left(\frac{p(\bx|\bz;\theta)p(\bz)}{q(\bz|\bx;\phi)}\right)^2 \right].    
\end{align}


\subsection{A-\MakeUppercase{\romannumeral 3}. Derivation of the IMBoost Loss Function}

Combining the two stable and computable loss functions in \eqref{eq:iwae} and \eqref{eq:cubo}, the final loss function of IMBoost is written as:

\begin{align*}
\hat{R}_t^{a}(\theta,\phi)=\hat{R}_t(\theta,\phi)+\lambda_{1,t}\cdot\hat{R}_{t}^i(\theta,\phi) - \lambda_{2,t}\cdot\hat{R}_{t}^o(\theta,\phi),
\end{align*}
where 
\begin{align*}
\hat{R}_t(\theta,\phi)=
\frac{\sum_{\boldsymbol{x}\in\mathcal{D}_t}l_i(\theta,\phi;\boldsymbol{x})\cdot I(l_i(\theta,\phi;\boldsymbol{x})\le\tau_t)}{\sum_{\boldsymbol{x}'\in\mathcal{D}_t}I(l_i(\theta,\phi;\boldsymbol{x}')\le\tau_t)},
\end{align*}
\begin{align*}
 \hat{R}_{t}^i(\theta,\phi)=\frac{1}{|\mathcal{I}_t|}\sum_{\boldsymbol{x}\in\mathcal{I}_t}{l_i(\theta,\phi;\boldsymbol{x})}, \quad \text{and} \quad
\hat{R}_{t}^o(\theta,\phi)=\frac{1}{|\mathcal{O}_t|}\sum_{\boldsymbol{x}\in\mathcal{O}_t}l_o(\theta,\phi;\boldsymbol{x}).
\end{align*}
It is important to note that the parameter $\phi$ is also optimized during training, since the encoder, along with the decoder, must be learned to minimize the overall loss.
At each iteration $t$, we optimize the above loss function $\hat{R}_t^{a}(\theta,\phi)$ with respect to $(\theta, \phi)$ using a stochastic gradient descent (SGD)-based optimizer such as Adam \citep{kingma2014adam}.

\section{B. Proof of Theoretical Results}
Before starting to prove Proposition 2, we first state three lemmas that are used throughout our proofs, all of which are similar to those in Appendix A of \citet{DBLP:conf/aaai/ChoHBK25}. 

\begin{lemma}
\label{lem:1}
If the IM assumption is satisfied, there exists $a_4>0$ such that  the following inequality holds:
\bea
\label{im}
\mathbb{P}_o\left(l(\theta;X)\le R_i(\theta) \right)\le a_4\cdot R_i(\theta).
\eea
\end{lemma}


\begin{lemma}[Conditional version of Theorem 3.6 in \citet{chung2006concentration}]
\label{lem:2}
Suppose that $Y_i$ for $i\in[n]$ are random variables satisfying $Y_i\le M$ and $\mathcal{H}$ is a given $\sigma$-algebra. 
Let assume that the conditional expectations $\E(Y_i|\mathcal{H})$s are independent. 
And let $Y=\sum_{i=1}^n Y_i$. 
Then, for any $\lambda>0$, we have
\begin{align*}
P\left(Y\ge \E(Y|\mathcal{H})+\lambda\Big|\mathcal{H}\right) \le \exp{\left(-\frac{\lambda^2}{2\left(\sum_{i=1}^n \E(Y_i^2|\mathcal{H})+M\lambda/3\right)}\right)} \text{ a.s.}.
\end{align*}
\end{lemma}

\begin{lemma}[Conditional and simpler version of Lemma 4 in \citet{ghadimi2016mini}]
\label{lem:3}
Suppose that $Y_i$s, $i\in[n]$, are random vectors with mean zero and $\mathcal{H}$ is a given $\sigma$-algebra. 
Let us assume that there exist positive values $\sigma_i^2>0$, $i\in[n]$, such that $E(\|Y_i\|^2)\le \sigma_i^2$. 
We also assume that the conditional expectations $\E(Y_i|\mathcal{H})$s are independent. 
Then for any $\lambda>0$, the following holds:
\begin{align*}
P\left(  \left\| \sum_{i=1}^n Y_i^2\right\|^2 \ge \lambda\sum_{i=1}^n \sigma_i^2 \Big| \mathcal{H} \right)\le \frac{1}{\lambda} \text{ a.s.}.
\end{align*}
\end{lemma}

\subsection{B-\MakeUppercase{\romannumeral 1}. Restatement and Proof of Proposition 1 from \citet{DBLP:conf/aaai/ChoHBK25}}

In this section, we restate Proposition 1 from \citet{DBLP:conf/aaai/ChoHBK25} and provide its proof using our notation.

\begin{lemma}
\label{lemma:4}
At the $t$-th update during \textit{amplification} and \textit{polarization} phases, we suppose that the current parameter $\theta_{t-1}$ satisfies $a_1\le R_i(\theta_{t-1})\le a_2\gamma^{-(t-1)}$. 
For a mini-batch $\mathcal{D}_t$, we denote the inlier set which is included in the truncated loss as $\mathcal{A}_t^{\tau}$. 
Similarly, we can define $\mathcal{B}_t^{\tau}$ for outliers. 
Then, under Assumptions 1 to 3 and for a given $\delta>0$, there exists positive constants $M_1$ and $M_2$ not depending on $t$ such that the following two inequalities holds:
\begin{align*}
|\mathcal{A}_t^{\tau}|\ge M_1 n_t\quad {and}\quad|\mathcal{B}_t^{\tau}|\le M_2 n_0,
\end{align*}
with a probability at least $1-\delta$.
\end{lemma}


\noindent
\textit{Proof)}
We prove this proposition with probability at least $1 - 4\delta$.  
Note that transforming the bound to $1 - \delta$ is straightforward by replacing $\delta$ with $\delta/4$.\\

\noindent
Prior to our analysis, we introduce several key constants that will be used throughout.
\begin{align*}
C_1 &= \sqrt{\frac{\log(2/\delta)}{2c^2(1-p_o)n_0}},\\
C_2 &= \max \left( \sqrt{\frac{\log(2/\delta)}{2(1-p_o))^2 n_0}}, \sqrt{\frac{\log(2/\delta)}{2p_o^2 n_0}} \right),\\
C_3 &= c(1-p_o)( 1-C_1)(1-C_2),\\
C_4 &= \left(7p_o(1+C_2) a_2 a_4 +\frac{1}{3}\log(1/\delta)\right),
\end{align*}
where $a_4$ is the same constant as in Lemma \ref{lem:1}. 
Here, we assume that $n_0$ is sufficiently large so that $C_1<1$ and $C_2<1$. 
For the mini-batch $\mathcal{D}_t = \{X_1, \ldots, X_{n_t}\}$ of size $n_t = n_0 \gamma^{t-1}$ sampled from $\mathcal{D}^{\text{tr}}$, we divide it into two subsets for analysis purposes:
\begin{itemize}
    \item $\mathcal{A}_t$, consisting of samples drawn from $\mathbb{P}_i$, and
    \item $\mathcal{B}_t$, consisting of samples drawn from $\mathbb{P}_o$.
\end{itemize}
We further define $\mathcal{A}_t^\tau$ and $\mathcal{B}_t^\tau$ as the subsets of $\mathcal{A}_t$ and $\mathcal{B}_t$, respectively, containing only those examples whose loss is smaller than the given threshold $\tau_t$, i.e., 

\begin{align*}
\mathcal{A}_t^\tau = \{X \in \mathcal{A}_t : l(\theta_{t-1}; X) \leq \tau_t\},
\end{align*}
\begin{align*}
\mathcal{B}_t^\tau = \{X \in \mathcal{B}_t : l(\theta_{t-1}; X) \leq \tau_t\},
\end{align*}
where $\tau_t = R_i(\theta_{t-1})$ and $\theta_{t-1}$ is the currently estimated parameter at the $(t-1)$-th update. 
Clearly, the samples used for computing $\mathbf{g}_t$ are the union of $\mathcal{A}_t^\tau$ and $\mathcal{B}_t^\tau$.
The following result bounds the size of $\mathcal{A}_t^\tau$ and $\mathcal{B}_t^\tau$. With a probability $1-4\delta$. 
We have 
\begin{align*}
    |\mathcal{A}_t^\tau| \ge C_3 n_t=C_3 n_0\gamma^{t-1} ,\ \  |\mathcal{B}_t^\tau | \le C_4 n_0,
\end{align*}
where $C_3$ and $C_4$ are defined above.

\paragraph{(i) Lower bound of $\mathcal{A}_t^{\tau}$}
By Hoeffding's inequality, for any $u>0,$ we have
\begin{align*}
\mathbb{P}(|\mathcal{A}_t|-(1-p_o)n_t\ge -u)\ge 
\mathbb{P}(||\mathcal{A}_t|-(1-p_o)n_t|\le u)\ge 1-2\exp(-2u^2/n_t).
\end{align*}
By substituting $u=\sqrt{\frac{n_t\log(2/\delta)}{2}}$ with $\delta>0$, we have
\begin{align}
\label{ineq1}
|\mathcal{A}_t|\ge (1-p_o)n_t \left(1-\sqrt{\frac{\log(2/\delta)}{2(1-p_o)^2n_t}}\right)\ge (1-p_o)(1-C_2)n_t,
\end{align}
with a probability at least $1-\delta$. 
Using the general Markov inequality and Assumption 3, we have
\begin{align*}
\mathbb{P}_i(l(\theta_{t-1} ; X) \le \tau_t) = 1- \mathbb{P}_i(l(\theta_{t-1} ; X) \ge \tau_t) \ge 1- \frac{\mathbb{E}_{i}[\sqrt{l(\theta_{t-1};X)}]}{\sqrt{R_i(\theta_{t-1})}}\ge c. 
\end{align*}
Let $\mathcal{F}_t:=\mathcal{F}\left(I(X_j\in\mathcal{X}_i),j\in[n_t]\right)$. 
By applying the conditional Hoeffding's inequality, we derive the following result: for any $u > 0$,
\begin{align*}
\mathbb{P}\Bigl(|\mathcal{A}_t^{\tau}|-c|\mathcal{A}_t|\ge -u \Big|\mathcal{F}_t\Bigr)&\ge
\mathbb{P}\Bigl(\big| |\mathcal{A}_t^{\tau}|-\E (|\mathcal{A}_t^{\tau}|) \big|\le u \Big|\mathcal{F}_t\Bigr)\\
&\ge 1-2\exp\bigl( -\frac{2u^2}{|\mathcal{A}_t|}\bigr)\text{ a.s.}
\end{align*}
With $u=\sqrt{\frac{|\mathcal{A}_t|}{2}\log(2/\delta)}$, 
\begin{align*}
\mathbb{P}\biggl(|\mathcal{A}_t^{\tau}|\ge c|\mathcal{A}_t| \biggl( 1-\sqrt{\frac{\log(2/\delta)}{2c^2 |\mathcal{A}_t|}} \biggr)  \Big|\mathcal{F}_t\biggr)\ge 1-\delta\text{ a.s.},
\end{align*}
and therefore
\begin{align}
\label{ineq3}
\mathbb{P}\biggl(|\mathcal{A}_t^{\tau}|\ge c|\mathcal{A}_t| \biggl( 1-\sqrt{\frac{\log(2/\delta)}{2c^2 |\mathcal{A}_t|}} \biggr)\biggr)\ge 1-\delta.
\end{align}
Combining (\ref{ineq1}) and (\ref{ineq3}), we have the following lower bound with a probability at least $1-2\delta$:
\begin{align}
\label{ineq9}
|\mathcal{A}_t^{\tau}|&\ge c|\mathcal{A}_t| \biggl( 1-\sqrt{\frac{\log(2/\delta)}{2c^2 |\mathcal{A}_t|}} \biggr)\nonumber\\
&\ge c(1-p_o)(1-C_2)n_t  \biggl( 1-\sqrt{\frac{\log(2/\delta)}{2c^2 (1-p_o)(1-C_2)n_t}} \biggr) \nonumber\\
&\ge c(1-p_o)(1-C_2)( 1-C_1)n_t \nonumber\\
&= C_3 n_t.
\end{align}

\paragraph{(ii) Upper bound of $\mathcal{B}_t^{\tau}$}
By Hoeffiding's inequality, we also have, for any $u>0$,
\begin{align*}
\mathbb{P}(|\mathcal{B}_t|-p_o n_t\le u)\ge 
\mathbb{P}(||\mathcal{B}_t|-p_o n_t|\le u)\ge 1-2\exp(-2t^2/n_t).
\end{align*}
By substituting $u=\sqrt{\frac{n_t\log(2/\delta)}{2}}$, we can obtain
\begin{align}
\label{ineq2}
|\mathcal{B}_t|\le p_o n_t \left( 1+\sqrt{\frac{\log(2/\delta)}{2p_o^2 n_t}} \right)\le p_o(1+C_2) n_t,
\end{align}
with a probability at least $1-\delta.$
Furthermore, by applying Lemma \ref{lem:1}, the following inequalities are obtained:
\begin{align}
\label{ineq4}
\mathbb{P}_o (l(\theta_{t-1}; X) \leq \tau_t) &= \mathbb{P}_o (l(\theta_{t-1}; X) \leq R_i(\theta_{t-1}))
\stackrel{(Lem. \ref{lem:1})}{\leq} a_4\cdot R_i(\theta_{t-1}) 
\le a_2 a_4 \gamma^{-(t-1)}.
\end{align}

\noindent
Let $\mathcal{G}_t:=\mathcal{F}\left(I(X_j\in\mathcal{X}_o),j\in[n_t]\right)$. 
We can bound the expectation of $|\mathcal{B}_t^\tau|$ given $\mathcal{G}_t$, i.e.,
\begin{align}
\label{ineq7}
\mathbb{E}\Bigl[|\mathcal{B}_t^{\tau}|\Big| \mathcal{G}_t\Bigr] &=
\mathbb{E}\left[\sum_{X\in\mathcal{B}_t}I\bigl(l(\theta_{t-1};X)\le \tau_t \bigr)\Big| \mathcal{G}_t\right]\nonumber \\
&{\le} |\mathcal{B}_t|\mathbb{P}_o\bigl( l(\theta_{t-1};X)\le\tau_t \bigr)
\stackrel{(\ref{ineq4})}{\le} |\mathcal{B}_t|a_2 a_4 \gamma^{-(t-1)} \text{ a.s.}.
\end{align}
Also, applying Lemma \ref{lem:2}, we obtain the following inequality for any $\lambda > 0$:
\begin{align*}
\mathbb{P}\left( |\mathcal{B}_t^{\tau}|\le \mathbb{E}\left[|\mathcal{B}_t^{\tau}| \big|\mathcal{G}_t\right] +\lambda\Big|\mathcal{G}_t \right)\ge 
1-\exp{\left( -\frac{\lambda^2}{2 { \mathbb{E}\Bigl[|\mathcal{B}_t^{\tau}|\Big| \mathcal{G}_t\Bigr]+\lambda/3} }\right)}
\text{ a.s.}
\end{align*}
For a given $\delta>0$, by substituting $\lambda=\frac{1}{6}\log(1/\delta)+\sqrt{\frac{1}{36}\log^2(1/\delta)+2\log(1/\delta)\mathbb{E}\bigl[|\mathcal{B}_t^{\tau}|\big| \mathcal{G}_t\bigr]}$, we have
\begin{align*}
 \mathbb{P}\left( |\mathcal{B}_t^{\tau}|\le \mathbb{E}\left[|\mathcal{B}_t^{\tau}| \big|\mathcal{G}_t\right] +\frac{1}{6}\log(1/\delta)+\sqrt{\frac{1}{36}\log^2(1/\delta)+2\log(1/\delta)\mathbb{E}\bigl[|\mathcal{B}_t^{\tau}|\big| \mathcal{G}_t\bigr]}\Big|\mathcal{G}_t \right)\ge 
1-\delta \text{ a.s.},
\end{align*}
and hence
\begin{align}
\label{ineq5}
\mathbb{P}\left( |\mathcal{B}_t^{\tau}|\le \mathbb{E}\left[|\mathcal{B}_t^{\tau}| \big|\mathcal{G}_t\right] +\frac{1}{6}\log(1/\delta)+\sqrt{\frac{1}{36}\log^2(1/\delta)+2\log(1/\delta)\mathbb{E}\bigl[|\mathcal{B}_t^{\tau}|\big| \mathcal{G}_t\bigr]} \right)\ge 
1-\delta.
\end{align}
From (\ref{ineq2}), (\ref{ineq7}), and (\ref{ineq5}), we derive the following inequality:
\begin{align}
\label{ineq8}
|\mathcal{B}_t^{\tau}| &\stackrel{(\ref{ineq5})}{\le} \mathbb{E}\left[|\mathcal{B}_t^{\tau}| \big|\mathcal{G}_t\right] +\frac{1}{6}\log(1/\delta)+\sqrt{\frac{1}{36}\log^2(1/\delta)+2\log(1/\delta)\mathbb{E}\bigl[|\mathcal{B}_t^{\tau}|\big| \mathcal{G}_t\bigr]} \nonumber\\
&\le 7\mathbb{E}\left[|\mathcal{B}_t^{\tau}| \big|\mathcal{G}_t\right]+\frac{1}{3}\log(1/\delta)\nonumber\\
&\stackrel{(\ref{ineq7})}{\le} 7|\mathcal{B}_t|a_2 a_4 \gamma^{-(t-1)}+\frac{1}{3}\log(1/\delta)\nonumber\\
&\stackrel{(\ref{ineq2})}{\le} 7p_o(1+C_2) n_0 \gamma^{t-1} a_2 a_4 \gamma^{-(t-1)}+\frac{1}{3}\log(1/\delta)\nonumber\\
&= 7p_o(1+C_2) n_0 a_2 a_4 +\frac{1}{3}\log(1/\delta)\nonumber \\
&\le \left(7p_o(1+C_2) a_2 a_4 +\frac{1}{3}\log(1/\delta)\right) n_0 =C_4 n_0.
\end{align}
with a probability at least $1-2\delta$.

\noindent
Therefore, combining (\ref{ineq9}) and (\ref{ineq8}), the proof is completed by setting $M_1=C_3$ and $M_2=C_4$. \qed

\newpage
\subsection{B-\MakeUppercase{\romannumeral 2}. Proof of Proposition 2}
\noindent
\textit{Proof)}
Similar to the proof of Lemma \ref{lemma:4}, we prove Proposition 2 with a probability of $1-7\delta$. 
Transforming $1-7\delta$ to $1-\delta$ can be done by using $\delta/7$ instead of $\delta$. \\
Before we carry out our analysis, we define a few more important constants
\begin{align*}
C_{5} &= c_2-c_2\frac{M_2}{M_1}\gamma^{-T_1}-G^2\gamma^{-T_1/2}\sqrt{\frac{2\log(2/\delta)}{M_1n_0}}-\frac{M_2}{M_1}G^2\gamma^{-T_1}(1+\sqrt{2\log(2/\delta)}),\\
C_{6,t} &= \frac{1-1/\gamma}{\mu a_2 - 2G^2(1/(\delta n_0 M_1)+M_2/M_1+(M_3/|\mathcal{I}_t|+M_4/|\mathcal{O}_t|)\gamma^{T_1})},\\
C_{7,t} &= M_3 c_1 + M_4 c_2 - G^2 \sqrt{{2\log(2/\delta)}} \sqrt{\frac{M_3^2}{|\mathcal{I}_t|}+\frac{M_4^2}{|\mathcal{O}_t|}} .
\end{align*}
Here, we assume that the \textit{amplification} phase has proceeded sufficiently so that $C_5$ is positive.
We further assume that the inlier-memorization effect is sufficiently strong, and that enough inliers and outliers are sampled to ensure that $C_{6,t}$ and $C_{7,t}$ are also positive. 

\noindent
\paragraph{(i) Decreasing property of $R_i(\theta_t)$}
During the \textit{amplification} phase, the lemma directly follows from Proposition 2 in \citet{DBLP:conf/aaai/ChoHBK25}.
Therefore, we will demonstrate that, with high probability, $R_i(\theta_{t}) \le a_2\gamma^{-t}$ for the \textit{polarization} phase, i.e., when $t > T_1$. 
For the convenience of theoretical analysis, we use the \textit{normalized} loss function with respect to a constant $(1+\lambda_{1,t}+\lambda_{2,t})$ as follows: 
\begin{align*}
    \hat{R}_t(\theta)=\frac{1}{1+\lambda_{1,t}+\lambda_{2,t}}\left[\hat{R}_t(\theta)+\lambda_{1,t} \hat{R}_{t}^i(\theta)-\lambda_{2,t} \hat{R}_{t}^o(\theta)\right],
\end{align*}
where
\begin{align*}    \hat{R}_t(\theta;\tau_t)&=\frac{\sum_{\boldsymbol{x}\in\mathcal{D}_t}l(\theta;\boldsymbol{x})\cdot I(l(\theta;\boldsymbol{x})\le\tau_t)}{\sum_{\boldsymbol{x}'\in\mathcal{D}_t}I(l(\theta;\boldsymbol{x}')\le\tau_t)},\\
    \hat{R}_{t}^i&=\frac{1}{|\mathcal{I}_{t}|}\sum_{\boldsymbol{x}\in \mathcal{I}_{t}}l(\theta;\boldsymbol{x}), \\
    \hat{R}_{t}^o&=\frac{1}{|\mathcal{O}_{t}|}\sum_{\boldsymbol{x}\in \mathcal{O}_{t}}l(\theta;\boldsymbol{x}),
\end{align*}
and $\lambda_{1,t}=M_3\cdot\gamma^{-(t-T_1-1)},\lambda_{2,t}=M_4\cdot\gamma^{-(t-T_1-1)}$ with $M_3,M_4>0$. 
It is straightforward to note that the update with the original (unnormalized) loss function can be recovered by scaling the learning rate by a factor of $(1 + \lambda_{1,t}+\lambda_{2,t})$.

Let $\mathbf{g}_t=\nabla_{\theta}\hat{R}(\theta_{t-1})$. 
Using the notation of $\mathcal{A}_t^\tau$ and $\mathcal{B}_t^\tau$, we can rewrite $\mathbf{g}_t$ as 
\begin{align*}
    \mathbf{g}_t = \frac{1}{1+\lambda_{1,t}+\lambda_{2,t}}\biggl[(1-b_t)\mathbf{g}_t^a + {b_t}\mathbf{g}_t^b+\lambda_{1,t}\mathbf{g}_t^i-\lambda_{2,t}\mathbf{g}_t^o\biggr], 
\end{align*}
where $\mathbf{g}_t^a = \frac{1}{|\mathcal{A}_t^\tau|}\sum_{X\in\mathcal{A}_t^\tau}\nabla l(\theta_{t-1};X)$,\ \ $\mathbf{g}_t^b = \frac{1}{|\mathcal{B}_t^\tau|}\sum_{X\in\mathcal{B}_t^\tau}\nabla l(\theta_{t-1};X)$, $\mathbf{g}_t^i = \frac{1}{|\mathcal{I}_{t}|}\sum_{X\in\mathcal{I}_{t}}\nabla l(\theta_{t-1};X)$, $\mathbf{g}_t^o = \frac{1}{|\mathcal{O}_{t}|}\sum_{X\in\mathcal{O}_{t}}\nabla l(\theta_{t-1};X)$, and $b_t$ is the proportion of samples from $\mathcal{B}_t^\tau$, that is, $b_t = {|\mathcal{B}_t^\tau|}/{(|\mathcal{A}_t^\tau|+|\mathcal{B}_t^\tau|})$.
Note that 
\begin{align}
\label{ineq22}
    b_t = \frac{|\mathcal{B}_t^\tau|}{|\mathcal{A}_t^\tau|+|\mathcal{B}_t^\tau|} \le \frac{|\mathcal{B}_t^\tau|}{|\mathcal{A}_t^\tau|} \le \frac{M_2}{1+M_1\gamma^{t-1}}<\frac{M_2}{M_1}\gamma^{-(t-1)},
\end{align}
with a probability of at least $1-4\delta$ by Lemma \ref{lemma:4}.

Since $R_i(\theta)$ is $L$-smooth by Assumption 2-(2), we have 
\begin{align}
\label{ineq12}
R_i(\theta_{t}) - R_i(\theta_{t-1}) &\overset{(a)}{\leq} \langle \nabla R_i(\theta_{t-1}), \theta_{t} - \theta_{t-1} \rangle + \frac{L}{2} \|\theta_{t} - \theta_{t-1}\|^2 \nonumber\\
&\overset{(b)}{=} \frac{\eta_t}{2} \|\nabla R_i(\theta_{t-1}) - \mathbf{g}_t \|^2 - \frac{\eta_t}{2} \left( \|\nabla R_i(\theta_{t-1})\|^2 + (1 - \eta_t L) \|\mathbf{g}_t\|^2 \right)\nonumber \\
&\overset{(c)}{\leq} \frac{\eta_t}{2(1+\lambda_{1,t}+\lambda_{2,t})} \biggl[ {(1 - b_t)} \|\nabla R_i(\theta_{t-1}) - \mathbf{g}_t^a \|^2 + {b_t} \|\nabla R_i(\theta_{t-1}) - \mathbf{g}_t^b \|^2 +{\lambda_{1,t}} \|\nabla R_i(\theta_{t-1}) - \mathbf{g}_t^i \|^2 \nonumber \\
&+{\lambda_{2,t}} \|\nabla R_i(\theta_{t-1}) + \mathbf{g}_t^o \|^2 \biggr]- \frac{\eta_t}{2} \left( \|\nabla R_i(\theta_{t-1})\|^2 + (1 - \eta_t L) \|\mathbf{g}_t\|^2 \right) \nonumber\\
&\overset{(d)}{\leq} \frac{\eta_t}{2(1+\lambda_{1,t}+\lambda_{2,t})} \biggl[ {(1 - b_t)} \|\nabla R_i(\theta_{t-1}) - \mathbf{g}_t^a \|^2 + {4G^2\left(b_t+\frac{\lambda_{1,t}}{|\mathcal{I}_t|}+\frac{\lambda_{2,t}}{|\mathcal{O}_t|}\right)} \biggr] - \eta_t \mu R_i(\theta_{t-1})\nonumber\\
&\overset{(d)}{\leq} \frac{\eta_t}{2(1+\lambda_{1,t}+\lambda_{2,t})} \biggl[ \|\nabla R_i(\theta_{t-1}) - \mathbf{g}_t^a \|^2 + {4G^2\left(b_t+\frac{\lambda_{1,t}}{|\mathcal{I}_t|}+\frac{\lambda_{2,t}}{|\mathcal{O}_t|}\right)} \biggr] - \eta_t \mu R_i(\theta_{t-1}),
\end{align}
where $(a)$ is due to Assumption 2-(1); 
$(b)$ follows the update of $\theta_{t} = \theta_{t-1} - \eta_t\mathbf{g}_{t}$; 
$(c)$ is due to the definition of $\mathbf{g}_t$ and the convexity of $\| \cdot \|^2$; 
$(d)$ follows the Assumption 2-(1),(2), and $\eta_t L \le 1$. 

Let $\mathcal{F}_t^{\tau}:=\mathcal{F}\left(I(X_j\in\mathcal{A}_t^{\tau}),j\in[n_t]\right)$. 
Then we have 
\begin{align*}
\mathbb{E} \left( \left\| \mathbf{g}_t^a - \nabla R_i(\theta_{t-1}) \right\|^2 \big| \mathcal{F}_t^{\tau}\right) &= 
\frac{1}{|\mathcal{A}_t^{\tau}|^2} \mathbb{E} \left(\Bigl\|  \sum_{X \in \mathcal{A}_t^{\tau}} \left( \nabla l(\theta_{t-1}; X) - \nabla R_i(\theta_{t-1}) \right) \Bigr\|^2 \bigg| \mathcal{F}_t^{\tau}\right) \\
&= \frac{1}{|\mathcal{A}_t^{\tau}|^2}  \mathbb{E} \left(\sum_{X \in \mathcal{A}_t^{\tau}}\Bigl\|  \nabla l(\theta_{t-1}; X) - \nabla R_i(\theta_{t-1}) \Bigr\|^2 \bigg| \mathcal{F}_t^{\tau}\right) \\
&\leq \frac{4G^2}{|\mathcal{A}_t^{\tau}|}, 
\end{align*}
where the last inequality holds due to Assumption 2-(1).
For a given $\delta>0$, by using Lemma \ref{lem:3} with $\lambda=1/\delta$, we have 
\begin{align*}
\mathbb{P}\left( \mathbb{P} \biggl(\left\| \mathbf{g}_t^a - \nabla R_i(\theta_{t-1}) \right\|^2  \le \frac{4G^2}{\delta|\mathcal{A}_t^{\tau}|}  \bigg| \mathcal{F}_t^{\tau} \biggr)\ge 1-\delta \right)=1,
\end{align*}
and hence
\begin{align}
\label{ineq10}
 \mathbb{P} \biggl(\left\| \mathbf{g}_t^a - \nabla R_i(\theta_{t-1}) \right\|^2  \le \frac{4G^2}{\delta|\mathcal{A}_t^{\tau}|}   \biggr)\ge 1-\delta. 
\end{align} 

Using the above bounds in \eqref{ineq22} and \eqref{ineq10}, we can further expand the bound in \eqref{ineq12} as follows:
\begin{align*}
R_i(\theta_{t}) - R_i(\theta_{t-1}) &\leq \frac{\eta_t}{2(1+\lambda_{1,t}+\lambda_{2,t})} 
\biggl[\frac{4G^2}{\delta M_1 n_0}\gamma^{-(t-1)} + 4G^2\gamma^{-(t-1)}\left(\frac{M_2}{M_1}+\left(\frac{M_3}{|\mathcal{I}_t|}+\frac{M_4}{|\mathcal{O}_t|}\right)\gamma^{T_1}\right) \biggr] - \eta_t \mu R_i(\theta_{t-1}) \\
&\leq \frac{\eta_t}{2} \biggl[ \frac{4G^2}{\delta M_1 n_0}\gamma^{-(t-1)} + 4G^2 \gamma^{-(t-1)} \left(\frac{M_2}{M_1}+\left(\frac{M_3}{|\mathcal{I}_t|}+\frac{M_4}{|\mathcal{O}_t|}\right)\gamma^{T_1}\right)\biggr] - \eta_t \mu R_i(\theta_{t-1}) \\
&= {2 \eta_t G^2} \left( \frac{1}{\delta n_0 M_1} + \frac{M_2}{M_1}+\left(\frac{M_3}{|\mathcal{I}_t|}+\frac{M_4}{|\mathcal{O}_t|}\right)\gamma^{T_1} \right) \gamma^{-(t-1)} - \eta_t \mu R_i(\theta_{t-1}).
\end{align*}
Hence, we have 
\begin{align*}
    R_i(\theta_{t}) &\le (1-\eta_t\mu)R_i(\theta_{t-1})+{2 \eta_t G^2} \left( \frac{1}{\delta n_0 M_1} + \frac{M_2}{M_1}+\left(\frac{M_3}{|\mathcal{I}_t|}+\frac{M_4}{|\mathcal{O}_t|}\right)\gamma^{T_1} \right) \gamma^{-(t-1)} \\ 
    &\le \gamma\left[ (1-\eta_t\mu)a_2 + {2 \eta_t G^2} \left( \frac{1}{\delta n_0 M_1} + \frac{M_2}{M_1}+\left(\frac{M_3}{|\mathcal{I}_t|}+\frac{M_4}{|\mathcal{O}_t|}\right)\gamma^{T_1} \right)\right] \gamma^{-t},
\end{align*} 
with a probability of at least $1-5\delta$.
Let us select a learning rate $\eta_t$ satisfying $\eta_t\ge C_{6,t}$. 
Then, we have $R_i(\theta_t)\le a_2 \gamma^{-t}$,
and the proof is completed. \qed

\noindent
\paragraph{(ii) Increasing property of $R_o(\theta_t)$}
By Assumption 2-(1) and 2-(2), we have 
\begin{align}
\label{ineq16}
R_o(\theta_{t}) - R_o(\theta_{t-1}) &{\ge} \langle \nabla R_i(\theta_{t-1}), \theta_{t} - \theta_{t-1} \rangle - \frac{L}{2} \|\theta_{t} - \theta_{t-1}\|^2 \nonumber\\
&{=} -{\eta_t} \langle \nabla R_o(\theta_{t-1}), \mathbf{g}_t \rangle  - \frac{\eta_t^2 L}{2} \|\mathbf{g}_t\|^2 \nonumber \\
&\ge \eta_t \left(-  \langle \nabla R_o(\theta_{t-1}), \mathbf{g}_t \rangle - \frac{\eta_t L G^2}{2}\right).
\end{align}
Thus, it is sufficient to show that the term next to $\eta_t$ in \eqref{ineq16} is larger than zero. 
The term $-  \langle \nabla R_o(\theta_{t-1}), \mathbf{g}_t \rangle$ can be reformulated as:
\begin{align}
\label{ineq21}
    -\langle \nabla R_o(\theta_{t-1}), \mathbf{g}_t \rangle &= \frac{1}{1+\lambda_{1,t}+\lambda_{2,t}}\biggl[ -(1-b_t)\langle \nabla R_o(\theta_{t-1}),\mathbf{g}_t^a\rangle - b_t\langle \nabla R_o(\theta_{t-1}), \mathbf{g}_t^b \rangle \nonumber \\
    &-\lambda_{1,t} \langle \nabla R_o(\theta_{t-1}), \mathbf{g}_t^i\rangle + 
    \lambda_{2,t} \langle \nabla R_o(\theta_{t-1}), \mathbf{g}_t^o\rangle\biggr]
\end{align}
By Conditional Hoeffding's inequality, for any $u>0$, we can achieve
\begin{align*}
\mathbb{P}\left( \langle \nabla R_o(\theta_{t-1}),\mathbf{g}_t^a\rangle-
\langle \nabla R_o(\theta_{t-1}),R_i(\theta_{t-1})\rangle\le u \Big|\mathcal{F}_t^{\tau}\right)&=
\mathbb{P}\left( \sum_{{X}\in\mathcal{A}_t^{\tau}}\langle \nabla R_o(\theta_{t-1}),l(\theta_{t-1};{X})-R_i(\theta_{t-1})\rangle\le |\mathcal{A}_t^{\tau}|\cdot u \Big|\mathcal{F}_t^{\tau}\right)
\nonumber\\
&\ge 
\mathbb{P}\left( \left|\sum_{{X}\in\mathcal{A}_t^{\tau}}\langle \nabla R_o(\theta_{t-1}),l(\theta_{t-1};{X})-R_i(\theta_{t-1})\rangle \right|\le |\mathcal{A}_t^{\tau}|\cdot u \Big|\mathcal{F}_t^{\tau}\right)
\nonumber \\
&\ge 
1-2\exp\left(\frac{u^2 \cdot |\mathcal{A}_t^{\tau}|}{2G^4}\right) \quad \text{a.s.},
\end{align*}
where $\mathcal{F}_t^{\tau}:=\mathcal{F}\left(I(X_j\in\mathcal{A}_t^{\tau}),j\in[n_t]\right)$.
By substituting $u={G^2}\sqrt{\frac{2\log(2/\delta)}{|\mathcal{A}_t^{\tau}|}}$ and by using the upper bound of $|\mathcal{A}_t^{\tau}|$ and Assumption 4, 
\begin{align}
\label{ineq17}
-\langle \nabla R_o(\theta_{t-1}), \mathbf{g}_t^a \rangle &\ge 
-\langle \nabla R_o(\theta_{t-1}), \nabla R_i(\theta_{t-1})\rangle - {G^2}\sqrt{\frac{2\log(2/\delta)}{|\mathcal{A}_t^{\tau}|}} \nonumber \\
&\ge c_2 - {G^2}\sqrt{\frac{2\log(2/\delta)}{|\mathcal{A}_t^{\tau}|}} 
\end{align}
with a probability larger than $1-\delta$. 
Similarly, using the conditional Hoeffding's inequality, for any $u>0$, we have
\begin{align*}
\mathbb{P}\left( \langle \nabla R_o(\theta_{t-1}),\mathbf{g}_t^b\rangle-
\langle \nabla R_o(\theta_{t-1}),\nabla R_o(\theta_{t-1})\rangle\le u \Big|\mathcal{G}_t^{\tau}\right)& \ge
\mathbb{P}\left( \left|\sum_{{X}\in\mathcal{B}_t^{\tau}}\langle \nabla R_o(\theta_{t-1}),l(\theta_{t-1};{X})-R_o(\theta_{t-1})\rangle \right|\le |\mathcal{B}_t^{\tau}|\cdot u \Big|\mathcal{G}_t^{\tau}\right)
\nonumber \\
&\ge 1-2\exp\left(\frac{u^2\cdot |\mathcal{B}_t^{\tau}|}{2G^4}\right) \quad\text{a.s.},
\end{align*}
where $\mathcal{G}_t^{\tau}:=\mathcal{F}\left(I(X_j\in\mathcal{B}_t^{\tau}),j\in[n_t]\right)$.
And by substituting $u={G^2}\sqrt{\frac{2\log(2/\delta)}{|\mathcal{B}_t^{\tau}|}}$  and using Assumption 2-(1), the inequality below holds with a probability larger than $1-\delta$: 
\begin{align}
\label{ineq18}
-\langle \nabla R_o(\theta_{t-1}), \mathbf{g}_t^b \rangle &\ge 
-\left\| \nabla R_o(\theta_{t-1})\right\|^2 - {G^2}\sqrt{\frac{2\log(2/\delta)}{|\mathcal{B}_t^{\tau}|}}
 \nonumber \\& \ge -G^2 - {G^2}\sqrt{\frac{2\log(2/\delta)}{|\mathcal{B}_t^{\tau}|}}. 
\end{align}

Regarding $\mathbf{g}_t^i$ and $\mathbf{g}_t^o$, we apply Hoeffding's inequality with Assumption 2-(1), thereby obtaining, for any $u>0$,  
\begin{align*}
&\mathbb{P}\left(\left(\lambda_{1,t}\langle \nabla R_o(\theta_{t-1}),\mathbf{g}_t^i \rangle-\lambda_{2,t}\langle \nabla R_o(\theta_{t-1}),\mathbf{g}_t^b \rangle \right)
-\left(\lambda_{1,t}\langle \nabla R_o(\theta_{t-1}),\nabla R_o(\theta_{t-1}) \rangle
-\lambda_{2,t}\langle \nabla R_o(\theta_{t-1}),\nabla R_i(\theta_{t-1}) \rangle \right)\ge u\right)\nonumber\\
&\ge \mathbb{P}\left(\left|\left(\lambda_{1,t}\langle \nabla R_o(\theta_{t-1}),\mathbf{g}_t^i \rangle-\lambda_{2,t}\langle \nabla R_o(\theta_{t-1}),\mathbf{g}_t^b \rangle \right)
-\left(\lambda_{1,t}\langle \nabla R_o(\theta_{t-1}),\nabla R_o(\theta_{t-1}) \rangle
-\lambda_{2,t}\langle \nabla R_o(\theta_{t-1}),\nabla R_i(\theta_{t-1}) \rangle \right)\right|\le u\right)\nonumber\\
&\ge 1-2\exp \left(-\frac{u^2}{2G^4 \left( \lambda_{1,t}^2/|\mathcal{I}_t|+\lambda_{2,t}/|\mathcal{O}_t| \right)} \right)
\end{align*}
By substituting $u=G^2\sqrt{2\log(2/\delta)}\sqrt{\frac{\lambda_{1,t}^2}{|\mathcal{I}_t|}+\frac{\lambda_{2,t}^2}{|\mathcal{O}_t|}}$, along with Assumption 4, we have
\begin{align}
\label{ineq19}
-\lambda_{1,t}\langle \nabla R_o(\theta_{t-1}),\mathbf{g}_t^i \rangle + \lambda_{2,t}\langle \nabla R_o(\theta_{t-1}),\mathbf{g}_t^b \rangle \ge 
\lambda_{1,t}c_1+\lambda_{2,t}c_2-G^2\sqrt{2\log(2/\delta)}\sqrt{\frac{\lambda_{1,t}^2}{|\mathcal{I}_t|}+\frac{\lambda_{2,t}^2}{|\mathcal{O}_t|}},
\end{align}
with a probability at least $1-\delta$. 



Combining \eqref{ineq21}-\eqref{ineq19} and by Lemma \ref{lemma:4}, we conclude that the following holds with probability at least $1-7\delta$:

\begin{align}
\label{ineq26}
    -\langle \nabla R_o(\theta_{t-1}), \mathbf{g}_t \rangle &= \frac{1}{1+\lambda_{1,t}+\lambda_{2,t}}\biggl[ -(1-b_t)\langle \nabla R_o(\theta_{t-1}),\mathbf{g}_t^a\rangle - b_t\langle \nabla R_o(\theta_{t-1}), \mathbf{g}_t^b \rangle \nonumber \\
    &-\lambda_{1,t} \langle \nabla R_o(\theta_{t-1}), \mathbf{g}_t^i\rangle + 
    \lambda_{2,t} \langle \nabla R_o(\theta_{t-1}), \mathbf{g}_t^o\rangle\biggr]\nonumber\\
    &\ge \frac{1}{1+\lambda_{1,t}+\lambda_{2,t}}\biggl[ 
    (1-b_t) \left( c_2 - G^2\sqrt{\frac{2\log(2/\delta)}{|\mathcal{A}_t^{\tau}|}}\right)
    -b_t\left( G^2+G^2 \sqrt{\frac{2\log(2/\delta)}{|\mathcal{B}_t^{\tau}|}}\right)\nonumber\\
    &+ \lambda_{1,t}c_1+\lambda_{2,t}c_2-G^2\sqrt{2\log(2/\delta)}\sqrt{\frac{\lambda_{1,t}^2}{|\mathcal{I}_t|}+\frac{\lambda_{2,t}^2}{|\mathcal{O}_t|}} \biggr]\nonumber\\
    &\ge \frac{1}{1+\lambda_{1,t}+\lambda_{2,t}}\biggl[
    (1-b_t)\left( c_2 - G^2\sqrt{\frac{2\log(2/\delta)}{M_1 n_0 \gamma^{t-1}}} \right) 
    -b_t\left( G^2+G^2 \sqrt{{2\log(2/\delta)}}\right) \nonumber\\
    &+ \lambda_{1,t}c_1+\lambda_{2,t}c_2-G^2\sqrt{2\log(2/\delta)}\sqrt{\frac{\lambda_{1,t}^2}{|\mathcal{I}_t|}+\frac{\lambda_{2,t}^2}{|\mathcal{O}_t|}} \biggr]\nonumber\\
    &\ge \frac{1}{1+M_3+M_4}\biggl[ \left( 1-\frac{M_2}{M_1}\gamma^{-T_1} \right)\left( c_2 - G^2\sqrt{\frac{2\log(2/\delta)}{M_1 n_0 \gamma^{T_1}}} \right) 
    - \frac{M_2}{M_1}G^2\gamma^{-T_1}\left( 1+\sqrt{2\log(2/\delta)} \right) \nonumber\\
    &+\gamma^{-(t-T_1-1)}\left(M_3 c_1 + M_4 c_2 - G^2 \sqrt{{2\log(2/\delta)}} \sqrt{\frac{M_3^2}{|\mathcal{I}_t|}+\frac{M_4^2}{|\mathcal{O}_t|}} \right)
    \biggr] \nonumber\\
    &= \frac{1}{1+M_3+M_4}\biggl[ \biggl( c_2-c_2\frac{M_2}{M_1}\gamma^{-T_1}-G^2\gamma^{-T_1/2}\sqrt{\frac{2\log(2/\delta)}{M_1n_0}}-\frac{M_2}{M_1}G^2\gamma^{-T_1}(1+\sqrt{2\log(2/\delta)})\biggr)\nonumber\\
    &+\gamma^{-(t-T_1-1)}\left(M_3 c_1 + M_4 c_2 - G^2 \sqrt{{2\log(2/\delta)}} \sqrt{\frac{M_3^2}{|\mathcal{I}_t|}+\frac{M_4^2}{|\mathcal{O}_t|}} \right)
    \biggr] \nonumber\\
    &=\frac{1}{1+M_3+M_4}\left( C_5 + C_{7,t}\gamma^{-(t-T_1-1)}\right).
\end{align}
Finally, combining \eqref{ineq16} with \eqref{ineq26}, we have 
\begin{align*}
    R_o(\theta_{t}) - R_o(\theta_{t-1}) &{\ge} \eta_t \left(-  \langle \nabla R_o(\theta_{t-1}), \mathbf{g}_t \rangle - \frac{\eta_t L G^2}{2}\right)\nonumber\\
    &\ge \eta_{t} \left( \frac{1}{1+M_3+M_4}\left( C_5 + C_{7,t}\gamma^{-(t-T_1-1)}\right)- \frac{\eta_t L G^2}{2}\right).
\end{align*}
Therefore, if we set the learning rate $\eta$ such that 
\begin{align*}
    \eta_t<\frac{2(C_5+C_{7,t})}{LG^2(1+M_3+M_4)},
\end{align*}
the inequality $R_o(\theta_{t}) - R_o(\theta_{t-1})>0$ holds, which is the completion of the proof. \qed


\newpage

\subsection{B-\MakeUppercase{\romannumeral 3}. Theoretical Insights into Proposition 2}

In this section, we discuss further implications of our theoretical results.

\paragraph{1. Why the ALTBI Loss Function is Necessary in the Warm-up Phase}
We observe from \eqref{ineq26} that, after the $t$-th iteration, the contribution of the trimmed loss to the increase in outlier risk is quantified by the constant $C_5$, which is given by:
\begin{align*}
c_2-c_2\frac{M_2}{M_1}\gamma^{-t}-G^2\gamma^{-t/2}\sqrt{\frac{2\log(2/\delta)}{M_1n_0}}-\frac{M_2}{M_1}G^2\gamma^{-t}(1+\sqrt{2\log(2/\delta)}).    
\end{align*}
This value can become negative when $t$ is small, indicating that the trimmed loss may adversely affect the increase of outlier risk in the early stages of training.
However, as training progresses, the proportion of outliers in the trimmed loss decreases exponentially, as shown in Lemma \ref{lemma:4}, leading to a gradual increase in this value.
Consequently, after a sufficient number of iterations, the value becomes positive, thereby contributing to the increase in outlier risk.
This explains why it is necessary to use the trimmed loss during the warm-up phase prior to introducing active learning.



\paragraph{2. The Optimal Ratio of Labeled Inlier and Outlier Samples}
We introduce a simple lemma below:

\begin{lemma}
\label{lemma:5}
Consider a function $f(a):(0,n)\to \mathbb{R}_{+}$ which is defined as:
\begin{align*}
f(a)= \frac{u_1^2}{a}+\frac{u_2^2}{n-a},
\end{align*}
where $u_1,u_2$ are two positive constants. 
Then, the function $f(a)$ is minimized when $a=\frac{u_1}{u_1+u_2} n$.
\end{lemma}
\textit{proof)}
We have 
\begin{align*}
f'(a)&=-\frac{u_1^2}{a^2}+\frac{u_2^2}{(n-a)^2}=\frac{-u_1^2(n-a)^2+u_2^2a^2}{a^2(n-a)^2}\\
&=\frac{\left((u_1+u_2)a-u_1n\right)\left((u_2-u_1)a+u_1n\right)}{a^2(n-a)^2}.
\end{align*}

\noindent
1) If $u_1=u_2$, it is obvious that $f(a)$ is minimized when $a=\frac{u_1}{u_1+u_2} n$.

\noindent
2) If $u_1 \neq u_2$, the solutions to $f'(a) = 0$ are $a = \frac{u_1}{u_1 + u_2}n$ and $a = \frac{u_1}{u_1 - u_2}$.

\noindent
\quad - If $u_1 > u_2$, note that $0 < \frac{u_1}{u_1 + u_2}n < n$ and $\frac{u_1}{u_1 - u_2} > n$. Hence, the valid solution that minimizes $f(a)$ is $a = \frac{u_1}{u_1 + u_2}  n$.

\noindent
\quad - If $u_1 < u_2$, we have $0 < \frac{u_1}{u_1 + u_2}n < n$ and $\frac{u_1}{u_1 - u_2} < 0$. Again, the valid minimizer is $a = \frac{u_1}{u_1 + u_2}  n$. \qed





\noindent
We can see from \eqref{ineq26} that the contribution of active learning to the increase in outlier risk is represented by the term $C_{7,t}\gamma^{-(t - T_1 - 1)}$, where $C_{7,t}$ is defined as:
\begin{align*}
\gamma^{-(t-T_1-1)}\left(M_3 c_1 + M_4 c_2 - G^2 \sqrt{{2\log(2/\delta)}} \sqrt{\frac{M_3^2}{|\mathcal{I}_t|}+\frac{M_4^2}{|\mathcal{O}_t|}} \right)
    \biggr].    
\end{align*}
When all other variables are fixed except for $|\mathcal{I}_t|$ and $|\mathcal{O}_t|$, the value is maximized when the term $\frac{M_3^2}{|\mathcal{I}_t|} + \frac{M_4^2}{|\mathcal{O}_t|}$ is minimized.
Given the constraint $|\mathcal{I}_t| + |\mathcal{O}_t| = n_a$, Lemma \ref{lemma:5} shows that the minimum is achieved when $\mathcal{I}_t=\frac{M_3}{M_3+M_4}n_a$ and $\mathcal{O}_t=\frac{M_4}{M_3+M_4}n_a$. 
Therefore, when the hyperparameters $\lambda_{1,t}$ and $\lambda_{2,t}$ are set to similar values (implying $M_3 \approx M_4$), the outlier risk increases the most when the numbers of inlier and outlier samples are approximately balanced, i.e., $|\mathcal{I}_t| \approx |\mathcal{O}_t| \approx \frac{n_a}{2}$.


\newpage

\section{C. Details of Experimental Analysis}

\subsection{C-\MakeUppercase{\romannumeral 1}. Benchmark Dataset Description}
We evaluate a total of 46 tabular datasets, and 6 image datasets, and 5 text datasets. 
These datasets are all obtained from a source known as ADBench. \citep{han2022adbench}. 
Table \ref{tab:adbench_summ} provides a summary of the basic imformation for all datasets we analyze.

\begin{table}[h!]
\renewcommand\thetable{C.1}
\centering
\setlength{\tabcolsep}{1mm}
\fontsize{8pt}{8pt}\selectfont
\begin{tabular}{llccccl}
\toprule
\textbf{Number} & \textbf{Dataset Name} & \textbf{\#Samples} & \textbf{\#Features} & \textbf{\#Anomaly} & \textbf{\%Anomaly} & \textbf{Category} \\
\midrule
\textbf{1}      & ALOI                  & 49534              & 27                  & 1508               & 3.04               & Image             \\
\textbf{2}      & annthyroid            & 7200               & 6                   & 534                & 7.42               & Healthcare        \\
\textbf{3}      & backdoor              & 95329              & 196                 & 2329               & 2.44               & Network           \\
\textbf{4}      & breastw               & 683                & 9                   & 239                & 34.99              & Healthcare        \\
\textbf{5}      & campaign              & 41188              & 62                  & 4640               & 11.27              & Finance           \\
\textbf{6}      & cardio                & 1831               & 21                  & 176                & 9.61               & Healthcare        \\
\textbf{7}      & Cardiotocography      & 2114               & 21                  & 466                & 22.04              & Healthcare        \\
\textbf{8}      & celeba                & 202599             & 39                  & 4547               & 2.24               & Image             \\
\textbf{9}      & census                & 299285             & 500                 & 18568              & 6.2                & Sociology         \\
\textbf{10}     & cover                 & 286048             & 10                  & 2747               & 0.96               & Botany            \\
\textbf{11}     & donors                & 619326             & 10                  & 36710              & 5.93               & Sociology         \\
\textbf{12}     & fault                 & 1941               & 27                  & 673                & 34.67              & Physical          \\
\textbf{13}     & fraud                 & 284807             & 29                  & 492                & 0.17               & Finance           \\
\textbf{14}     & glass                 & 214                & 7                   & 9                  & 4.21               & Forensic          \\
\textbf{15}     & Hepatitis             & 80                 & 19                  & 13                 & 16.25              & Healthcare        \\
\textbf{16}     & http                  & 567498             & 3                   & 2211               & 0.39               & Web               \\
\textbf{17}     & InternetAds           & 1966               & 1555                & 368                & 18.72              & Image             \\
\textbf{18}     & Ionosphere            & 351                & 32                  & 126                & 35.9               & Oryctognosy       \\
\textbf{19}     & landsat               & 6435               & 36                  & 1333               & 20.71              & Astronautics      \\
\textbf{20}     & letter                & 1600               & 32                  & 100                & 6.25               & Image             \\
\textbf{21}     & Lymphography          & 148                & 18                  & 6                  & 4.05               & Healthcare        \\
\textbf{22}     & magic.gamma           & 19020              & 10                  & 6688               & 35.16              & Physical          \\
\textbf{23}     & mammography           & 11183              & 6                   & 260                & 2.32               & Healthcare        \\
\textbf{24}     & mnist                 & 7603               & 100                 & 700                & 9.21               & Image             \\
\textbf{25}     & musk                  & 3062               & 166                 & 97                 & 3.17               & Chemistry         \\
\textbf{26}     & optdigits             & 5216               & 64                  & 150                & 2.88               & Image             \\
\textbf{27}     & PageBlocks            & 5393               & 10                  & 510                & 9.46               & Document          \\
\textbf{28}     & pendigits             & 6870               & 16                  & 156                & 2.27               & Image             \\
\textbf{29}     & Pima                  & 768                & 8                   & 268                & 34.9               & Healthcare        \\
\textbf{30}     & satellite             & 6435               & 36                  & 2036               & 31.64              & Astronautics      \\
\textbf{31}     & satimage-2            & 5803               & 36                  & 71                 & 1.22               & Astronautics      \\
\textbf{32}     & shuttle               & 49097              & 9                   & 3511               & 7.15               & Astronautics      \\
\textbf{33}     & skin                  & 245057             & 3                   & 50859              & 20.75              & Image             \\
\textbf{34}     & smtp                  & 95156              & 3                   & 30                 & 0.03               & Web               \\
\textbf{35}     & SpamBase              & 4207               & 57                  & 1679               & 39.91              & Document          \\
\textbf{36}     & speech                & 3686               & 400                 & 61                 & 1.65               & Linguistics       \\
\textbf{37}     & Stamps                & 340                & 9                   & 31                 & 9.12               & Document          \\
\textbf{38}     & thyroid               & 3772               & 6                   & 93                 & 2.47               & Healthcare        \\
\textbf{39}     & vertebral             & 240                & 6                   & 30                 & 12.5               & Biology           \\
\textbf{40}     & vowels                & 1456               & 12                  & 50                 & 3.43               & Linguistics       \\
\textbf{41}     & Waveform              & 3443               & 21                  & 100                & 2.9                & Physics           \\
\textbf{42}     & WBC                   & 223                & 9                   & 10                 & 4.48               & Healthcare        \\
\textbf{43}     & WDBC                  & 367                & 30                  & 10                 & 2.72               & Healthcare        \\
\textbf{44}     & Wilt                  & 4819               & 5                   & 257                & 5.33               & Botany            \\
\textbf{45}     & wine                  & 129                & 13                  & 10                 & 7.75               & Chemistry         \\
\textbf{46}     & WPBC                  & 198                & 33                  & 47                 & 23.74              & Healthcare        \\
\textbf{47}     & yeast                 & 1484               & 8                   & 507                & 34.16              & Biology           \\
\textbf{48}     & CIFAR10               & 5263               & 512                 & 263                & 5                  & Image             \\
\textbf{49}     & FashionMNIST          & 6315               & 512                 & 315                & 5                  & Image             \\
\textbf{50}     & MNIST-C               & 10000              & 512                 & 500                & 5                  & Image             \\
\textbf{51}     & MVTec-AD              & 5354               & 512                 & 1258               & 23.5               & Image             \\
\textbf{52}     & SVHN                  & 5208               & 512                 & 260                & 5                  & Image             \\
\textbf{53}     & Agnews                & 10000              & 768                 & 500                & 5                  & NLP               \\
\textbf{54}     & Amazon                & 10000              & 768                 & 500                & 5                  & NLP               \\
\textbf{55}     & Imdb                  & 10000              & 768                 & 500                & 5                  & NLP               \\
\textbf{56}     & Yelp                  & 10000              & 768                 & 500                & 5                  & NLP               \\
\textbf{57}     & 20news        & 11905              & 768                 & 591                & 4.96               & NLP              \\
\bottomrule
\end{tabular}
\caption{Description of \texttt{ADBench} datasets}
\label{tab:adbench_summ}
\end{table}

\subsection{C-\MakeUppercase{\romannumeral 2}. Detailed AUC and AP Results over ADBench Datasets}
Table \ref{tab:adbench1}- \ref{tab:adbench4} provide detailed results of averaged AUC and AP along with their standard deviations for each method over the \texttt{ADBench} datasets, including both training and test evaluations.

\begin{sidewaystable}[h!]
\renewcommand\thetable{C.2}
\centering
\setlength{\tabcolsep}{1.0mm}
\fontsize{6pt}{7pt}\selectfont
\begin{tabular}{lcccccccccccccccccc|c}
\toprule
                 & XGBOD+RD & DSAD+RD & DSAD+HC & DSAD+AB & SB+RD & SB+HC & SB+DB & SB+AB & SB-NCE+RD & SB-NCE+HC & SB-NCE+DB & SB-NCE+AB & OC+RD & OC+HC & OC+AB & OC-NCE+RD & OC-NCE+HC & OC-NCE+AB & IMBoost \\
\midrule
ALOI             & 0.805 & 0.542   & 0.542   & 0.542   & 0.522 & 0.529 & 0.519 & 0.521 & 0.538     & 0.538     & 0.532     & 0.527     & 0.538 & 0.544 & 0.541 & 0.524     & 0.544     & 0.520     & 0.549   \\
annthyroid       & 0.961 & 0.588   & 0.701   & 0.610   & 0.572 & 0.553 & 0.636 & 0.594 & 0.654     & 0.716     & 0.668     & 0.703     & 0.641 & 0.557 & 0.567 & 0.693     & 0.701     & 0.716     & 0.783   \\
backdoor         & 0.993 & 0.885   & 0.905   & 0.905   & 0.727 & 0.585 & 0.787 & 0.734 & 0.667     & 0.877     & 0.879     & 0.898     & 0.785 & 0.819 & 0.912 & 0.671     & 0.910     & 0.903     & 0.941   \\
breastw          & 0.922 & 0.880   & 0.960   & 0.880   & 0.579 & 0.495 & 0.596 & 0.629 & 0.985     & 0.819     & 0.977     & 0.968     & 0.945 & 0.886 & 0.941 & 0.978     & 0.937     & 0.984     & 0.977   \\
campaign         & 0.925 & 0.542   & 0.547   & 0.546   & 0.499 & 0.401 & 0.560 & 0.403 & 0.636     & 0.575     & 0.648     & 0.566     & 0.537 & 0.540 & 0.588 & 0.614     & 0.650     & 0.614     & 0.768   \\
cardio           & 0.890 & 0.564   & 0.790   & 0.620   & 0.663 & 0.612 & 0.726 & 0.617 & 0.663     & 0.612     & 0.726     & 0.617     & 0.840 & 0.827 & 0.789 & 0.840     & 0.827     & 0.789     & 0.969   \\
Cardiotocography & 0.857 & 0.547   & 0.656   & 0.550   & 0.685 & 0.594 & 0.724 & 0.788 & 0.922     & 0.659     & 0.879     & 0.908     & 0.773 & 0.614 & 0.749 & 0.906     & 0.663     & 0.913     & 0.902   \\
celeba           & 0.964 & 0.516   & 0.518   & 0.517   & 0.579 & 0.475 & 0.785 & 0.535 & 0.902     & 0.844     & 0.887     & 0.888     & 0.677 & 0.542 & 0.561 & 0.863     & 0.877     & 0.891     & 0.921   \\
census           & 0.904 & 0.521   & 0.519   & 0.524   & 0.448 & 0.552 & 0.400 & 0.510 & 0.754     & 0.527     & 0.545     & 0.531     & 0.360 & 0.514 & 0.396 & 0.784     & 0.623     & 0.545     & 0.903   \\
cover            & 0.994 & 0.656   & 0.671   & 0.671   & 0.815 & 0.848 & 0.790 & 0.777 & 0.905     & 0.992     & 0.973     & 0.995     & 0.532 & 0.630 & 0.590 & 0.792     & 0.776     & 0.736     & 0.994   \\
donors           & 1.000 & 0.667   & 0.803   & 0.654   & 0.734 & 0.706 & 0.739 & 0.683 & 0.999     & 0.965     & 0.954     & 0.932     & 0.809 & 0.753 & 0.810 & 0.997     & 0.970     & 0.970     & 0.999   \\
fault            & 0.693 & 0.630   & 0.628   & 0.637   & 0.505 & 0.468 & 0.464 & 0.468 & 0.712     & 0.591     & 0.583     & 0.579     & 0.516 & 0.603 & 0.579 & 0.678     & 0.644     & 0.610     & 0.636   \\
fraud            & 0.959 & 0.874   & 0.890   & 0.892   & 0.807 & 0.794 & 0.833 & 0.795 & 0.855     & 0.927     & 0.840     & 0.932     & 0.880 & 0.897 & 0.889 & 0.881     & 0.933     & 0.930     & 0.957   \\
glass            & 0.500 & 0.784   & 0.930   & 0.861   & 0.390 & 0.464 & 0.422 & 0.422 & 0.769     & 0.809     & 0.422     & 0.422     & 0.787 & 0.844 & 0.810 & 0.832     & 0.994     & 0.940     & 0.837   \\
Hepatitis        & 0.500 & 0.909   & 0.885   & 0.925   & 0.587 & 0.591 & 0.547 & 0.624 & 0.969     & 0.919     & 0.927     & 0.973     & 0.810 & 0.734 & 0.801 & 0.978     & 0.925     & 0.960     & 0.869   \\
http             & 1.000 & 0.923   & 1.000   & 1.000   & 0.926 & 0.937 & 0.946 & 0.658 & 0.991     & 0.999     & 1.000     & 0.999     & 0.994 & 0.999 & 0.999 & 0.998     & 1.000     & 1.000     & 0.997   \\
InternetAds      & 0.773 & 0.678   & 0.719   & 0.689   & 0.715 & 0.663 & 0.637 & 0.666 & 0.738     & 0.691     & 0.761     & 0.697     & 0.705 & 0.659 & 0.705 & 0.739     & 0.685     & 0.702     & 0.840   \\
Ionosphere       & 0.736 & 0.882   & 0.879   & 0.894   & 0.570 & 0.542 & 0.599 & 0.552 & 0.936     & 0.861     & 0.903     & 0.958     & 0.852 & 0.868 & 0.855 & 0.916     & 0.919     & 0.935     & 0.822   \\
landsat          & 0.891 & 0.658   & 0.661   & 0.672   & 0.458 & 0.431 & 0.406 & 0.410 & 0.751     & 0.627     & 0.545     & 0.590     & 0.516 & 0.618 & 0.555 & 0.698     & 0.633     & 0.582     & 0.807   \\
letter           & 0.723 & 0.795   & 0.797   & 0.797   & 0.484 & 0.475 & 0.407 & 0.467 & 0.680     & 0.637     & 0.647     & 0.654     & 0.717 & 0.782 & 0.773 & 0.757     & 0.757     & 0.771     & 0.643   \\
Lymphography     & 0.500 & 0.718   & 0.922   & 0.937   & 0.694 & 0.762 & 0.727 & 0.753 & 0.825     & 1.000     & 0.897     & 1.000     & 0.816 & 0.887 & 0.967 & 0.812     & 0.923     & 1.000     & 0.989   \\
magic.gamma      & 0.882 & 0.726   & 0.758   & 0.729   & 0.652 & 0.537 & 0.601 & 0.616 & 0.652     & 0.537     & 0.601     & 0.616     & 0.695 & 0.656 & 0.642 & 0.695     & 0.656     & 0.642     & 0.822   \\
mammography      & 0.914 & 0.708   & 0.690   & 0.703   & 0.353 & 0.250 & 0.375 & 0.342 & 0.632     & 0.648     & 0.695     & 0.658     & 0.609 & 0.501 & 0.692 & 0.751     & 0.829     & 0.718     & 0.922   \\
musk             & 0.996 & 0.975   & 1.000   & 1.000   & 0.857 & 0.907 & 0.739 & 0.914 & 0.999     & 0.990     & 0.999     & 0.996     & 0.922 & 0.845 & 0.882 & 0.995     & 1.000     & 1.000     & 0.999   \\
optdigits        & 0.977 & 0.658   & 0.703   & 0.706   & 0.690 & 0.691 & 0.701 & 0.657 & 0.989     & 0.980     & 0.998     & 0.998     & 0.612 & 0.671 & 0.543 & 0.975     & 0.994     & 0.998     & 0.998   \\
Pageblocks       & 0.948 & 0.658   & 0.781   & 0.679   & 0.542 & 0.509 & 0.561 & 0.541 & 0.915     & 0.782     & 0.897     & 0.890     & 0.720 & 0.703 & 0.711 & 0.900     & 0.849     & 0.918     & 0.908   \\
pendigits        & 0.984 & 0.711   & 0.724   & 0.734   & 0.623 & 0.721 & 0.628 & 0.692 & 0.979     & 0.984     & 0.862     & 0.989     & 0.791 & 0.779 & 0.845 & 0.988     & 0.998     & 0.997     & 0.978   \\
Pima             & 0.652 & 0.562   & 0.569   & 0.564   & 0.593 & 0.523 & 0.611 & 0.598 & 0.731     & 0.674     & 0.676     & 0.652     & 0.649 & 0.574 & 0.622 & 0.655     & 0.623     & 0.668     & 0.752   \\
satellite        & 0.914 & 0.692   & 0.760   & 0.689   & 0.631 & 0.579 & 0.666 & 0.548 & 0.923     & 0.613     & 0.903     & 0.864     & 0.765 & 0.703 & 0.782 & 0.927     & 0.754     & 0.860     & 0.813   \\
satimage-2       & 0.982 & 0.866   & 0.986   & 0.906   & 0.890 & 0.953 & 0.891 & 0.891 & 0.974     & 0.987     & 0.983     & 0.976     & 0.973 & 0.980 & 0.909 & 0.982     & 0.989     & 0.986     & 0.991   \\
shuttle          & 0.999 & 0.663   & 0.977   & 0.572   & 0.980 & 0.976 & 0.940 & 0.932 & 0.955     & 0.976     & 0.975     & 0.969     & 0.892 & 0.912 & 0.875 & 0.944     & 0.926     & 0.913     & 0.989   \\
skin             & 0.999 & 0.711   & 0.721   & 0.715   & 0.554 & 0.663 & 0.621 & 0.478 & 0.925     & 0.681     & 0.621     & 0.580     & 0.549 & 0.484 & 0.693 & 0.778     & 0.484     & 0.746     & 0.915   \\
smtp             & 0.955 & 0.852   & 0.835   & 0.835   & 0.676 & 0.676 & 0.781 & 0.676 & 0.676     & 0.877     & 0.779     & 0.780     & 0.835 & 0.871 & 0.866 & 0.835     & 0.876     & 0.842     & 0.891   \\
SpamBase         & 0.880 & 0.539   & 0.579   & 0.545   & 0.629 & 0.378 & 0.606 & 0.412 & 0.872     & 0.421     & 0.838     & 0.804     & 0.650 & 0.518 & 0.615 & 0.851     & 0.690     & 0.821     & 0.813   \\
speech           & 0.621 & 0.501   & 0.466   & 0.477   & 0.489 & 0.440 & 0.490 & 0.440 & 0.522     & 0.464     & 0.486     & 0.440     & 0.467 & 0.488 & 0.463 & 0.514     & 0.502     & 0.463     & 0.502   \\
Stamps           & 0.640 & 0.515   & 0.819   & 0.545   & 0.404 & 0.403 & 0.389 & 0.407 & 0.894     & 0.922     & 0.837     & 0.929     & 0.611 & 0.533 & 0.649 & 0.722     & 0.973     & 0.959     & 0.942   \\
thyroid          & 0.905 & 0.768   & 0.910   & 0.847   & 0.574 & 0.538 & 0.543 & 0.444 & 0.877     & 0.955     & 0.953     & 0.958     & 0.815 & 0.790 & 0.782 & 0.937     & 0.985     & 0.987     & 0.957   \\
vertebral        & 0.617 & 0.550   & 0.547   & 0.551   & 0.514 & 0.502 & 0.501 & 0.554 & 0.655     & 0.764     & 0.816     & 0.879     & 0.557 & 0.545 & 0.493 & 0.670     & 0.791     & 0.744     & 0.658   \\
vowels           & 0.788 & 0.833   & 0.906   & 0.854   & 0.700 & 0.647 & 0.629 & 0.624 & 0.915     & 0.951     & 0.940     & 0.860     & 0.827 & 0.861 & 0.847 & 0.932     & 0.974     & 0.989     & 0.983   \\
Waveform         & 0.769 & 0.692   & 0.786   & 0.764   & 0.709 & 0.643 & 0.679 & 0.647 & 0.815     & 0.841     & 0.859     & 0.813     & 0.712 & 0.736 & 0.738 & 0.834     & 0.823     & 0.839     & 0.802   \\
WBC              & 0.500 & 0.774   & 1.000   & 0.831   & 0.519 & 0.414 & 0.590 & 0.629 & 0.698     & 1.000     & 1.000     & 0.959     & 0.825 & 0.957 & 0.892 & 0.873     & 1.000     & 1.000     & 0.974   \\
WDBC             & 0.500 & 0.812   & 0.930   & 0.905   & 0.915 & 0.890 & 0.893 & 0.892 & 0.929     & 0.948     & 0.913     & 0.944     & 0.834 & 0.860 & 0.869 & 0.873     & 0.994     & 0.995     & 0.993   \\
Wilt             & 0.869 & 0.439   & 0.436   & 0.437   & 0.619 & 0.620 & 0.654 & 0.635 & 0.693     & 0.618     & 0.730     & 0.673     & 0.442 & 0.436 & 0.434 & 0.617     & 0.469     & 0.411     & 0.634   \\
wine             & 0.500 & 0.758   & 1.000   & 1.000   & 0.652 & 0.669 & 0.670 & 0.828 & 0.781     & 1.000     & 0.998     & 1.000     & 0.592 & 0.806 & 0.758 & 0.750     & 1.000     & 1.000     & 0.997   \\
WPBC             & 0.437 & 0.592   & 0.603   & 0.602   & 0.555 & 0.550 & 0.558 & 0.525 & 0.690     & 0.691     & 0.711     & 0.682     & 0.563 & 0.522 & 0.548 & 0.647     & 0.644     & 0.695     & 0.649   \\
yeast            & 0.600 & 0.478   & 0.471   & 0.479   & 0.521 & 0.524 & 0.496 & 0.526 & 0.567     & 0.555     & 0.569     & 0.585     & 0.445 & 0.500 & 0.492 & 0.575     & 0.482     & 0.490     & 0.523   \\
CIFAR10          & 0.740 & 0.713   & 0.781   & 0.735   & 0.757 & 0.694 & 0.684 & 0.772 & 0.821     & 0.824     & 0.807     & 0.868     & 0.806 & 0.762 & 0.824 & 0.849     & 0.865     & 0.896     & 0.918   \\
MNIST-C          & 0.916 & 0.613   & 0.656   & 0.634   & 0.662 & 0.608 & 0.621 & 0.614 & 0.795     & 0.779     & 0.773     & 0.751     & 0.672 & 0.643 & 0.671 & 0.802     & 0.797     & 0.789     & 0.936   \\
MVTec-AD         & 0.688 & 0.704   & 0.762   & 0.740   & 0.681 & 0.745 & 0.694 & 0.718 & 0.773     & 0.812     & 0.769     & 0.792     & 0.672 & 0.715 & 0.693 & 0.706     & 0.787     & 0.731     & 0.809   \\
SVHN             & 0.688 & 0.501   & 0.501   & 0.501   & 0.511 & 0.503 & 0.502 & 0.507 & 0.522     & 0.504     & 0.521     & 0.517     & 0.507 & 0.509 & 0.506 & 0.520     & 0.502     & 0.512     & 0.603   \\
mnist            & 0.962 & 0.769   & 0.796   & 0.783   & 0.678 & 0.714 & 0.676 & 0.667 & 0.867     & 0.811     & 0.819     & 0.831     & 0.689 & 0.757 & 0.710 & 0.849     & 0.857     & 0.826     & 0.829   \\
FashionMNIST     & 0.913 & 0.744   & 0.843   & 0.763   & 0.795 & 0.729 & 0.770 & 0.765 & 0.899     & 0.825     & 0.892     & 0.901     & 0.851 & 0.785 & 0.873 & 0.888     & 0.876     & 0.918     & 0.932   \\
20news           & 0.669 & 0.543   & 0.542   & 0.553   & 0.536 & 0.522 & 0.511 & 0.548 & 0.603     & 0.581     & 0.604     & 0.592     & 0.552 & 0.536 & 0.550 & 0.587     & 0.608     & 0.592     & 0.785   \\
Agnews           & 0.807 & 0.572   & 0.573   & 0.575   & 0.524 & 0.496 & 0.500 & 0.518 & 0.688     & 0.632     & 0.593     & 0.647     & 0.543 & 0.549 & 0.551 & 0.696     & 0.609     & 0.625     & 0.918   \\
Amazon           & 0.742 & 0.493   & 0.490   & 0.492   & 0.497 & 0.537 & 0.499 & 0.495 & 0.501     & 0.480     & 0.520     & 0.509     & 0.483 & 0.501 & 0.495 & 0.545     & 0.520     & 0.509     & 0.882   \\
Imdb             & 0.716 & 0.499   & 0.498   & 0.498   & 0.529 & 0.484 & 0.482 & 0.513 & 0.585     & 0.476     & 0.527     & 0.547     & 0.523 & 0.499 & 0.527 & 0.588     & 0.533     & 0.546     & 0.833   \\
Yelp             & 0.753 & 0.494   & 0.498   & 0.498   & 0.504 & 0.468 & 0.444 & 0.488 & 0.540     & 0.505     & 0.496     & 0.505     & 0.486 & 0.477 & 0.468 & 0.549     & 0.532     & 0.477     & 0.858   \\
\midrule
\textbf{Average} & 0.806 & 0.674   & 0.734   & 0.698   & 0.622 & 0.600 & 0.622 & 0.608 & 0.784     & 0.759     & 0.775     & 0.777     & 0.687 & 0.686 & 0.700 & 0.785     & 0.784     & 0.791     & 0.856   \\
\midrule
\textbf{std}     & 0.022 & 0.053   & 0.040   & 0.044   & 0.078 & 0.070 & 0.090 & 0.079 & 0.068     & 0.056     & 0.051     & 0.054     & 0.059 & 0.050 & 0.051 & 0.061     & 0.038     & 0.042     & 0.022    \\
\bottomrule
\end{tabular}
\caption{Training AUC and Standard Deviation on ADBench}
\label{tab:adbench1}
\end{sidewaystable}

\begin{sidewaystable}[h!]
\renewcommand\thetable{C.3}
\centering
\setlength{\tabcolsep}{1.0mm}
\fontsize{6pt}{7pt}\selectfont
\begin{tabular}{lcccccccccccccccccc|c}
\toprule
                 & XGBOD+RD & DSAD+RD & DSAD+HC & DSAD+AB & SB+RD & SB+HC & SB+DB & SB+AB & SB-NCE+RD & SB-NCE+HC & SB-NCE+DB & SB-NCE+AB & OC+RD & OC+HC & OC+AB & OC-NCE+RD & OC-NCE+HC & OC-NCE+AB & IMBoost \\
\midrule
ALOI             & 0.147 & 0.046   & 0.048   & 0.048   & 0.044 & 0.047 & 0.043 & 0.044 & 0.044     & 0.060     & 0.041     & 0.054     & 0.042 & 0.047 & 0.044 & 0.039     & 0.056     & 0.049     & 0.042   \\
annthyroid       & 0.760 & 0.100   & 0.406   & 0.188   & 0.121 & 0.118 & 0.199 & 0.151 & 0.217     & 0.411     & 0.261     & 0.337     & 0.185 & 0.107 & 0.140 & 0.261     & 0.369     & 0.375     & 0.434   \\
backdoor         & 0.934 & 0.191   & 0.365   & 0.442   & 0.056 & 0.042 & 0.070 & 0.067 & 0.389     & 0.450     & 0.342     & 0.377     & 0.084 & 0.158 & 0.271 & 0.254     & 0.728     & 0.437     & 0.773   \\
breastw          & 0.855 & 0.740   & 0.927   & 0.747   & 0.621 & 0.516 & 0.606 & 0.694 & 0.961     & 0.786     & 0.947     & 0.942     & 0.872 & 0.753 & 0.861 & 0.949     & 0.898     & 0.965     & 0.943   \\
campaign         & 0.577 & 0.127   & 0.131   & 0.130   & 0.140 & 0.092 & 0.146 & 0.094 & 0.233     & 0.238     & 0.205     & 0.205     & 0.132 & 0.124 & 0.140 & 0.173     & 0.257     & 0.190     & 0.309   \\
cardio           & 0.679 & 0.168   & 0.586   & 0.275   & 0.398 & 0.302 & 0.422 & 0.296 & 0.398     & 0.302     & 0.422     & 0.296     & 0.562 & 0.430 & 0.474 & 0.562     & 0.430     & 0.474     & 0.844   \\
Cardiotocography & 0.692 & 0.303   & 0.524   & 0.311   & 0.452 & 0.366 & 0.519 & 0.604 & 0.800     & 0.492     & 0.731     & 0.826     & 0.583 & 0.389 & 0.572 & 0.781     & 0.524     & 0.814     & 0.691   \\
celeba           & 0.341 & 0.021   & 0.021   & 0.021   & 0.030 & 0.022 & 0.062 & 0.025 & 0.161     & 0.194     & 0.137     & 0.212     & 0.036 & 0.023 & 0.023 & 0.121     & 0.171     & 0.195     & 0.132   \\
census           & 0.449 & 0.059   & 0.059   & 0.059   & 0.051 & 0.068 & 0.048 & 0.058 & 0.204     & 0.135     & 0.062     & 0.076     & 0.045 & 0.059 & 0.047 & 0.230     & 0.095     & 0.078     & 0.477   \\
cover            & 0.947 & 0.017   & 0.032   & 0.032   & 0.112 & 0.088 & 0.040 & 0.036 & 0.505     & 0.873     & 0.771     & 0.926     & 0.009 & 0.016 & 0.011 & 0.082     & 0.074     & 0.278     & 0.613   \\
donors           & 1.000 & 0.081   & 0.141   & 0.078   & 0.134 & 0.122 & 0.173 & 0.092 & 0.979     & 0.902     & 0.755     & 0.903     & 0.133 & 0.110 & 0.162 & 0.922     & 0.902     & 0.913     & 0.990   \\
fault            & 0.555 & 0.469   & 0.480   & 0.476   & 0.383 & 0.370 & 0.338 & 0.368 & 0.543     & 0.505     & 0.483     & 0.482     & 0.383 & 0.468 & 0.394 & 0.508     & 0.538     & 0.474     & 0.472   \\
fraud            & 0.731 & 0.054   & 0.219   & 0.198   & 0.193 & 0.241 & 0.225 & 0.254 & 0.359     & 0.662     & 0.219     & 0.690     & 0.049 & 0.124 & 0.104 & 0.190     & 0.651     & 0.637     & 0.660   \\
glass            & 0.040 & 0.253   & 0.660   & 0.511   & 0.033 & 0.058 & 0.034 & 0.034 & 0.275     & 0.597     & 0.034     & 0.034     & 0.282 & 0.465 & 0.438 & 0.321     & 0.941     & 0.872     & 0.135   \\
Hepatitis        & 0.161 & 0.874   & 0.856   & 0.913   & 0.292 & 0.395 & 0.265 & 0.486 & 0.926     & 0.839     & 0.809     & 0.941     & 0.437 & 0.539 & 0.632 & 0.943     & 0.850     & 0.928     & 0.704   \\
http             & 0.996 & 0.295   & 0.998   & 0.998   & 0.164 & 0.216 & 0.496 & 0.311 & 0.583     & 0.998     & 0.997     & 0.997     & 0.423 & 0.882 & 0.737 & 0.660     & 0.997     & 0.956     & 0.461   \\
InternetAds      & 0.493 & 0.418   & 0.537   & 0.479   & 0.542 & 0.403 & 0.338 & 0.427 & 0.536     & 0.485     & 0.573     & 0.534     & 0.538 & 0.403 & 0.479 & 0.555     & 0.516     & 0.504     & 0.747   \\
Ionosphere       & 0.682 & 0.809   & 0.827   & 0.854   & 0.574 & 0.543 & 0.534 & 0.547 & 0.904     & 0.822     & 0.873     & 0.946     & 0.801 & 0.820 & 0.762 & 0.873     & 0.894     & 0.897     & 0.721   \\
landsat          & 0.710 & 0.306   & 0.323   & 0.327   & 0.213 & 0.240 & 0.173 & 0.186 & 0.516     & 0.408     & 0.256     & 0.404     & 0.202 & 0.250 & 0.222 & 0.369     & 0.333     & 0.370     & 0.495   \\
letter           & 0.348 & 0.326   & 0.374   & 0.375   & 0.065 & 0.073 & 0.063 & 0.075 & 0.202     & 0.276     & 0.274     & 0.284     & 0.252 & 0.334 & 0.307 & 0.279     & 0.385     & 0.459     & 0.133   \\
Lymphography     & 0.039 & 0.466   & 0.920   & 0.921   & 0.182 & 0.532 & 0.272 & 0.486 & 0.585     & 1.000     & 0.707     & 1.000     & 0.439 & 0.732 & 0.772 & 0.550     & 0.920     & 1.000     & 0.848   \\
magic.gamma      & 0.822 & 0.613   & 0.701   & 0.615   & 0.526 & 0.425 & 0.492 & 0.542 & 0.526     & 0.425     & 0.492     & 0.542     & 0.613 & 0.559 & 0.548 & 0.613     & 0.559     & 0.548     & 0.779   \\
mammography      & 0.501 & 0.076   & 0.105   & 0.093   & 0.036 & 0.040 & 0.036 & 0.053 & 0.251     & 0.381     & 0.211     & 0.373     & 0.043 & 0.049 & 0.072 & 0.239     & 0.392     & 0.257     & 0.629   \\
musk             & 0.987 & 0.647   & 1.000   & 1.000   & 0.268 & 0.425 & 0.396 & 0.436 & 0.983     & 0.859     & 0.974     & 0.941     & 0.501 & 0.396 & 0.429 & 0.928     & 1.000     & 1.000     & 0.986   \\
optdigits        & 0.682 & 0.055   & 0.145   & 0.133   & 0.063 & 0.073 & 0.081 & 0.055 & 0.898     & 0.952     & 0.986     & 0.987     & 0.039 & 0.130 & 0.047 & 0.708     & 0.879     & 0.986     & 0.898   \\
Pageblocks       & 0.692 & 0.217   & 0.529   & 0.312   & 0.242 & 0.138 & 0.217 & 0.231 & 0.656     & 0.574     & 0.720     & 0.735     & 0.298 & 0.311 & 0.332 & 0.649     & 0.623     & 0.735     & 0.721   \\
pendigits        & 0.614 & 0.051   & 0.065   & 0.066   & 0.085 & 0.094 & 0.037 & 0.176 & 0.838     & 0.942     & 0.536     & 0.945     & 0.059 & 0.102 & 0.108 & 0.849     & 0.957     & 0.970     & 0.403   \\
Pima             & 0.515 & 0.411   & 0.441   & 0.412   & 0.451 & 0.418 & 0.459 & 0.437 & 0.566     & 0.556     & 0.527     & 0.516     & 0.470 & 0.422 & 0.443 & 0.493     & 0.503     & 0.489     & 0.635   \\
satellite        & 0.847 & 0.520   & 0.671   & 0.526   & 0.647 & 0.503 & 0.681 & 0.504 & 0.882     & 0.572     & 0.861     & 0.816     & 0.725 & 0.542 & 0.710 & 0.896     & 0.702     & 0.837     & 0.748   \\
satimage-2       & 0.805 & 0.088   & 0.932   & 0.355   & 0.398 & 0.858 & 0.198 & 0.198 & 0.926     & 0.954     & 0.922     & 0.920     & 0.594 & 0.825 & 0.197 & 0.877     & 0.946     & 0.951     & 0.938   \\
shuttle          & 0.994 & 0.175   & 0.885   & 0.136   & 0.891 & 0.856 & 0.745 & 0.587 & 0.631     & 0.856     & 0.772     & 0.815     & 0.511 & 0.592 & 0.578 & 0.687     & 0.706     & 0.719     & 0.970   \\
skin             & 0.994 & 0.310   & 0.318   & 0.317   & 0.240 & 0.301 & 0.284 & 0.233 & 0.595     & 0.332     & 0.284     & 0.295     & 0.233 & 0.212 & 0.344 & 0.428     & 0.212     & 0.387     & 0.727   \\
smtp             & 0.101 & 0.384   & 0.555   & 0.555   & 0.145 & 0.145 & 0.225 & 0.145 & 0.145     & 0.542     & 0.228     & 0.394     & 0.197 & 0.200 & 0.202 & 0.197     & 0.429     & 0.545     & 0.143   \\
SpamBase         & 0.829 & 0.407   & 0.475   & 0.409   & 0.566 & 0.352 & 0.539 & 0.392 & 0.805     & 0.464     & 0.795     & 0.774     & 0.540 & 0.403 & 0.472 & 0.773     & 0.636     & 0.777     & 0.790   \\
speech           & 0.085 & 0.053   & 0.054   & 0.047   & 0.020 & 0.015 & 0.019 & 0.015 & 0.030     & 0.031     & 0.026     & 0.015     & 0.018 & 0.018 & 0.016 & 0.036     & 0.031     & 0.016     & 0.037   \\
Stamps           & 0.252 & 0.142   & 0.688   & 0.250   & 0.121 & 0.168 & 0.105 & 0.179 & 0.520     & 0.909     & 0.623     & 0.812     & 0.210 & 0.223 & 0.287 & 0.431     & 0.932     & 0.885     & 0.600   \\
thyroid          & 0.743 & 0.103   & 0.633   & 0.433   & 0.097 & 0.067 & 0.131 & 0.049 & 0.568     & 0.814     & 0.742     & 0.750     & 0.208 & 0.197 & 0.175 & 0.549     & 0.811     & 0.812     & 0.588   \\
vertebral        & 0.269 & 0.192   & 0.249   & 0.283   & 0.144 & 0.152 & 0.152 & 0.187 & 0.319     & 0.503     & 0.593     & 0.666     & 0.155 & 0.171 & 0.138 & 0.299     & 0.577     & 0.493     & 0.243   \\
vowels           & 0.343 & 0.251   & 0.532   & 0.351   & 0.113 & 0.094 & 0.056 & 0.078 & 0.602     & 0.865     & 0.746     & 0.615     & 0.248 & 0.344 & 0.323 & 0.646     & 0.913     & 0.949     & 0.844   \\
Waveform         & 0.152 & 0.296   & 0.569   & 0.527   & 0.075 & 0.075 & 0.104 & 0.068 & 0.423     & 0.550     & 0.610     & 0.558     & 0.436 & 0.477 & 0.474 & 0.485     & 0.624     & 0.561     & 0.212   \\
WBC              & 0.045 & 0.316   & 1.000   & 0.667   & 0.148 & 0.109 & 0.284 & 0.260 & 0.463     & 1.000     & 1.000     & 0.921     & 0.334 & 0.700 & 0.604 & 0.522     & 1.000     & 1.000     & 0.499   \\
WDBC             & 0.027 & 0.236   & 0.706   & 0.638   & 0.338 & 0.379 & 0.367 & 0.379 & 0.544     & 0.698     & 0.467     & 0.612     & 0.350 & 0.465 & 0.466 & 0.331     & 0.907     & 0.914     & 0.794   \\
Wilt             & 0.434 & 0.049   & 0.049   & 0.049   & 0.074 & 0.073 & 0.090 & 0.080 & 0.106     & 0.097     & 0.155     & 0.097     & 0.047 & 0.046 & 0.046 & 0.075     & 0.056     & 0.046     & 0.098   \\
wine             & 0.078 & 0.497   & 1.000   & 1.000   & 0.152 & 0.303 & 0.303 & 0.526 & 0.407     & 1.000     & 0.985     & 1.000     & 0.246 & 0.546 & 0.537 & 0.422     & 1.000     & 1.000     & 0.974   \\
WPBC             & 0.247 & 0.389   & 0.436   & 0.437   & 0.313 & 0.348 & 0.296 & 0.291 & 0.511     & 0.594     & 0.546     & 0.556     & 0.323 & 0.316 & 0.341 & 0.422     & 0.561     & 0.547     & 0.393   \\
yeast            & 0.459 & 0.332   & 0.350   & 0.342   & 0.345 & 0.355 & 0.324 & 0.342 & 0.376     & 0.392     & 0.379     & 0.410     & 0.316 & 0.348 & 0.343 & 0.387     & 0.353     & 0.352     & 0.360   \\
CIFAR10          & 0.239 & 0.131   & 0.350   & 0.187   & 0.254 & 0.191 & 0.166 & 0.306 & 0.396     & 0.591     & 0.378     & 0.599     & 0.307 & 0.217 & 0.346 & 0.375     & 0.607     & 0.623     & 0.674   \\
MNIST-C          & 0.722 & 0.122   & 0.265   & 0.190   & 0.157 & 0.147 & 0.123 & 0.123 & 0.289     & 0.502     & 0.356     & 0.368     & 0.196 & 0.180 & 0.156 & 0.309     & 0.494     & 0.391     & 0.807   \\
MVTec-AD         & 0.500 & 0.362   & 0.511   & 0.485   & 0.248 & 0.458 & 0.259 & 0.316 & 0.409     & 0.631     & 0.492     & 0.532     & 0.271 & 0.447 & 0.364 & 0.366     & 0.603     & 0.493     & 0.533   \\
SVHN             & 0.191 & 0.053   & 0.056   & 0.055   & 0.054 & 0.053 & 0.053 & 0.055 & 0.056     & 0.062     & 0.057     & 0.061     & 0.053 & 0.054 & 0.054 & 0.056     & 0.062     & 0.060     & 0.115   \\
mnist            & 0.774 & 0.303   & 0.455   & 0.359   & 0.248 & 0.299 & 0.253 & 0.260 & 0.558     & 0.577     & 0.462     & 0.607     & 0.259 & 0.330 & 0.297 & 0.536     & 0.648     & 0.572     & 0.527   \\
FashionMNIST     & 0.585 & 0.265   & 0.571   & 0.318   & 0.496 & 0.331 & 0.360 & 0.498 & 0.687     & 0.653     & 0.657     & 0.783     & 0.553 & 0.380 & 0.612 & 0.631     & 0.718     & 0.803     & 0.737   \\
20news           & 0.181 & 0.068   & 0.070   & 0.077   & 0.063 & 0.069 & 0.063 & 0.085 & 0.114     & 0.145     & 0.120     & 0.128     & 0.064 & 0.067 & 0.071 & 0.109     & 0.150     & 0.132     & 0.372   \\
Agnews           & 0.307 & 0.066   & 0.068   & 0.067   & 0.066 & 0.053 & 0.056 & 0.062 & 0.107     & 0.177     & 0.103     & 0.193     & 0.067 & 0.061 & 0.064 & 0.105     & 0.123     & 0.112     & 0.557   \\
Amazon           & 0.214 & 0.049   & 0.049   & 0.049   & 0.050 & 0.056 & 0.050 & 0.050 & 0.051     & 0.054     & 0.053     & 0.061     & 0.048 & 0.050 & 0.048 & 0.055     & 0.056     & 0.059     & 0.348   \\
Imdb             & 0.186 & 0.049   & 0.049   & 0.049   & 0.056 & 0.047 & 0.047 & 0.053 & 0.065     & 0.051     & 0.054     & 0.063     & 0.055 & 0.049 & 0.054 & 0.067     & 0.057     & 0.058     & 0.423   \\
Yelp             & 0.233 & 0.048   & 0.049   & 0.049   & 0.049 & 0.049 & 0.042 & 0.050 & 0.059     & 0.054     & 0.047     & 0.053     & 0.048 & 0.045 & 0.046 & 0.060     & 0.063     & 0.048     & 0.490   \\
\midrule
\textbf{Average} & 0.505 & 0.253   & 0.439   & 0.355   & 0.229 & 0.235 & 0.231 & 0.239 & 0.468     & 0.540     & 0.489     & 0.543     & 0.283 & 0.311 & 0.315 & 0.443     & 0.551     & 0.561     & 0.565   \\
\midrule
\textbf{std}     & 0.033 & 0.055   & 0.061   & 0.064   & 0.066 & 0.071 & 0.089 & 0.083 & 0.105     & 0.082     & 0.098     & 0.078     & 0.071 & 0.074 & 0.078 & 0.100     & 0.059     & 0.068     & 0.062   \\
\bottomrule
\end{tabular}
\caption{Training AP and Standard Deviation on ADBench}
\label{tab:adbench2}
\end{sidewaystable}

\begin{sidewaystable}[h!]
\renewcommand\thetable{C.4}
\centering
\setlength{\tabcolsep}{1.0mm}
\fontsize{6pt}{7pt}\selectfont
\begin{tabular}{lcccccccccccccccccc|c}
\toprule
                 & XGBOD+RD & DSAD+RD & DSAD+HC & DSAD+AB & SB+RD & SB+HC & SB+DB & SB+AB & SB-NCE+RD & SB-NCE+HC & SB-NCE+DB & SB-NCE+AB & OC+RD & OC+HC & OC+AB & OC-NCE+RD & OC-NCE+HC & OC-NCE+AB & IMBoost \\
\midrule
ALOI             & 0.828 & 0.551   & 0.551   & 0.552   & 0.525 & 0.532 & 0.522 & 0.524 & 0.514     & 0.530     & 0.531     & 0.524     & 0.547 & 0.552 & 0.548 & 0.522     & 0.557     & 0.524     & 0.547   \\
annthyroid       & 0.958 & 0.620   & 0.725   & 0.635   & 0.556 & 0.528 & 0.628 & 0.575 & 0.639     & 0.714     & 0.691     & 0.714     & 0.647 & 0.579 & 0.569 & 0.685     & 0.688     & 0.725     & 0.789   \\
backdoor         & 0.993 & 0.884   & 0.903   & 0.903   & 0.734 & 0.588 & 0.784 & 0.733 & 0.677     & 0.880     & 0.877     & 0.899     & 0.787 & 0.817 & 0.909 & 0.670     & 0.907     & 0.900     & 0.940   \\
breastw          & 0.915 & 0.884   & 0.947   & 0.885   & 0.523 & 0.520 & 0.611 & 0.625 & 0.975     & 0.800     & 0.963     & 0.964     & 0.931 & 0.892 & 0.936 & 0.964     & 0.930     & 0.969     & 0.967   \\
campaign         & 0.922 & 0.542   & 0.547   & 0.545   & 0.491 & 0.407 & 0.558 & 0.406 & 0.628     & 0.578     & 0.646     & 0.564     & 0.541 & 0.545 & 0.592 & 0.617     & 0.649     & 0.610     & 0.765   \\
cardio           & 0.912 & 0.596   & 0.775   & 0.639   & 0.713 & 0.657 & 0.747 & 0.657 & 0.713     & 0.657     & 0.747     & 0.657     & 0.835 & 0.830 & 0.773 & 0.835     & 0.830     & 0.773     & 0.948   \\
Cardiotocography & 0.857 & 0.567   & 0.658   & 0.566   & 0.694 & 0.600 & 0.753 & 0.795 & 0.918     & 0.657     & 0.874     & 0.921     & 0.775 & 0.614 & 0.747 & 0.907     & 0.656     & 0.918     & 0.898   \\
celeba           & 0.964 & 0.519   & 0.521   & 0.520   & 0.581 & 0.478 & 0.783 & 0.539 & 0.905     & 0.850     & 0.888     & 0.890     & 0.679 & 0.544 & 0.562 & 0.864     & 0.881     & 0.893     & 0.920   \\
census           & 0.898 & 0.522   & 0.520   & 0.524   & 0.449 & 0.551 & 0.398 & 0.508 & 0.751     & 0.525     & 0.543     & 0.528     & 0.357 & 0.512 & 0.394 & 0.784     & 0.621     & 0.544     & 0.896   \\
cover            & 0.994 & 0.651   & 0.666   & 0.666   & 0.818 & 0.844 & 0.786 & 0.777 & 0.902     & 0.988     & 0.973     & 0.994     & 0.532 & 0.624 & 0.591 & 0.781     & 0.773     & 0.730     & 0.994   \\
donors           & 1.000 & 0.668   & 0.804   & 0.655   & 0.734 & 0.705 & 0.738 & 0.682 & 0.999     & 0.964     & 0.954     & 0.931     & 0.809 & 0.753 & 0.811 & 0.997     & 0.970     & 0.969     & 0.999   \\
fault            & 0.683 & 0.650   & 0.649   & 0.652   & 0.493 & 0.464 & 0.439 & 0.458 & 0.716     & 0.592     & 0.579     & 0.576     & 0.514 & 0.620 & 0.583 & 0.684     & 0.641     & 0.609     & 0.602   \\
fraud            & 0.951 & 0.887   & 0.902   & 0.898   & 0.842 & 0.828 & 0.867 & 0.833 & 0.872     & 0.932     & 0.868     & 0.935     & 0.888 & 0.898 & 0.900 & 0.882     & 0.939     & 0.935     & 0.952   \\
glass            & 0.500 & 0.815   & 0.909   & 0.900   & 0.228 & 0.243 & 0.233 & 0.233 & 0.586     & 0.826     & 0.233     & 0.233     & 0.811 & 0.807 & 0.759 & 0.840     & 0.974     & 0.923     & 0.905   \\
Hepatitis        & 0.500 & 0.833   & 0.752   & 0.807   & 0.637 & 0.661 & 0.748 & 0.643 & 0.750     & 0.566     & 0.665     & 0.702     & 0.723 & 0.699 & 0.618 & 0.745     & 0.766     & 0.747     & 0.781   \\
http             & 1.000 & 0.922   & 1.000   & 1.000   & 0.925 & 0.937 & 0.943 & 0.659 & 0.990     & 1.000     & 1.000     & 1.000     & 0.994 & 0.999 & 0.998 & 0.998     & 1.000     & 1.000     & 0.997   \\
InternetAds      & 0.766 & 0.717   & 0.764   & 0.747   & 0.755 & 0.680 & 0.657 & 0.687 & 0.759     & 0.725     & 0.765     & 0.741     & 0.755 & 0.669 & 0.745 & 0.776     & 0.719     & 0.750     & 0.881   \\
Ionosphere       & 0.777 & 0.890   & 0.945   & 0.892   & 0.536 & 0.525 & 0.538 & 0.534 & 0.905     & 0.839     & 0.882     & 0.938     & 0.875 & 0.894 & 0.869 & 0.918     & 0.946     & 0.917     & 0.807   \\
landsat          & 0.888 & 0.651   & 0.658   & 0.666   & 0.466 & 0.430 & 0.408 & 0.403 & 0.736     & 0.633     & 0.545     & 0.580     & 0.501 & 0.609 & 0.540 & 0.675     & 0.638     & 0.571     & 0.798   \\
letter           & 0.734 & 0.834   & 0.832   & 0.833   & 0.505 & 0.512 & 0.413 & 0.499 & 0.666     & 0.649     & 0.634     & 0.647     & 0.752 & 0.825 & 0.815 & 0.780     & 0.810     & 0.780     & 0.611   \\
Lymphography     & 0.500 & 0.939   & 0.953   & 0.969   & 0.617 & 0.632 & 0.625 & 0.625 & 0.783     & 0.992     & 1.000     & 0.992     & 0.911 & 0.922 & 0.934 & 0.958     & 0.981     & 0.981     & 0.981   \\
magic.gamma      & 0.874 & 0.723   & 0.754   & 0.725   & 0.646 & 0.537 & 0.594 & 0.611 & 0.646     & 0.537     & 0.594     & 0.611     & 0.689 & 0.649 & 0.636 & 0.689     & 0.649     & 0.636     & 0.821   \\
mammography      & 0.903 & 0.720   & 0.718   & 0.727   & 0.384 & 0.293 & 0.415 & 0.388 & 0.653     & 0.688     & 0.705     & 0.693     & 0.660 & 0.520 & 0.732 & 0.786     & 0.843     & 0.747     & 0.936   \\
musk             & 1.000 & 0.976   & 1.000   & 1.000   & 0.866 & 0.910 & 0.771 & 0.914 & 0.999     & 0.989     & 0.999     & 0.996     & 0.910 & 0.908 & 0.932 & 0.992     & 1.000     & 1.000     & 0.999   \\
optdigits        & 0.955 & 0.646   & 0.692   & 0.695   & 0.648 & 0.646 & 0.658 & 0.611 & 0.988     & 0.991     & 0.999     & 1.000     & 0.610 & 0.653 & 0.519 & 0.974     & 0.995     & 0.996     & 0.997   \\
Pageblocks       & 0.942 & 0.674   & 0.789   & 0.686   & 0.526 & 0.486 & 0.549 & 0.525 & 0.902     & 0.764     & 0.884     & 0.880     & 0.714 & 0.691 & 0.699 & 0.889     & 0.836     & 0.904     & 0.897   \\
pendigits        & 0.984 & 0.749   & 0.769   & 0.764   & 0.590 & 0.716 & 0.636 & 0.703 & 0.973     & 0.980     & 0.856     & 0.972     & 0.800 & 0.804 & 0.843 & 0.984     & 0.996     & 0.994     & 0.977   \\
Pima             & 0.608 & 0.589   & 0.613   & 0.583   & 0.619 & 0.486 & 0.610 & 0.577 & 0.720     & 0.661     & 0.663     & 0.601     & 0.651 & 0.586 & 0.633 & 0.663     & 0.642     & 0.663     & 0.732   \\
satellite        & 0.902 & 0.672   & 0.750   & 0.670   & 0.638 & 0.591 & 0.670 & 0.562 & 0.920     & 0.620     & 0.903     & 0.870     & 0.760 & 0.693 & 0.780 & 0.921     & 0.756     & 0.864     & 0.815   \\
satimage-2       & 0.987 & 0.846   & 0.987   & 0.890   & 0.895 & 0.973 & 0.903 & 0.903 & 0.988     & 0.991     & 0.996     & 0.969     & 0.974 & 0.987 & 0.918 & 0.983     & 0.991     & 0.992     & 0.991   \\
shuttle          & 0.999 & 0.672   & 0.976   & 0.571   & 0.979 & 0.978 & 0.950 & 0.939 & 0.956     & 0.978     & 0.976     & 0.968     & 0.887 & 0.911 & 0.875 & 0.943     & 0.924     & 0.914     & 0.988   \\
skin             & 0.999 & 0.712   & 0.722   & 0.715   & 0.554 & 0.665 & 0.621 & 0.479 & 0.926     & 0.683     & 0.621     & 0.581     & 0.550 & 0.484 & 0.693 & 0.779     & 0.484     & 0.745     & 0.916   \\
smtp             & 0.950 & 0.800   & 0.731   & 0.731   & 0.540 & 0.540 & 0.587 & 0.540 & 0.540     & 0.683     & 0.634     & 0.654     & 0.718 & 0.742 & 0.749 & 0.718     & 0.730     & 0.687     & 0.740   \\
SpamBase         & 0.876 & 0.550   & 0.584   & 0.553   & 0.622 & 0.392 & 0.604 & 0.434 & 0.856     & 0.437     & 0.824     & 0.794     & 0.646 & 0.534 & 0.623 & 0.837     & 0.690     & 0.801     & 0.810   \\
speech           & 0.572 & 0.517   & 0.533   & 0.524   & 0.551 & 0.459 & 0.485 & 0.458 & 0.516     & 0.458     & 0.499     & 0.458     & 0.531 & 0.472 & 0.513 & 0.571     & 0.520     & 0.513     & 0.515   \\
Stamps           & 0.598 & 0.514   & 0.823   & 0.442   & 0.452 & 0.421 & 0.379 & 0.354 & 0.809     & 0.939     & 0.754     & 0.898     & 0.558 & 0.467 & 0.621 & 0.660     & 0.937     & 0.924     & 0.911   \\
thyroid          & 0.919 & 0.757   & 0.943   & 0.868   & 0.571 & 0.544 & 0.572 & 0.444 & 0.858     & 0.922     & 0.919     & 0.953     & 0.803 & 0.786 & 0.772 & 0.912     & 0.981     & 0.987     & 0.970   \\
vertebral        & 0.575 & 0.571   & 0.572   & 0.619   & 0.529 & 0.645 & 0.646 & 0.590 & 0.664     & 0.824     & 0.764     & 0.876     & 0.478 & 0.564 & 0.479 & 0.608     & 0.799     & 0.784     & 0.668   \\
vowels           & 0.829 & 0.870   & 0.953   & 0.932   & 0.613 & 0.541 & 0.565 & 0.501 & 0.917     & 0.928     & 0.936     & 0.925     & 0.777 & 0.900 & 0.892 & 0.942     & 0.983     & 0.997     & 0.994   \\
Waveform         & 0.737 & 0.739   & 0.820   & 0.809   & 0.720 & 0.684 & 0.722 & 0.667 & 0.894     & 0.883     & 0.900     & 0.834     & 0.792 & 0.791 & 0.797 & 0.919     & 0.861     & 0.887     & 0.752   \\
WBC              & 0.500 & 0.832   & 0.954   & 0.930   & 0.756 & 0.662 & 0.712 & 0.713 & 0.836     & 0.836     & 0.951     & 0.840     & 0.910 & 0.909 & 0.906 & 0.944     & 0.865     & 0.882     & 0.956   \\
WDBC             & 0.500 & 0.687   & 0.910   & 0.905   & 0.895 & 0.839 & 0.839 & 0.839 & 0.908     & 0.932     & 0.907     & 0.937     & 0.892 & 0.729 & 0.739 & 0.833     & 0.919     & 0.928     & 0.990   \\
Wilt             & 0.838 & 0.400   & 0.398   & 0.400   & 0.613 & 0.647 & 0.639 & 0.630 & 0.695     & 0.622     & 0.760     & 0.670     & 0.420 & 0.385 & 0.384 & 0.639     & 0.467     & 0.359     & 0.649   \\
wine             & 0.500 & 0.527   & 1.000   & 1.000   & 0.667 & 0.702 & 0.706 & 0.754 & 0.800     & 1.000     & 1.000     & 1.000     & 0.419 & 0.545 & 0.712 & 0.577     & 1.000     & 0.991     & 0.994   \\
WPBC             & 0.391 & 0.542   & 0.553   & 0.545   & 0.465 & 0.439 & 0.439 & 0.499 & 0.607     & 0.468     & 0.539     & 0.588     & 0.546 & 0.535 & 0.524 & 0.556     & 0.649     & 0.591     & 0.593   \\
yeast            & 0.566 & 0.480   & 0.476   & 0.483   & 0.480 & 0.500 & 0.469 & 0.491 & 0.534     & 0.513     & 0.548     & 0.544     & 0.452 & 0.489 & 0.479 & 0.574     & 0.497     & 0.499     & 0.522   \\
CIFAR10          & 0.726 & 0.714   & 0.785   & 0.734   & 0.751 & 0.692 & 0.683 & 0.763 & 0.816     & 0.824     & 0.807     & 0.867     & 0.806 & 0.759 & 0.816 & 0.847     & 0.857     & 0.893     & 0.911   \\
MNIST-C          & 0.913 & 0.611   & 0.654   & 0.631   & 0.655 & 0.607 & 0.620 & 0.612 & 0.789     & 0.773     & 0.768     & 0.750     & 0.666 & 0.644 & 0.649 & 0.797     & 0.792     & 0.784     & 0.931   \\
MVTec-AD         & 0.687 & 0.761   & 0.786   & 0.772   & 0.688 & 0.729 & 0.682 & 0.732 & 0.791     & 0.805     & 0.794     & 0.740     & 0.736 & 0.765 & 0.695 & 0.772     & 0.795     & 0.741     & 0.762   \\
SVHN             & 0.672 & 0.490   & 0.491   & 0.491   & 0.502 & 0.507 & 0.491 & 0.499 & 0.502     & 0.496     & 0.497     & 0.509     & 0.500 & 0.498 & 0.504 & 0.508     & 0.495     & 0.507     & 0.583   \\
mnist            & 0.956 & 0.771   & 0.800   & 0.786   & 0.674 & 0.706 & 0.673 & 0.661 & 0.862     & 0.807     & 0.815     & 0.831     & 0.689 & 0.760 & 0.712 & 0.845     & 0.854     & 0.825     & 0.823   \\
FashionMNIST     & 0.904 & 0.746   & 0.846   & 0.764   & 0.796 & 0.731 & 0.776 & 0.762 & 0.899     & 0.827     & 0.897     & 0.899     & 0.854 & 0.786 & 0.875 & 0.889     & 0.878     & 0.918     & 0.933   \\
20news           & 0.616 & 0.537   & 0.540   & 0.541   & 0.516 & 0.491 & 0.503 & 0.523 & 0.565     & 0.545     & 0.563     & 0.558     & 0.536 & 0.537 & 0.535 & 0.560     & 0.590     & 0.587     & 0.774   \\
Agnews           & 0.801 & 0.570   & 0.569   & 0.571   & 0.518 & 0.497 & 0.498 & 0.518 & 0.689     & 0.624     & 0.597     & 0.640     & 0.540 & 0.546 & 0.562 & 0.687     & 0.610     & 0.619     & 0.915   \\
Amazon           & 0.731 & 0.485   & 0.482   & 0.485   & 0.508 & 0.526 & 0.498 & 0.479 & 0.479     & 0.460     & 0.506     & 0.495     & 0.500 & 0.496 & 0.478 & 0.529     & 0.522     & 0.502     & 0.870   \\
Imdb             & 0.688 & 0.499   & 0.498   & 0.498   & 0.525 & 0.480 & 0.498 & 0.530 & 0.572     & 0.492     & 0.551     & 0.553     & 0.533 & 0.509 & 0.530 & 0.571     & 0.510     & 0.553     & 0.806   \\
Yelp             & 0.745 & 0.487   & 0.492   & 0.491   & 0.491 & 0.492 & 0.449 & 0.502 & 0.525     & 0.505     & 0.497     & 0.514     & 0.469 & 0.500 & 0.477 & 0.523     & 0.532     & 0.464     & 0.840   \\
\midrule
\textbf{Average} & 0.798 & 0.677   & 0.736   & 0.701   & 0.619 & 0.598 & 0.619 & 0.598 & 0.772     & 0.744     & 0.759     & 0.761     & 0.687 & 0.680 & 0.693 & 0.777     & 0.778     & 0.780     & 0.847   \\
\midrule
\textbf{std}     & 0.025 & 0.048   & 0.044   & 0.047   & 0.085 & 0.076 & 0.093 & 0.086 & 0.065     & 0.069     & 0.056     & 0.061     & 0.061 & 0.060 & 0.061 & 0.064     & 0.046     & 0.050     & 0.031   \\
\bottomrule
\end{tabular}
\caption{Test AUC and Standard Deviation on ADBench}
\label{tab:adbench3}
\end{sidewaystable}

\begin{sidewaystable}[h!]
\renewcommand\thetable{C.5}
\centering
\setlength{\tabcolsep}{1.0mm}
\fontsize{6pt}{7pt}\selectfont
\begin{tabular}{lcccccccccccccccccc|c}
\toprule
                 & XGBOD+RD & DSAD+RD & DSAD+HC & DSAD+AB & SB+RD & SB+HC & SB+DB & SB+AB & SB-NCE+RD & SB-NCE+HC & SB-NCE+DB & SB-NCE+AB & OC+RD & OC+HC & OC+AB & OC-NCE+RD & OC-NCE+HC & OC-NCE+AB & IMBoost \\
\midrule
ALOI             & 0.160 & 0.054   & 0.054   & 0.054   & 0.045 & 0.054 & 0.047 & 0.050 & 0.045     & 0.055     & 0.047     & 0.050     & 0.048 & 0.052 & 0.052 & 0.045     & 0.056     & 0.048     & 0.047   \\
annthyroid       & 0.694 & 0.112   & 0.434   & 0.189   & 0.118 & 0.118 & 0.192 & 0.129 & 0.199     & 0.399     & 0.262     & 0.357     & 0.176 & 0.114 & 0.136 & 0.248     & 0.355     & 0.399     & 0.427   \\
backdoor         & 0.922 & 0.205   & 0.383   & 0.453   & 0.062 & 0.043 & 0.069 & 0.073 & 0.385     & 0.469     & 0.339     & 0.385     & 0.086 & 0.160 & 0.272 & 0.245     & 0.740     & 0.443     & 0.767   \\
breastw          & 0.857 & 0.674   & 0.873   & 0.693   & 0.526 & 0.478 & 0.593 & 0.651 & 0.917     & 0.726     & 0.884     & 0.904     & 0.794 & 0.689 & 0.813 & 0.890     & 0.834     & 0.902     & 0.882   \\
campaign         & 0.548 & 0.127   & 0.131   & 0.129   & 0.132 & 0.091 & 0.142 & 0.091 & 0.228     & 0.240     & 0.200     & 0.191     & 0.132 & 0.124 & 0.141 & 0.168     & 0.254     & 0.175     & 0.299   \\
cardio           & 0.688 & 0.170   & 0.542   & 0.250   & 0.520 & 0.368 & 0.451 & 0.330 & 0.520     & 0.368     & 0.451     & 0.330     & 0.620 & 0.451 & 0.468 & 0.620     & 0.451     & 0.468     & 0.798   \\
Cardiotocography & 0.655 & 0.313   & 0.529   & 0.319   & 0.481 & 0.388 & 0.587 & 0.643 & 0.798     & 0.502     & 0.733     & 0.841     & 0.613 & 0.402 & 0.601 & 0.786     & 0.530     & 0.822     & 0.687   \\
celeba           & 0.340 & 0.022   & 0.022   & 0.022   & 0.032 & 0.023 & 0.064 & 0.026 & 0.174     & 0.214     & 0.141     & 0.229     & 0.038 & 0.023 & 0.024 & 0.131     & 0.180     & 0.211     & 0.132   \\
census           & 0.408 & 0.060   & 0.060   & 0.060   & 0.052 & 0.069 & 0.048 & 0.059 & 0.201     & 0.133     & 0.063     & 0.077     & 0.046 & 0.059 & 0.048 & 0.229     & 0.095     & 0.079     & 0.459   \\
cover            & 0.936 & 0.017   & 0.029   & 0.029   & 0.103 & 0.078 & 0.041 & 0.034 & 0.499     & 0.870     & 0.778     & 0.922     & 0.009 & 0.015 & 0.012 & 0.079     & 0.083     & 0.275     & 0.615   \\
donors           & 1.000 & 0.081   & 0.140   & 0.078   & 0.134 & 0.120 & 0.171 & 0.091 & 0.977     & 0.899     & 0.755     & 0.901     & 0.132 & 0.109 & 0.160 & 0.921     & 0.900     & 0.911     & 0.990   \\
fault            & 0.509 & 0.487   & 0.502   & 0.493   & 0.377 & 0.369 & 0.320 & 0.343 & 0.544     & 0.487     & 0.475     & 0.470     & 0.386 & 0.471 & 0.397 & 0.502     & 0.534     & 0.463     & 0.436   \\
fraud            & 0.693 & 0.061   & 0.259   & 0.225   & 0.249 & 0.304 & 0.269 & 0.308 & 0.431     & 0.719     & 0.267     & 0.739     & 0.064 & 0.156 & 0.132 & 0.239     & 0.707     & 0.698     & 0.691   \\
glass            & 0.046 & 0.449   & 0.480   & 0.556   & 0.054 & 0.055 & 0.054 & 0.054 & 0.261     & 0.607     & 0.054     & 0.054     & 0.307 & 0.295 & 0.268 & 0.324     & 0.862     & 0.848     & 0.350   \\
Hepatitis        & 0.167 & 0.479   & 0.296   & 0.385   & 0.275 & 0.320 & 0.344 & 0.355 & 0.335     & 0.281     & 0.279     & 0.416     & 0.351 & 0.274 & 0.218 & 0.360     & 0.323     & 0.389     & 0.443   \\
http             & 0.996 & 0.292   & 0.998   & 0.997   & 0.168 & 0.216 & 0.496 & 0.300 & 0.578     & 0.998     & 0.997     & 0.997     & 0.417 & 0.875 & 0.726 & 0.653     & 0.997     & 0.945     & 0.450   \\
InternetAds      & 0.466 & 0.408   & 0.572   & 0.526   & 0.585 & 0.413 & 0.355 & 0.461 & 0.573     & 0.530     & 0.624     & 0.585     & 0.604 & 0.423 & 0.513 & 0.604     & 0.563     & 0.567     & 0.782   \\
Ionosphere       & 0.722 & 0.846   & 0.921   & 0.855   & 0.521 & 0.519 & 0.464 & 0.505 & 0.837     & 0.792     & 0.844     & 0.909     & 0.830 & 0.857 & 0.777 & 0.884     & 0.923     & 0.873     & 0.693   \\
landsat          & 0.694 & 0.304   & 0.326   & 0.327   & 0.223 & 0.232 & 0.170 & 0.182 & 0.486     & 0.411     & 0.242     & 0.385     & 0.193 & 0.246 & 0.212 & 0.340     & 0.332     & 0.357     & 0.471   \\
letter           & 0.261 & 0.347   & 0.385   & 0.386   & 0.062 & 0.069 & 0.055 & 0.062 & 0.137     & 0.225     & 0.226     & 0.185     & 0.250 & 0.339 & 0.322 & 0.277     & 0.363     & 0.436     & 0.117   \\
Lymphography     & 0.044 & 0.420   & 0.857   & 0.867   & 0.091 & 0.125 & 0.108 & 0.108 & 0.462     & 0.917     & 1.000     & 0.917     & 0.502 & 0.630 & 0.638 & 0.539     & 0.881     & 0.881     & 0.750   \\
magic.gamma      & 0.809 & 0.609   & 0.694   & 0.611   & 0.514 & 0.425 & 0.485 & 0.537 & 0.514     & 0.425     & 0.485     & 0.537     & 0.606 & 0.553 & 0.542 & 0.606     & 0.553     & 0.542     & 0.774   \\
mammography      & 0.379 & 0.105   & 0.142   & 0.126   & 0.054 & 0.069 & 0.068 & 0.079 & 0.291     & 0.413     & 0.266     & 0.398     & 0.061 & 0.095 & 0.105 & 0.299     & 0.448     & 0.319     & 0.642   \\
musk             & 0.999 & 0.734   & 1.000   & 1.000   & 0.300 & 0.433 & 0.392 & 0.436 & 0.980     & 0.832     & 0.969     & 0.933     & 0.487 & 0.530 & 0.596 & 0.924     & 1.000     & 1.000     & 0.981   \\
optdigits        & 0.611 & 0.048   & 0.119   & 0.111   & 0.062 & 0.065 & 0.069 & 0.047 & 0.870     & 0.960     & 0.982     & 0.991     & 0.034 & 0.110 & 0.042 & 0.711     & 0.868     & 0.985     & 0.869   \\
Pageblocks       & 0.641 & 0.260   & 0.529   & 0.328   & 0.221 & 0.132 & 0.223 & 0.231 & 0.624     & 0.543     & 0.690     & 0.711     & 0.306 & 0.327 & 0.333 & 0.612     & 0.619     & 0.677     & 0.684   \\
pendigits        & 0.596 & 0.055   & 0.065   & 0.066   & 0.062 & 0.083 & 0.035 & 0.177 & 0.800     & 0.928     & 0.521     & 0.905     & 0.054 & 0.092 & 0.098 & 0.834     & 0.929     & 0.941     & 0.419   \\
Pima             & 0.447 & 0.415   & 0.447   & 0.414   & 0.465 & 0.353 & 0.433 & 0.434 & 0.533     & 0.495     & 0.464     & 0.419     & 0.446 & 0.417 & 0.424 & 0.491     & 0.481     & 0.464     & 0.588   \\
satellite        & 0.829 & 0.513   & 0.665   & 0.518   & 0.650 & 0.517 & 0.684 & 0.515 & 0.882     & 0.576     & 0.860     & 0.819     & 0.722 & 0.533 & 0.710 & 0.888     & 0.701     & 0.839     & 0.754   \\
satimage-2       & 0.736 & 0.068   & 0.937   & 0.373   & 0.441 & 0.876 & 0.258 & 0.258 & 0.895     & 0.930     & 0.941     & 0.928     & 0.573 & 0.805 & 0.178 & 0.842     & 0.930     & 0.927     & 0.872   \\
shuttle          & 0.994 & 0.172   & 0.882   & 0.135   & 0.882 & 0.862 & 0.754 & 0.593 & 0.635     & 0.862     & 0.776     & 0.806     & 0.504 & 0.584 & 0.572 & 0.667     & 0.688     & 0.710     & 0.972   \\
skin             & 0.994 & 0.310   & 0.318   & 0.317   & 0.238 & 0.302 & 0.283 & 0.234 & 0.597     & 0.333     & 0.283     & 0.295     & 0.234 & 0.212 & 0.344 & 0.430     & 0.212     & 0.387     & 0.728   \\
smtp             & 0.066 & 0.326   & 0.438   & 0.438   & 0.041 & 0.041 & 0.106 & 0.041 & 0.041     & 0.315     & 0.107     & 0.169     & 0.085 & 0.092 & 0.106 & 0.085     & 0.300     & 0.352     & 0.122   \\
SpamBase         & 0.820 & 0.423   & 0.475   & 0.424   & 0.566 & 0.361 & 0.541 & 0.414 & 0.784     & 0.473     & 0.790     & 0.765     & 0.549 & 0.417 & 0.493 & 0.764     & 0.633     & 0.763     & 0.789   \\
speech           & 0.027 & 0.016   & 0.017   & 0.017   & 0.019 & 0.019 & 0.016 & 0.016 & 0.017     & 0.038     & 0.019     & 0.016     & 0.021 & 0.018 & 0.021 & 0.022     & 0.031     & 0.021     & 0.040   \\
Stamps           & 0.216 & 0.111   & 0.534   & 0.138   & 0.157 & 0.167 & 0.164 & 0.097 & 0.407     & 0.734     & 0.443     & 0.628     & 0.155 & 0.151 & 0.235 & 0.304     & 0.747     & 0.654     & 0.465   \\
thyroid          & 0.695 & 0.108   & 0.705   & 0.380   & 0.135 & 0.051 & 0.124 & 0.038 & 0.499     & 0.756     & 0.623     & 0.791     & 0.188 & 0.167 & 0.134 & 0.502     & 0.869     & 0.859     & 0.582   \\
vertebral        & 0.228 & 0.203   & 0.244   & 0.285   & 0.145 & 0.239 & 0.239 & 0.254 & 0.245     & 0.606     & 0.546     & 0.681     & 0.140 & 0.231 & 0.138 & 0.222     & 0.606     & 0.492     & 0.298   \\
vowels           & 0.531 & 0.223   & 0.412   & 0.325   & 0.094 & 0.052 & 0.046 & 0.044 & 0.633     & 0.746     & 0.771     & 0.670     & 0.269 & 0.272 & 0.291 & 0.562     & 0.826     & 0.964     & 0.871   \\
Waveform         & 0.099 & 0.344   & 0.653   & 0.611   & 0.077 & 0.062 & 0.108 & 0.054 & 0.425     & 0.645     & 0.643     & 0.638     & 0.526 & 0.569 & 0.560 & 0.615     & 0.675     & 0.631     & 0.208   \\
WBC              & 0.045 & 0.297   & 0.550   & 0.573   & 0.283 & 0.269 & 0.332 & 0.316 & 0.494     & 0.390     & 0.510     & 0.354     & 0.298 & 0.430 & 0.339 & 0.398     & 0.348     & 0.324     & 0.519   \\
WDBC             & 0.027 & 0.084   & 0.702   & 0.675   & 0.453 & 0.247 & 0.238 & 0.247 & 0.284     & 0.708     & 0.634     & 0.716     & 0.340 & 0.186 & 0.207 & 0.369     & 0.808     & 0.831     & 0.768   \\
Wilt             & 0.305 & 0.041   & 0.041   & 0.041   & 0.067 & 0.085 & 0.085 & 0.076 & 0.090     & 0.090     & 0.142     & 0.084     & 0.043 & 0.040 & 0.040 & 0.074     & 0.051     & 0.040     & 0.101   \\
wine             & 0.077 & 0.138   & 1.000   & 1.000   & 0.144 & 0.190 & 0.199 & 0.212 & 0.204     & 1.000     & 1.000     & 1.000     & 0.104 & 0.148 & 0.160 & 0.176     & 1.000     & 0.833     & 0.956   \\
WPBC             & 0.204 & 0.282   & 0.295   & 0.285   & 0.232 & 0.210 & 0.199 & 0.231 & 0.343     & 0.293     & 0.385     & 0.339     & 0.319 & 0.335 & 0.283 & 0.267     & 0.413     & 0.345     & 0.304   \\
yeast            & 0.397 & 0.339   & 0.351   & 0.341   & 0.326 & 0.342 & 0.310 & 0.331 & 0.362     & 0.356     & 0.363     & 0.385     & 0.322 & 0.347 & 0.335 & 0.390     & 0.349     & 0.352     & 0.368   \\
CIFAR10          & 0.180 & 0.118   & 0.325   & 0.171   & 0.245 & 0.171 & 0.153 & 0.290 & 0.367     & 0.551     & 0.354     & 0.567     & 0.290 & 0.198 & 0.318 & 0.351     & 0.562     & 0.580     & 0.640   \\
MNIST-C          & 0.701 & 0.125   & 0.265   & 0.188   & 0.159 & 0.150 & 0.125 & 0.125 & 0.292     & 0.497     & 0.352     & 0.369     & 0.196 & 0.180 & 0.155 & 0.309     & 0.494     & 0.394     & 0.801   \\
MVTec-AD         & 0.487 & 0.343   & 0.434   & 0.413   & 0.282 & 0.436 & 0.276 & 0.339 & 0.385     & 0.563     & 0.468     & 0.450     & 0.295 & 0.412 & 0.314 & 0.353     & 0.507     & 0.374     & 0.435   \\
SVHN             & 0.135 & 0.054   & 0.053   & 0.054   & 0.054 & 0.056 & 0.052 & 0.052 & 0.055     & 0.053     & 0.053     & 0.055     & 0.054 & 0.054 & 0.053 & 0.054     & 0.052     & 0.055     & 0.084   \\
mnist            & 0.732 & 0.299   & 0.437   & 0.343   & 0.242 & 0.277 & 0.253 & 0.241 & 0.542     & 0.562     & 0.455     & 0.590     & 0.256 & 0.318 & 0.291 & 0.520     & 0.629     & 0.561     & 0.517   \\
FashionMNIST     & 0.550 & 0.271   & 0.569   & 0.320   & 0.501 & 0.324 & 0.364 & 0.497 & 0.684     & 0.654     & 0.662     & 0.774     & 0.559 & 0.379 & 0.612 & 0.628     & 0.720     & 0.797     & 0.730   \\
20news           & 0.132 & 0.074   & 0.070   & 0.073   & 0.059 & 0.065 & 0.061 & 0.077 & 0.098     & 0.119     & 0.093     & 0.123     & 0.065 & 0.068 & 0.067 & 0.088     & 0.128     & 0.124     & 0.341   \\
Agnews           & 0.253 & 0.069   & 0.070   & 0.069   & 0.068 & 0.057 & 0.059 & 0.067 & 0.116     & 0.174     & 0.106     & 0.173     & 0.072 & 0.064 & 0.068 & 0.114     & 0.117     & 0.103     & 0.544   \\
Amazon           & 0.132 & 0.049   & 0.049   & 0.049   & 0.053 & 0.054 & 0.052 & 0.051 & 0.052     & 0.048     & 0.054     & 0.051     & 0.052 & 0.052 & 0.049 & 0.056     & 0.056     & 0.054     & 0.305   \\
Imdb             & 0.128 & 0.056   & 0.056   & 0.056   & 0.061 & 0.052 & 0.053 & 0.061 & 0.074     & 0.058     & 0.064     & 0.068     & 0.064 & 0.059 & 0.060 & 0.073     & 0.061     & 0.066     & 0.351   \\
Yelp             & 0.171 & 0.050   & 0.050   & 0.050   & 0.051 & 0.051 & 0.045 & 0.056 & 0.058     & 0.052     & 0.049     & 0.054     & 0.049 & 0.051 & 0.050 & 0.060     & 0.059     & 0.049     & 0.476   \\
\midrule
\textbf{Average} & 0.477 & 0.240   & 0.410   & 0.337   & 0.232 & 0.221 & 0.227 & 0.221 & 0.434     & 0.502     & 0.466     & 0.511     & 0.280 & 0.285 & 0.285 & 0.417     & 0.516     & 0.519     & 0.547   \\
\midrule
\textbf{std}     & 0.032 & 0.063   & 0.076   & 0.073   & 0.075 & 0.062 & 0.090 & 0.081 & 0.104     & 0.093     & 0.095     & 0.092     & 0.080 & 0.069 & 0.074 & 0.104     & 0.071     & 0.095     & 0.086   \\
\bottomrule
\end{tabular}
\caption{Test AP and Standard Deviation on ADBench}
\label{tab:adbench4}
\end{sidewaystable}

\clearpage
\subsection{C-\MakeUppercase{\romannumeral 3}. Query Strategies for Active Learning}
\vspace{0.2cm}
\vspace{0.2cm}
\textbf{Comparison of Query Strategies} \hspace{0.2cm} We compare the effectiveness of three query strategies for active learning, including Random (RD), Confidence Poles (CP), Decision Boundary Using Mixture Model (MM). 
Table \ref{tab:query_strategies} shows the averaged test AUC scores and standard deviations on \texttt{ADBench} datasets after five ronuds of active learning.
Among all strategies, the MM strategy achieves the best final performance with an AUC of 0.847, demonstrating its advantage in selecting informative samples for model improvement.
In addition, MM shows the smallest standard deviation of 0.031, suggesting high robustness and consistency across different trials.
\begin{table}[h!]
\renewcommand\thetable{C.6}
\centering
\setlength{\tabcolsep}{2mm} 
\fontsize{10pt}{11.6pt}\selectfont 
\begin{tabular}{l|ccc}
\hline
\textbf{Query Strategies} & \textbf{RD} & \textbf{CP} & \textbf{MM}    \\
\hline
\textbf{$\text{AUC} \pm \text{std}$}  & $0.839 \pm 0.040$  & $0.797 \pm 0.035$  & $\textbf{0.847} \pm 0.031$  \\
\hline
\end{tabular}
\caption{Averaged results of test AUC scores and standard deviations with various query strategies}
\label{tab:query_strategies}
\end{table}

\noindent
\textbf{Performance over Active Learning Rounds} \hspace{0.2cm} Table \ref{tab:RD_per_round} reports the averaged test AUC scores and standard deviations on \texttt{ADBench} datasets for each query strategy at every round of active learning.
We observe that all strategies improve their performance as the number of labeled samples increases.
Notably, the MM strategy outperforms both RD and CP in all rounds except for Round1, where RD shows slightly better performance.
The MM strategy achieves the highest AUC of 0.847 at Round 5. 
\begin{table}[h!]
\renewcommand\thetable{C.7}
\centering
\setlength{\tabcolsep}{2mm} 
\fontsize{10pt}{11.6pt}\selectfont 
\begin{tabular}{l|ccccc}
\hline
\textbf{Query Strategies} & \textbf{RD} & \textbf{CP} & \textbf{MM} \\
\hline
\textbf{Round1} & $0.796 \pm 0.042$  & $0.771 \pm 0.047$   & $0.785 \pm 0.048$  \\
\textbf{Round2} & $0.809 \pm 0.046$  & $0.778 \pm 0.040$   & $0.811 \pm 0.046$  \\
\textbf{Round3} & $0.824 \pm 0.043$  & $0.790 \pm 0.039$   & $0.827 \pm 0.038$  \\
\textbf{Round4} & $0.831 \pm 0.042$  & $0.799 \pm 0.036$   & $0.837 \pm 0.037$  \\
\textbf{Round5} & $0.839 \pm 0.040$  & $0.797 \pm 0.035$   & $0.847 \pm 0.031$  \\
\hline
\end{tabular}
\caption{Averaged results of test AUC scores and standard deviations over rounds with various query strategies}
\label{tab:RD_per_round}
\end{table}

\subsection{C-\MakeUppercase{\romannumeral 4}. Ablation Studies}
\vspace{0.2cm}
\textbf{Weighting Parameters for Inlier and Outlier Losses} \hspace{0.2cm} 
Table \ref{tab:ablation_lambda1} and \ref{tab:ablation_lambda2} show the averaged test AUC scores on \texttt{ADBench} datasets for different values of two weighting parameters.
The best performance, with an AUC of 0.847 is achieved when $\lambda_{1,t}=2$ and $\lambda_{2,t}=1$, indicating that appropriate non-zero weights of the inlier and outlier terms is crucial for optimal performance.
In contrast, setting $\lambda_{1,t}$ or $\lambda_{2,t}$ to zero (i.e., $\lambda = 0$) yields the lowest performance, which demonstrates ignoring either component significantly degrades the model.
These findings underscore the importance of carefully balancing the contributions of inlier and outlier loss components, as overly small or excessively large weights lead to suboptimal results.
\begin{table}[h!]
\renewcommand\thetable{C.8}
\centering
\setlength{\tabcolsep}{2mm} 
\fontsize{10pt}{11.6pt}\selectfont 
\begin{tabular}{l|ccccccc}
\hline
$\mathbf{\lambda_{1,t}}$ & \textbf{0} & \textbf{1} & \textbf{2} & \textbf{3} & \textbf{5} & \textbf{7} & \textbf{10} \\
\hline
\textbf{AUC}  & 0.685  & 0.827  & \textbf{0.847}  & 0.829  & 0.818  & 0.809  & 0.804  \\
\hline
\end{tabular}
\caption{Averaged results of test AUC scores with various values of $\lambda_{1,t}$}
\label{tab:ablation_lambda1}
\end{table}

\begin{table}[h!]
\renewcommand\thetable{C.9}
\centering
\setlength{\tabcolsep}{2mm} 
\fontsize{10pt}{11.6pt}\selectfont 
\begin{tabular}{l|ccccccc}
\hline
$\mathbf{\lambda_{2,t}}$ & \textbf{0} & \textbf{1} & \textbf{2} & \textbf{3} & \textbf{5} & \textbf{7} & \textbf{10} \\
\hline
\textbf{AUC}  & 0.758  & \textbf{0.847}  & 0.826  & 0.794  & 0.752  & 0.739  & 0.725  \\
\hline
\end{tabular}
\caption{Averaged results of test AUC scores with various values of $\lambda_{2,t}$}
\label{tab:ablation_lambda2}
\end{table}

\newpage
\noindent
\textbf{Inlier Risk Ratio in Adaptive Threshold} \hspace{0.2cm} 
As shown in Table \ref{tab:ablation_xi}, the best averaged test AUC score on \texttt{ADBench} datasets is 0.847, which is obtained at $\xi = 0.4$.
In contrast, extreme values close to 0.0 or 1.0 result in lower AUCs, indicating that appropriately balancing the contributions from quantile risk and inlier risk is essential for optimal performance.
These findings highlight the importance of choosing an appropriately balanced value of $\xi$, rather than relying solely on either component.
\begin{table}[h!]
\renewcommand\thetable{C.10}
\centering
\setlength{\tabcolsep}{2mm} 
\fontsize{10pt}{11.6pt}\selectfont 
\begin{tabular}{l|cccccc}
\hline
$\mathbf{\xi}$ & \textbf{0.0} & \textbf{0.2} & \textbf{0.4} & \textbf{0.6} & \textbf{0.8} & \textbf{1.0} \\
\hline
\textbf{AUC}  & 0.836  & 0.840  & \textbf{0.847} & 0.832  & 0.838  & 0.835  \\
\hline
\end{tabular}
\caption{Averaged results of test AUC scores with various values of $\xi$}
\label{tab:ablation_xi}
\end{table}\

\noindent
\textbf{Posterior Probability} \hspace{0.2cm} Table \ref{tab:ablation_alpha} shows the averaged test AUC scores on \texttt{ADBench} datasets for various values of posterior probability threshold $\alpha$.
The best performance is achieved at $\alpha=0.4$ with an AUC of 0.847.
This observation is consistent with the theoretical justification that setting $\alpha$ close to 0.5 helps balance the number of labeld inliers and outliers, thereby maximizing the increase in outlier risk and enhancing overall performance.
\begin{table}[h!]
\renewcommand\thetable{C.11}
\centering
\setlength{\tabcolsep}{2mm} 
\fontsize{10pt}{11.6pt}\selectfont 
\begin{tabular}{l|ccccc}
\hline
$\mathbf{\alpha}$ & \textbf{0.3} & \textbf{0.4} & \textbf{0.5} & \textbf{0.6} & \textbf{0.7}  \\
\hline
\textbf{AUC}  & 0.843  & \textbf{0.847} & 0.841  & 0.842  & 0.842  \\
\hline
\end{tabular}
\caption{Averaged results of test AUC scores with various values of $\alpha$}
\label{tab:ablation_alpha}
\end{table}

\noindent
\textbf{Warm-up Phase} \hspace{0.2cm} Table \ref{tab:ablation_T_1} presents the averaged test AUC scores on \texttt{ADBench} datasets for various values of the warm-up phase length $T_1$.
The best performance is observed at $T_1=40$, with an AUC of 0.847.
These results highlight the importance of selecting an appropriate warm-up duration before transitioning to the polarization phase.
In contrast, both shorter and longer warm-up durations result in performance degradation, highlighting the importance of selecting appropriate value of $T_1$.
The warm-up phase should be long enough to stabilize the model, but not so long that it delays the polarization phase and negatively affects performance.
\begin{table}[h!]
\renewcommand\thetable{C.12}
\centering
\setlength{\tabcolsep}{2mm} 
\fontsize{10pt}{11.6pt}\selectfont 
\begin{tabular}{l|ccccc}
\hline
$\mathbf{T_1}$ & \textbf{10} & \textbf{20} & \textbf{30} & \textbf{40} & \textbf{50}  \\
\hline
\textbf{AUC}  & 0.824  & 0.840  & 0.836  & \textbf{0.847}  & 0.829  \\
\hline
\end{tabular}
\caption{Averaged results of test AUC scores with various values of $T_1$}
\label{tab:ablation_T_1}
\end{table}

\noindent
\textbf{Running Time} \hspace{0.2cm} Table \ref{tab:ablation_running_time} reports the running times (in seconds) for processing all \texttt{ADBench} datasets using the best-performing methods from each category: OC-NCE+AB for the OC-based methods, SB-NCE+RD for the SB-based methods and our proposed method, IMBoost.
All measurements are obtained using the full \texttt{ADBench} datasets with a single random seed.
Among the three, IMBoost achieves the lowest running time of 4484.17 seconds, demonstrating its superior computational efficiency.
\begin{table}[h!]
\renewcommand\thetable{C.13}
\centering
\setlength{\tabcolsep}{2mm} 
\fontsize{10pt}{11.6pt}\selectfont 
\begin{tabular}{l|ccc}
\hline
\textbf{Query Strategies} & \textbf{OC-NCE+AB} & \textbf{SB-NCE+RD} & \textbf{IMBoost} \\
\hline
\textbf{Running time (s)} & 6900.44            & 6966.73            & 4484.17     \\
\hline
\end{tabular}
\caption{Running time with various query strategies}
\label{tab:ablation_running_time}
\end{table}


\end{document}